\documentclass{article}

\usepackage[dblblindworkshop, final]{neurips_2025}
\usepackage{xcolor}         % colors

\usepackage{times}
\usepackage{latexsym}
\usepackage{enumitem}
\usepackage[T1]{fontenc}
\usepackage[utf8]{inputenc}

\usepackage{inconsolata}

\usepackage{graphicx}

\usepackage{amsmath}
\usepackage{hyperref}       % hyperlinks
\usepackage{url}            % simple URL typesetting
\usepackage{amsfonts}       % blackboard math symbols
\usepackage{nicefrac}       % compact symbols for 1/2, etc.
\usepackage{svg}[inkscapelatex=false]
\usepackage{mathabx}

\usepackage{multirow} 
\usepackage{graphicx}
\usepackage[export]{adjustbox}
\usepackage{float}
\usepackage{epsfig}
\usepackage{graphicx}

\usepackage{amssymb} % http://ctan.org/pkg/amssymb
\usepackage{caption}
\usepackage{xparse} % data cards
\usepackage{tabularx}
\usepackage{subcaption}
\usepackage{fancyhdr}
\usepackage[hang,flushmargin]{footmisc} % remove indentation in footnote
\usepackage{pifont} % http://ctan.org/pkg/pifont
\usepackage{color}
\usepackage{wrapfig}
\usepackage{subcaption}
\usepackage{multirow}
\usepackage{boxedminipage}
\usepackage{framed}
\usepackage{soul}
\usepackage{xspace}
\usepackage{booktabs} % professional-quality tables

\usepackage{makecell}
\usepackage{xcolor}
\usepackage{vcell}
\usepackage{colortbl}
\usepackage[nodisplayskipstretch]{setspace}
\usepackage{seqsplit} 
\usepackage{array}
\usepackage{todonotes}
\usepackage{CJKutf8} % Load the CJKutf8 package(Chinese)
\addtocontents{toc}{\protect\setcounter{tocdepth}{-1}}
\usepackage[final,nopatch=footnote]{microtype}

\newcommand{\header}[1]{\text{#1}}

\definecolor{backred}{RGB}{242, 121, 112}
\definecolor{backblue}{RGB}{210, 230, 250}
\definecolor{backgreen}{RGB}{187, 253, 207}
\definecolor{backyellow}{RGB}{253, 255, 167}
\definecolor{backpurple}{RGB}{233, 213, 255}
\newcommand{\high}{\cellcolor{backpurple}}

\newcommand{\best}{\cellcolor{backgreen}}

\definecolor{my_green}{RGB}{51,102,0}
\definecolor{my_yellow}{RGB}{255,165,0}
\definecolor{my_red}{RGB}{204, 0, 0}

\newcommand{\red}[1]{\textcolor{red}{#1}}

\newcommand{\green}[1]{\textcolor{my_green}{#1}}

\definecolor{shadecolor}{RGB}{237,237,237}

\usepackage{tcolorbox}

\definecolor{myboxcolor}{RGB}{207,234,241} 
\definecolor{myframe}{RGB}{73,108,136} 

\newtcolorbox{mybody}{
  colback=myboxcolor,
  colframe=myframe,
  boxrule=1pt, % Adjust the border thickness
  left=1pt,
  right=1pt,
  top=1pt,
  bottom=1pt,
}

\title{\textit{VisAidMath}: Benchmarking Visual-Aided Mathematical Reasoning}
\workshoptitle{VLM4RWD}

\author{
        Jingkun Ma$^1$~~~
        Runzhe Zhan$^1$~~~
        Yang Li$^1$~~~Di Sun$^2$\\
        \textbf{Hou Pong Chan}$^3$~~~
        \textbf{Lidia S. Chao}$^1$~~~
        \textbf{Derek F. Wong}$^1$\thanks{Corresponding author.}\\
    $^1$NLP\textsuperscript{2}CT Lab, Department of Computer and Information Science, University of Macau \\
    $^2$ University of Macau \quad $^3$DAMO Academy, Alibaba Group \\
    \texttt{nlp2ct.\{jingkun, runzhe, liyang\}@gmail.com},~~\texttt{yc47479@um.edu.mo}\\
    \texttt{houpong.chan@alibaba-inc.com},~~\texttt{\{lidiasc, derekfw\}@um.edu.mo} \\
    ~\\
    \textbf{Homepage:} \url{https://nlp2ct.github.io/VisAidMathHomepage/} \\
    \textbf{Evaluation:} \url{https://www.codabench.org/competitions/7634/}\\
    }

\begin{document}

\maketitle

\begin{abstract}
A hallmark of advanced artificial intelligence is the capacity to progress from passive visual perception to the strategic modification of visual information to facilitate complex reasoning. This advanced capability, however, remains critically underdeveloped in current Large Multi-modal Models (LMMs). The deficiency is often masked by evaluation metrics that prioritize final-answer accuracy, creating an illusion of competence where genuine reasoning is absent. Using the domain of geometric problem-solving as a precise instrument, we probe this issue through tasks that require constructing visual aids.
To this end, we introduce \textbf{VisAidMath}, a challenging benchmark, and our novel Three-Layered Funnel Evaluation Framework. This framework moves beyond simple accuracy (ACCU) to scrutinize the generation of valid visual aids (PVA) and the soundness of subsequent reasoning steps (SPRS). Our extensive experiments on state-of-the-art models, including Doubao-Seed-1.6 and o4, reveal a profound ``Reasoning Illusion''. We observe that high surface-level accuracy conceals a catastrophic failure in the models' ability to produce valid visual aids or to reason from them. Our findings expose a fundamental schism between visual perception and logical deduction in modern LMMs. 
% We thus contribute a rigorous benchmark and evaluation paradigm to steer the development of future models capable of true, visually-grounded reasoning. 
We host an evaluation platform at CodaBench for testing publicly.
\end{abstract}

\section{Introduction}
\begin{wrapfigure}{r}{0.5\textwidth}
    \centering
    \vspace{-0.8cm}
    \includegraphics[width=0.9\linewidth]{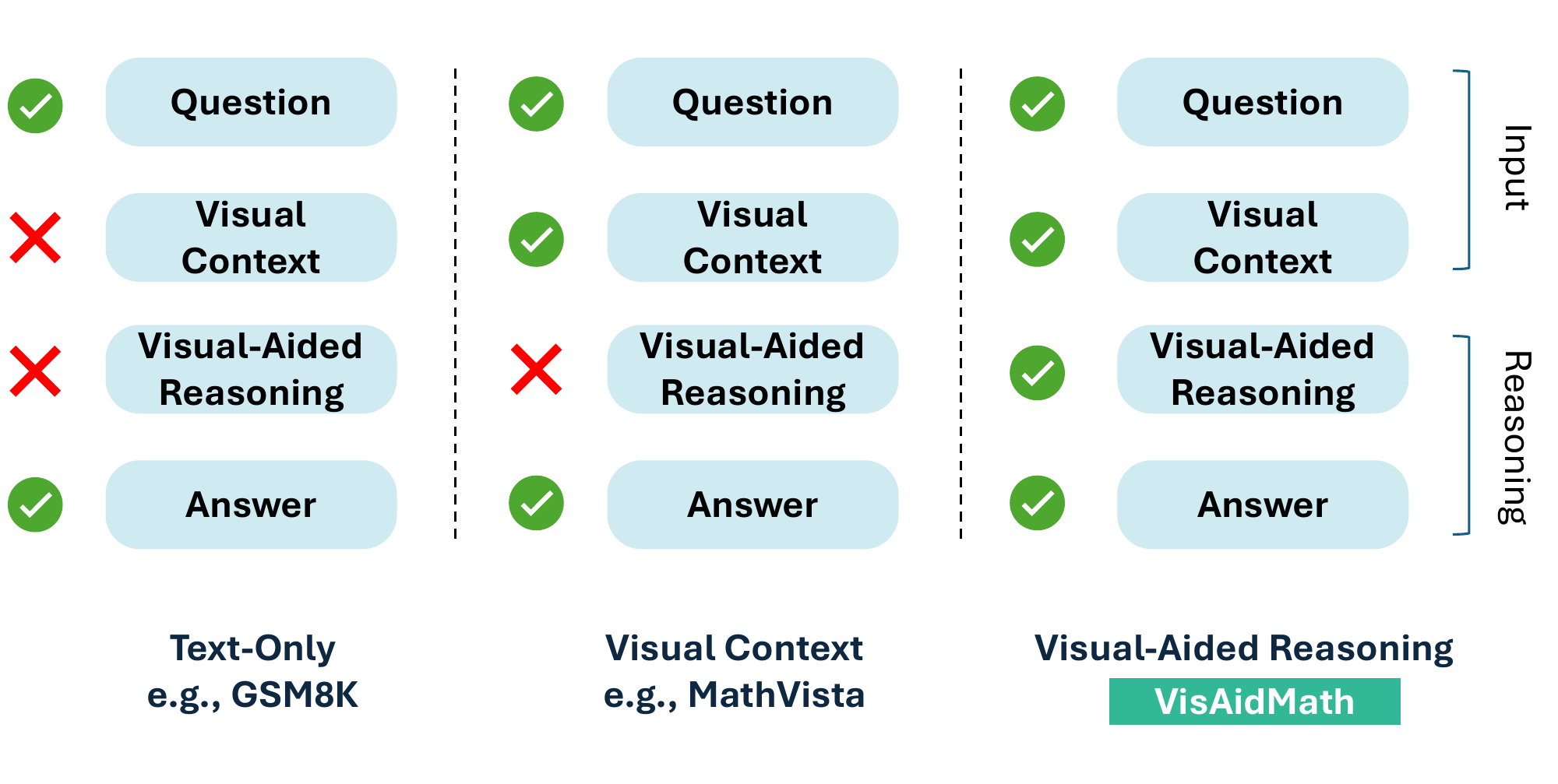}
    \caption{Comparison between VisAidMath and other benchmarks. Our work particularly focuses on utilization of explicit and implicit visual context during reasoning process.}
    \label{Figure: Motivation}
    \vspace{-0.5cm}
\end{wrapfigure}
Mathematical problem-solving (MPS) remains a significant hurdle for Large Language Models (LLMs) and Large Multi-modal Models (LMMs) \cite{hendrycks2021measuring, lewkowycz2022solving, wu2020knowledge, wu2021lime}. The complexity intensifies when problems are presented in a multi-modal format \cite{lindstrom2022clevr, masry2022chartqa}. While some research has explored multi-modal MPS by incorporating visual contexts \cite{masry2022chartqa,lu2023mathvista}, these efforts predominantly focus on traditional vision-language task paradigms. Consequently, they tend to evaluate text-only reasoning steps, often underutilizing the visual information crucial for complex reasoning \cite{su2025thinking}. This creates a significant gap between evaluating surface-level ``Think about Images'' and ``Think with Images'' \cite{su2025thinking}, a gap our work aims to address as illustrated in Figure \ref{Figure: Motivation}.

\begin{figure*}[t]
  \centering
  \includegraphics[width=1.0\textwidth]{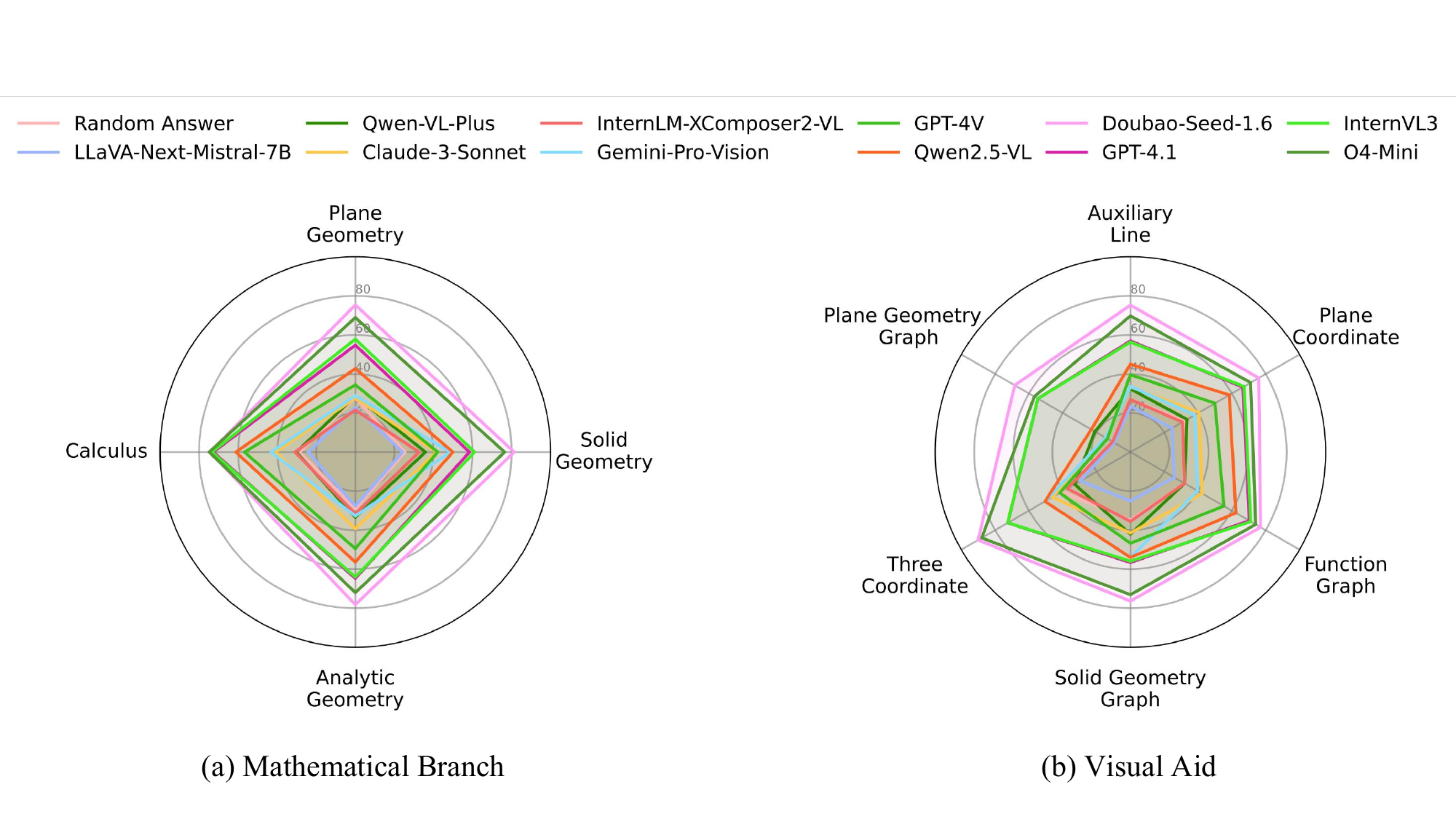}
  \caption{Accuracies of all LMM on visual-aided mathematical reasoning task across four branches and six visual aids.} % The best performing GPT-4V cannot exceed 60\% accuracy score.
  \vspace{-0.3cm}
  \label{Figure: Mathematics branch Radar}
  \vspace{-0.4cm}
\end{figure*}

To bridge this gap, we introduce \textbf{VisAidMath}, a benchmark specifically designed to compel and evaluate the process of visual-aided mathematical reasoning. Drawing from diverse, high-quality sources, VisAidMath consists of 1,200 problems structured to necessitate generation or utilization of visual aids. This design moves beyond simple visual comprehension to directly target a model's ability to perform complex spatial and logical deductions grounded in visual context. Crucially, VisAidMath provides a unified testbed to evaluate model capabilities across the full evolutionary spectrum of “Thinking with Images” paradigm \cite{su2025thinking}. Problems within our benchmark can be explored through the selection of analytical tools \cite{hu2022promptcap,wu2024mind,liu2025visual,qi2024cogcom}, the programmatic creation of visual aids like auxiliary lines \cite{gupta2023visual,fu2025refocus,chervonyi2025gold,weng2025geosketch}, or by engaging in intrinsic spatial imagination to foresee the solution path \cite{team2024chameleon,guo2025can,zhao2025cot}.

To demonstrate benchmark's unique value, we propose a three-layered funnel evaluation framework that moves beyond standard accuracy (ACCU), which often masks procedural flaws. This framework assesses Process-Verified Accuracy (PVA), filtering out answers from flawed reasoning, and Solution Process Robustness Score (SPRS) to quantify the fine-grained quality of solution. Applying this rigorous evaluation reveals a universal “reasoning illusion”: a dramatic collapse from high accuracy to poor process quality across all tested models. This illusion is most pronounced on the core tasks of VisAidMath, where even top models like Doubao-Seed-1.6 and O4-Mini show a staggering performance drop. This collapse reveals a fundamental weakness in generating and utilizing visual aids, a deficiency completely missed by standard metrics.

This paper's contributions are threefold: (1) We introduce \textbf{VisAidMath}, a novel benchmark that mandates visual-aided reasoning for mathematical problem-solving. (2) We propose a new \textbf{evaluation framework} that uncovers the widespread ``reasoning illusion'' in current SOTA models. (3) Through comprehensive analysis, we provide quantitative proof that VisAidMath is uniquely effective at exposing these deep-seated reasoning failures, thereby establishing its value and pinpointing critical areas for the future development of more robust and reliable multi-modal models.

\section{\textit{VisAidMath}}\label{sec:Dataset}
\subsection{Data Creation}\label{data creation}

\paragraph{Principles}
A typical problem within our \textit{VisAidMath} benchmark comprises four parts: \textbf{Visual Context (C)}, \textbf{Question (Q)}, \textbf{Visual Aids (V)}, and \textbf{Answer (A)}.
The main task involves prompting the model to generate visual aids that assist in mathematical reasoning, a key distinction from other benchmarks as detailed in appendix \ref{related work: multimodal benchmark}. While the visual context may be optional, spatial descriptions are included as essential data elements within the question. Given that many text-based LLMs lack image understanding or generation capabilities, we have additionally annotated precise captions for both the visual context and the visual aids through annotation. This allows us to extend the evaluation scenarios to models that are constructed with limited modality.

\paragraph{Data Sources and Categories}
We collected the VisAidMath benchmark from both English and Chinese sources. 
All data sources were categorized to ensure a balanced range of difficulty. A comprehensive description of our data sources, collection methodology, metadata, and translation process is detailed in \ref{Appendix: data source} and \ref{Appendix: metadata}.
To ensure diversity and balance, we manually collected and annotated a range of categories within the benchmark.
Detailed categories and examples from different categorizations can be found in section \ref{Appendix: examples for different categorizations}.

\paragraph{Construction Pipeline} 
As depicted in Figure \ref{Figure: Data Collection And Annotation Pipeline}, we propose a construction pipeline for the VisAidMath dataset, which incorporates multi-round verification and dynamic quality control based on feedback. The dataset creation pipeline involves four key roles (see Appendix \ref{key role detail}). To prepare the benchmark for the wide English research community, we perform several steps. This includes machine translation, for which we selected DeepL and Baidu Translate after a manual sampling process confirmed their high quality for technical content. This is followed by data processing and release preparation. The detailed processes can be found in Appendix \ref{dataset collection detail}, with the annotation process specified in Appendix \ref{annotation details}.

\subsection{Benchmark Attributes}
The distribution of data sources is presented in Figure \ref{Data Source and Complexity Distribution}, providing a comprehensive overview of the dataset's origins. Additionally, the mathematical branches within the dataset exhibit a well-balanced distribution. This balance enables a broader exploration of diverse mathematical knowledge.
Further details on other attributes can be found in Appendix \ref{Appendix: Dataset Analysis}.

\begin{figure*}[t]
  % \begin{center}
  \centering
    \includegraphics[width=0.9\textwidth]{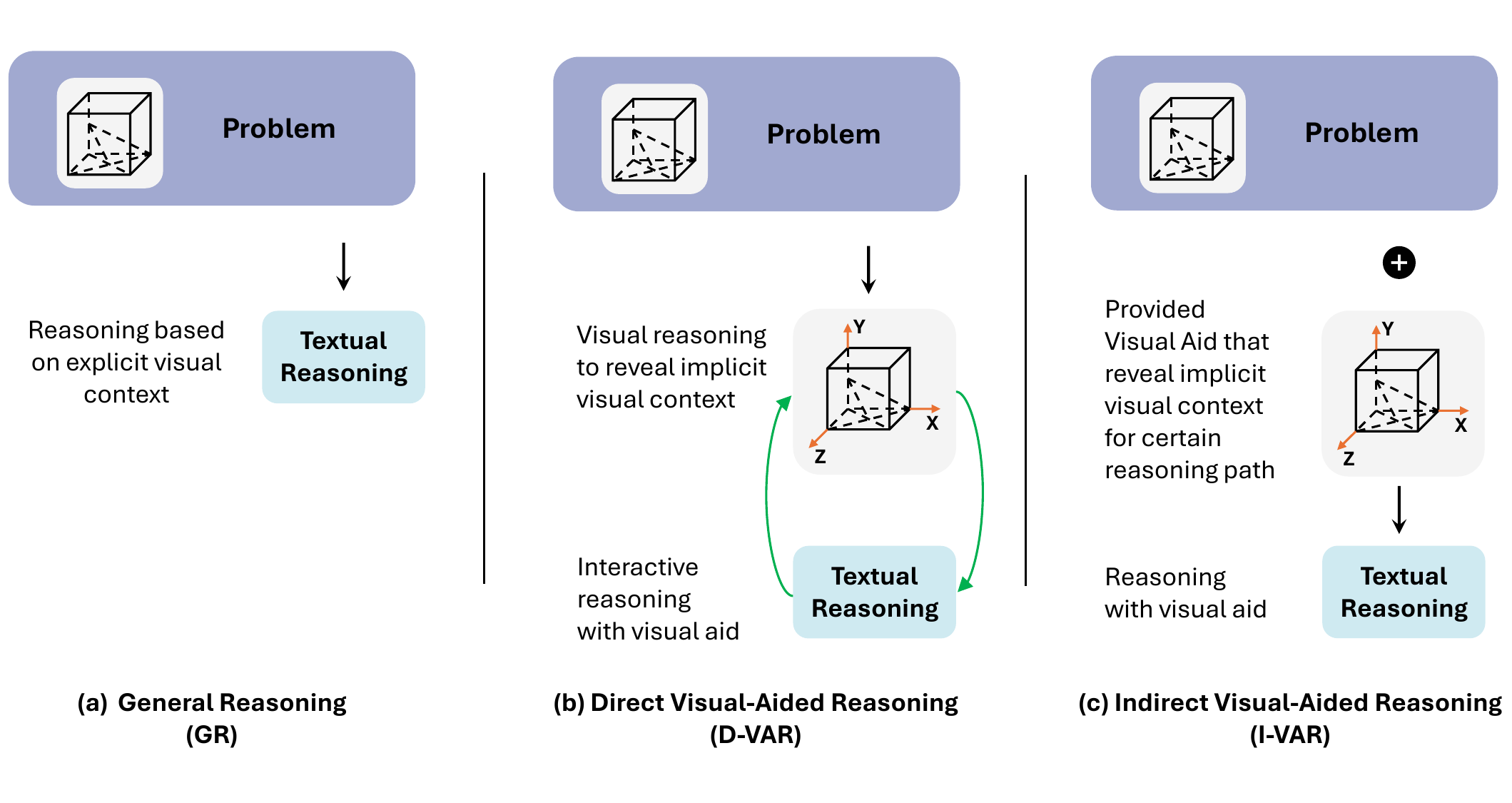}
    % \includesvg[width=0.65\textwidth]{graphs/double_layer_pie_chart_no_labels.svg}
    \vspace{-0.2cm}
    \caption{Comparison of different tasks: a) General Reasoning: provide MPS reasoning steps directly. b) Direct Visual-Aided Reasoning: create visual aids that disclose implicit visual context within problem, incorporating with textual reasoning to solve mathematical problem. c) Indirect Reasoning: solve the mathematical problem based on given visual aids. Direct visual-aided reasoning require the model to perform visual reasoning for visual aids generation.}
    \label{Figure: Task Comparison}
    \vspace{-0.3cm}
  % \end{center}
\end{figure*}

\begin{table*}[htbp!]
\centering
\caption{Formal definitions of the reasoning tasks. $V_g$ denotes a generated visual aid (output), while $V_p$ denotes a provided visual aid (input).}
\resizebox{\linewidth}{!}{%
\begin{tabular}{l|c|c|l}
\toprule
\textbf{Task Name} & \textbf{Abbr.} & \textbf{Formal Expression} & \textbf{Core Characteristic} \\
\midrule
General Reasoning & GR & $(C, Q) \rightarrow A$ & Directly solves the problem. \\
Direct Visual-Aided Reasoning & D-VAR & $(C, Q) \rightarrow (V_g, A)$ & \textbf{Generates} a visual aid to assist reasoning. \\
Indirect Visual-Aided Reasoning & I-VAR & $(C, Q, V_p) \rightarrow A$ & \textbf{Utilizes} a provided visual aid for reasoning. \\
\bottomrule
\end{tabular}%
}
\label{tab:task_definitions}
\end{table*}

\subsection{Task Definition}

Our work introduces a series of novel mathematical reasoning tasks centered on the use of visual aids. 

To formalize these tasks, we first define their basic components: the visual context (C), the textual question (Q), and the final answer (A). Crucially, a visual aid (V) in our framework is a \textbf{textual description} of a geometric construction.

We establish a baseline task, \textbf{General Reasoning (GR)}, and introduce two novel formulations based on how visual aids are utilized: \textbf{Direct Visual-Aided Reasoning (D-VAR)} tasks the model with generating a visual aid $V_{\text{g}}$ to solve a problem. In contrast, \textbf{Indirect Visual-Aided Reasoning (I-VAR)} requires the model to leverage a provided one $V_{\text{p}}$. The formal definitions and core distinctions for these tasks are detailed in Table~\ref{tab:task_definitions}.                               

To accommodate language models that only accept textual inputs, the visual context `C` is replaced by its textual description, denoted as $C_{\text{txt}}$. This creates text-only variants for each task, such as $(C_{\text{txt}}, Q) \rightarrow A$ for the GR task. This approach ensures a fair comparison across both multimodal and text-only models, as the visual aids themselves are consistently represented as text in all settings.

As defined in Table~\ref{tab:task_definitions}, the introduction of the D-VAR and I-VAR tasks distinguishes VisAidMath from traditional benchmarks. Unlike tasks that only require understanding existing visual elements, our approach compels models to engage in a deeper, two-step reasoning process: first, planning or interpreting geometric constructions (the visual aids), and then executing the subsequent reasoning steps. This novel modality, focused on the generation and utilization of textual visual aids, allows us to specifically probe the spatial and logical planning capabilities of MLLMs. We illustrate the distinctions between these tasks in Figure~\ref{Figure: Task Comparison}.
\begin{table*}[t!]
\vspace{-3mm}
\caption{Accuracy scores on \textbf{Direct Visual-Aided Reasoning (D-VAR)} task upon \textbf{VisAidMath}. Meanings of all abbreviations are: ALL $\rightarrow$ overall accuracy, PLG $\rightarrow$ plane geometry, SDG $\rightarrow$ solid geometry, AYG$\rightarrow$ analytic geometry, CAL: calculus and functions. Visual Aids Types: AXL $\rightarrow$ auxiliary line, RTC $\rightarrow$ rectangular coordinate, THC $\rightarrow$ rectangular three-dimensional coordinate, PLG $\rightarrow$ plane geometry graph, SDG $\rightarrow$ solid geometry graph, FUG $\rightarrow$ function graph. The highest scores in chunk and in general have been emphasized with purple and green to facilitate comparison respectively.}
\centering
 \renewcommand\tabcolsep{5pt} % column space
 \resizebox{1\linewidth}{!}{
    \begin{tabular}{l|c|cccc|cccccc}
    \toprule
    \header{Model} & \header{ALL} & \header{PLG} & \header{SDG} & \header{AYG} & \header{CAL} & \header{AXL} & \header{RTC} & \header{THC} & \header{PLG} & \header{SDG} & \header{FUG}  \\ 
    \midrule
    \multicolumn{12}{l}{\hfill \textit{Heuristics  Baselines}} \\
    \midrule
    % \rowcolor[rgb]{0.93,0.93,0.93} \multicolumn{11}{l}{\textit{Heuristic baselines}} \\
    Random Answer & 24.42 & 21.54 & 34.31 & 21.45 & 20.07 & 24.44 & 20.87 & 35.16 & 10.53 & 32.89 & 21.50 \\
    Frequent Answer & 40.83 & 28.92 & 50.65 & 40.36 & 44.22 & 32.79 & 47.25 & 74.73 & 20.00 & 47.73 & 44.53 \\
    \midrule
    \multicolumn{12}{l}{\hfill \textit{Large Language Models (LLMs): Text-Only Input} } \\
    \midrule
    Llama2-7B & 26.83 & 21.85 & 34.64 & 30.55 & 20.75 & 26.68 & 25.23 & 39.56 & 11.58 & 30.26 & 26.49 \\
    Mistral-7b-Instruct-v0.2 & 27.42 & 27.38 & 30.72 & 27.64 & 23.81 & 27.57 & 28.21 & 28.57 & 11.58 & 27.63 & 26.87 \\
    GPT3.5 & 37.58 & 32.31 & 42.16 & 37.45 & 38.78 & 37.56 & 38.30 & 40.66 & 13.68 & 42.11 & 38.20 \\
    GPT4 & \high{51.92} & \high{41.54} & \high{52.29} & \high{50.91} & \high{63.95} & \high{45.75} & \high{54.59} & \high{60.44} & \high{23.16} & \high{53.29} & \high{61.23} \\
    \midrule
    \multicolumn{12}{l}{\hfill \textit{Large Multimodal Models (LMMs): Text-Only Input} } \\
    \midrule
    LLaVA-Next-Mistral-7B & 23.08 & 21.23 & 22.55 & 25.45 & 23.47 & 22.21 & 23.62 & 25.27 & 8.42 & 26.32 & 25.34 \\
    InternLM-XComposer2-VL & 33.17 & 24.62 & 44.12 & 32.36 & 31.97 & 30.40 & 33.03 & 46.15 & 10.53 & 41.45 & 34.17 \\
    Qwen-VL-Plus & 34.75 & 30.15 & 43.46 & 33.82 & 31.63 & 34.43 & 34.63 & 48.35 & \high{21.05} & 44.74 & 32.63 \\
    Gemini-Pro-Vision & 38.42 & 31.08 & \high{48.37} & 31.27 & 42.86 & 34.72 & 37.84 & \high{49.45} & 18.95 & \high{51.97} & 39.54 \\
    Claude-3-Sonnet & 38.58 & 31.38 & 43.46 & 39.27 & 40.82 & 36.66 & 40.14 & 46.15 & 14.74 & 43.42 & 42.23 \\
    GPT4V & \high{47.00} & \high{35.08} & 47.06 & \high{50.55} & \high{56.80} & \high{41.43} & \high{50.69} & 48.35 & 15.79 & 47.37 & \high{55.66} \\
    \midrule
    \multicolumn{12}{l}{\hfill \textit{Large Multimodal Models (LMMs): Multimodal Input}} \\
    \midrule
    LLaVA-Next-Mistral-7B & 24.58 & 22.77 & 24.18 & 27.64 & 24.15 & 23.55 & 24.54 & 29.67 & 9.47 & 25.00 & 25.91 \\
    InternLM-XComposer2-VL & 29.00 & 21.54 & 32.68 & 31.64 & 30.95 & 26.97 & 30.73 & 37.36 & 10.53 & 35.53 & 32.05 \\
    Qwen-VL-Plus & 32.00 & 28.62 & 35.95 & 33.45 & 30.27 & 32.34 & 33.49 & 32.97 & 21.05 & 42.11 & 32.05 \\
    Gemini-Pro-Vision & 38.33 & 28.92 & 48.69 & 32.73 & 43.20 & 33.68 & 38.07 & 50.55 & 14.74 & 53.95 & 39.73 \\
    Claude-3-Sonnet & 37.08 & 27.69 & 41.50 & 39.27 & 40.82 & 33.38 & 40.60 & 46.15 & 14.74 & 41.45 & 42.42 \\
    GPT4V & 45.33 & 34.46 & 42.16 & 49.45 & 56.80 & 39.64 & 50.00 & 41.76 & 13.68 & 46.71 & 55.28 \\
    VL-Cogito & 49.17  & 40.31 & 53.92 & 53.74 & 49.45  & 45.31 & 53.85 & 52.40 & 55.26 & 50.23 & 20.00  \\
    Qwen2.5-VL-72B & 52.25  & 42.77 & 50.00 & 61.22 & 56.36  & 45.01 & 50.55 & 62.38 & 53.95 & 58.49 & 23.16  \\
    GPT4.1 & 62.42  & 54.77 & 58.50 & 72.79 & 64.73  & 56.93 & 72.53 & 70.25 & 56.58 & 66.51 & 54.74  \\
    InternVL3.5-38B & 63.92  & 57.85 & 61.11 & 73.47 & 64.00  & 56.33 & 72.53 & 71.21 & 55.92 & 67.20 & 54.74  \\
    O4-Mini & 73.00  & 68.92 & 76.47 & \best{74.83} & 72.00  & 69.75 & 87.91 & 74.09 & 73.03 & 71.10 & 56.84  \\
    Doubao-Seed-1.6 & \best{77.33}  & \best{75.38} & \best{81.37} & 74.49 & \best{78.18}  & \best{75.26} & \best{90.11} & \best{76.97} & \best{76.32} & \best{75.92} & \best{68.42}  \\
    \bottomrule
    \end{tabular}
    }
    \vspace{-0.4cm}
% \vspace{-3mm}
\label{Table: 2AidAnsw}
\end{table*}

\begin{table*}[t!]
% % \vspace{-3mm}
\caption{Comprehensive performance evaluation of different models on three reasoning tasks. We assess multiple models on \textit{General Reasoning}, \textit{Direct Visual-aided Reasoning}, and \textit{Indirect Visual-aided Reasoning}. The metrics include \textbf{ACCU} (Accuracy, in \%) and our proposed \textbf{PVA}, and \textbf{SPRS}. SPRS is composed of four sub-dimensions, each rated on a 0-10 scale: \textbf{LogiR} (Logical Rigor), \textbf{ToolF} (Tool-Free Feasibility), \textbf{InfoT} (Information Traceability), and \textbf{GeneR} (Generality of the Method). A key finding is the significant gap between accuracy (ACCU) and process robustness (SPRS) across all models, underscoring that a correct answer does not guarantee a sound and verifiable reasoning process. For each task, the best score for each metric is highlighted in \textbf{green}.}
\small
\centering
 \renewcommand\tabcolsep{9pt} % column space
 \resizebox{0.95\linewidth}{!}{
    \begin{tabular}{l|c|c|c|cccc}
    \toprule
    \header{Model} & \header{ACCU} & \header{PVA} & \header{SPRS} & \header{LogiR} & \header{ToolF} & \header{InfoT} & \header{GeneR}  \\ 
    \midrule
    \multicolumn{7}{l}{\hfill \textit{Task 1: General Reasoning}} \\
    \midrule
    VL-Cogito & 48.00 & 24.00 & 11.50 & 5.86 & 8.81 & 8.83 & 7.80 \\
    Qwen2.5-VL-72B & 53.17 & 37.39 & 24.50 & 7.42 & 9.22 & 9.39 & 8.73 \\
    InternVL3.5-38B & 62.00 & 50.59 & 41.42 & 8.33 & 9.33 & 9.77 & 9.25 \\
    GPT-4.1 & 59.42 & 49.03 & 39.58 & 8.55 & 9.43 & 9.72 & 9.22 \\
    O4-Mini & 71.92 & 63.34 & 55.00 & 9.26 & 9.68 & 9.94 & 9.66 \\
    Doubao-Seed-1.6 & 78.75 & 68.13 & 55.25 & 9.09 & 9.72 & 9.93 & 9.61 \\

    \midrule
    \multicolumn{7}{l}{\hfill \textit{Task 2: Direct Visual-aided Reasoning}} \\
    \midrule
    VL-Cogito & 49.17 & 20.20 & 8.58 & 5.20 & 8.50 & 8.31 & 6.94 \\
    Qwen2.5-VL-72B & 52.25 & 34.79 & 21.42 & 7.13 & 9.18 & 9.27 & 8.54 \\
    InternVL3.5-38B & 63.92 & 20.20 & 30.67 & 7.93 & 9.23 & 9.68 & 9.19 \\
    GPT-4.1 & 62.42 & 52.34 & 44.17 & 8.72 & 9.44 & 9.78 & 9.38 \\
    O4-Mini & 73.00 & 53.84 & 37.92 & 8.29 & 9.53 & 9.44 & 9.23 \\
    Doubao-Seed-1.6 & 77.33 & 62.03 & 47.58 & 8.63 & 9.64 & 9.72 & 9.40 \\
    \midrule
    \multicolumn{7}{l}{\hfill \textit{Task 3: Indirect Visual-aided Reasoning}} \\
    \midrule
    VL-Cogito & 48.67 & 25.50 & 12.67 & 6.02 & 8.82 & 8.84 & 7.89 \\
    Qwen2.5-VL-72B & 54.67 & 37.99 & 25.92 & 7.36 & 9.22 & 9.44 & 8.67 \\
    InternVL3.5-38B & 60.00 & 47.72 & 38.67 & 8.24 & 9.39 & 9.76 & 9.26 \\
    GPT-4.1 & 60.50 & 51.38 & 44.67 & 8.72 & 9.46 & 9.75 & 9.32 \\
    O4-Mini & 73.50 & 65.02 & \best{62.08} & \best{9.31} & 9.74 & 9.94 & \best{9.65} \\
    Doubao-Seed-1.6 & \best{81.00} & \best{69.84} & 59.25 & 9.06 & \best{9.76} & \best{9.90} & 9.56 \\
    \bottomrule
    \end{tabular}
    }
    % \vspace{-2mm}
% \vspace{-1mm}
\label{Table: Three Layer Evaluate}
% \vspace{-0.4cm}
\end{table*}

% \vspace{-0.5cm}
\section{Experiments}
\subsection{Models}
To comprehensively evaluate the challenges posed by VisAidMath, we selected a wide spectrum of models, with a particular focus on large multi-modal models designed for complex reasoning. Our selection encompasses both leading open-source and proprietary systems. Our evaluation includes:
1) Open-source LLMs: Llama-2-7B \cite{llama2}, Mistral-7B-Instruct-v0.2 \cite{mistral};
2) Closed-source LLMs: GPT-3.5-turbo \cite{gpt35}, GPT-4-turbo \cite{gpt4};
3) Open-source LMMs: LLaVA-Next-Mistral-7B \cite{liu2024llava}, InternLM-XComposer2-VL \cite{internlm-xcomposer2}, VL-Cogito\cite{yuan2025vl}, Qwen2.5-VL-72B\cite{bai2025qwen2}, InternVL3.5-38B\cite{wang2025internvl3};
4) Closed-source LMMs: Qwen-VL-Plus \cite{bai2023qwen}, Gemini-Pro-Vision \cite{geminiprov}, Claude-3-Sonnet \cite{anthropic2024claude}, GPT-4-Vision \cite{gpt4v}, GPT-4.1\cite{gpt4.1}, O4-Mini\cite{o4mini}, and Doubao-Seed-1.6\cite{guo2025seed1}. 
This extensive selection allows for a robust analysis of performance trends across different model architectures, scales, and training paradigms. Detailed experimental settings and hyperparameters are provided in Appendix~\ref{append: hyperparameters}.

\subsection{Three-Layered Funnel Evaluation}
To move beyond surface-level correctness and enable a deeper assessment of mathematical reasoning, we introduce \textbf{The Three-Layered Funnel Evaluation Framework}. This hierarchical methodology is designed to scrutinize model outputs with increasing depth and rigor, moving from surface-level answer checking to fine-grained component analysis, and culminating in a holistic judgment of the entire reasoning process. This approach allows us to distinguish not only correct from incorrect answers, but also robustly derived solutions from those that are superficially correct but procedurally flawed.
\paragraph{Final Answer Correctness (ACCU)}
Given that VisAidMath comprises mathematics problems with deterministic answers, we evaluate the correctness of the final answer using Standard Accuracy (ACCU) as a baseline metric.
To accurately extract the final answer from model outputs, we follow the approach of \cite{lu2023mathvista} and employ GPT-4o mini as answer extractor, as it demonstrated a success rate of 99\% in preliminary experiments with 200 examples. The prompts used to construct the answer extractor are described in detail in section \ref{Appendix: Answer Extractor}.
ACCU is calculated as:
\begin{equation} \label{eq:accu}
    \text{ACCU} = \frac{1}{N} \sum_{j=1}^{N} a_j
\end{equation}
where $N$ is the total number of samples and $a_j \in \{0, 1\}$ is the binary correctness of the final answer for sample $j$.

\paragraph{Process-Verified Accuracy (PVA)}
Recognizing that standard accuracy (ACCU) cannot distinguish valid reasoning from "lucky guesses," we introduce PVA to act as a strict reliability filter. PVA refines the initial accuracy score by considering a solution valid only if its final answer is correct \textit{and} its reasoning process is judged to be holistically sound.

Instead of relying on a rigid, rule-based threshold, we leverage the evaluator model's own meta-reasoning capabilities. For each solution, we employ a two-step assessment process. First, we prompt the model to provide granular scores across several evaluation dimensions ($\mathcal{D} = {\text{LogiR, ToolF, InfoT, GeneR}}$). Second, based on its own fine-grained analysis, we prompt the model to make a holistic, binary judgment on the overall validity of the reasoning process. This yields a binary validity score, $V_j$, for each sample $j$.

Formally, PVA is defined as:
\begin{equation} \label{eq:pva}
\text{PVA} = \frac{1}{N} \sum_{j=1}^{N} a_j \cdot V_j
\end{equation}
where $a_j$ is 1 if the final answer of sample $j$ is correct and 0 otherwise, and $V_j$ is the binary validity judgment (1 for valid, 0 for invalid) directly provided by the evaluator model. This approach allows the assessment to capture nuanced flaws that a simple threshold might miss, treating the LLM not just as a scorer, but as a qualitative judge of the entire reasoning chain.

\paragraph{Solution Process Robustness Score (SPRS)}
The framework culminates in the Solution Process Robustness Score (SPRS), which offers the most granular level of assessment. Distinct from the binary filtering of PVA, SPRS provides a continuous, fine-grained quality score for all correctly answered problems. It quantifies the overall robustness of a solution by multiplicatively aggregating its scores across all process dimensions. This mechanism, reflecting a ``short-plank effect,'' heavily penalizes any ``weak links'' in the reasoning chain. It is calculated as:
\begin{equation} \label{eq:sprs}
    \text{SPRS} = \frac{1}{N} \sum_{j=1}^{N} a_j \cdot \prod_{i \in \mathcal{D}} \left( \frac{S_{i,j}}{10} \right)
\end{equation}
The gating by $a_j$ ensures that only correctly answered samples contribute to the score. The product $\prod$ over the normalized dimension scores ($S_{i,j}/10$) ensures that even a single low-quality dimension significantly depresses the overall score for that sample, thus rewarding solutions that are consistently strong across all aspects of reasoning.

\subsection{Main Results}\label{Experiment: Result}
\paragraph{Initial Performance via Standard Accuracy}
The complete performance results are presented in Tables \ref{Table: 2Answ}, \ref{Table: 2AidAnsw}, and \ref{Table: Aid2Answ}, with a summary of leading models in Table \ref{Table: Three Layer Evaluate}. On the metric of standard accuracy (ACCU), top-performing models demonstrate high scores. For instance, Doubao-Seed-1.6 achieves a peak ACCU of \textbf{81.00\%} on the Indirect Visual-aided Reasoning task and \textbf{77.33\%} on the Direct Visual-aided Reasoning task. While these figures indicate a strong capability for arriving at correct final answers, they alone do not validate the underlying reasoning processes used to obtain them.

\paragraph{Performance Collapse under Deeper Scrutiny}
The insufficiency of ACCU as a standalone metric becomes evident when our Funnel Evaluation Framework is applied. As shown in Table \ref{Table: Three Layer Evaluate}, a significant and universal decline is observed from ACCU to both PVA and SPRS. On the Direct Visual-aided Reasoning task, for example, Doubao-Seed-1.6’s 77.33\% ACCU declines to a PVA of 62.03\% and an SPRS of 47.58\%. We define this stark discrepancy between apparent correctness and procedural soundness as the ``\textbf{reasoning illusion}''. This phenomenon provides powerful evidence that VisAidMath effectively probes deep reasoning deficiencies that are masked by conventional accuracy-only evaluations. A quantitative analysis of this illusion and its implications for the VisAidMath benchmark is the central focus of Section~\ref{sec:analysis}.

\section{Analysis}\label{sec:analysis}

Following the discovery of the universal ``reasoning illusion'', this chapter provides a deep analysis of this phenomenon. We first offer quantitative proof demonstrating how VisAidMath uniquely exposes model deficiencies, and then present a qualitative diagnosis to understand the root causes behind these failures.

\subsection{The Reasoning Gap Quantified}
\label{sec:quantitative_analysis}

\begin{figure}[t!]
  \centering
  \begin{subfigure}[t]{0.44\linewidth}
    \centering
    \includegraphics[width=\linewidth]{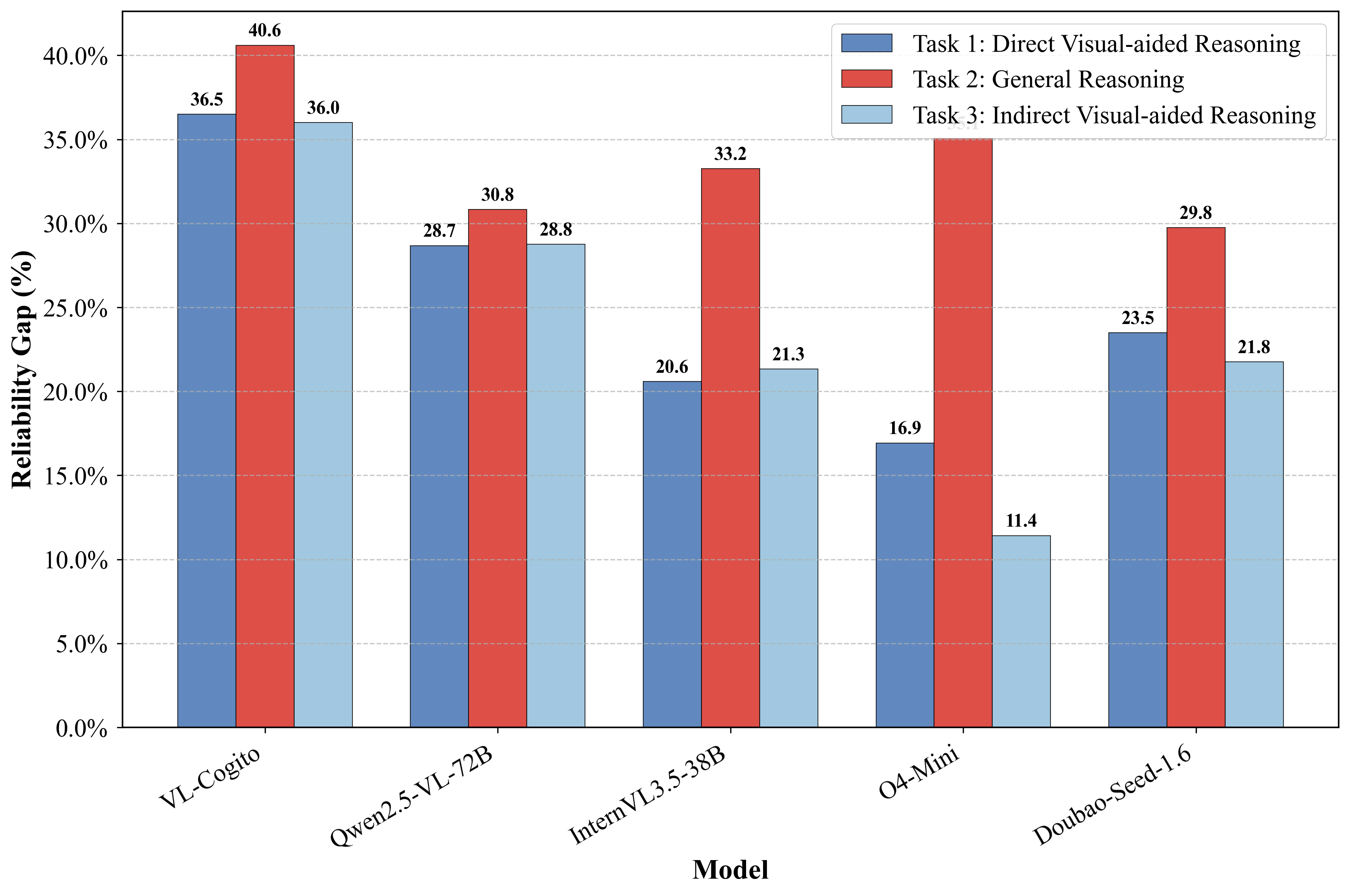}
    \caption{The \textbf{Reliability Gap} (``ACCU - PVA'') across the three reasoning tasks.}
    \label{Figure: delta graph pva}
  \end{subfigure}%
  \hspace{0.02\linewidth}%
  \begin{subfigure}[t]{0.44\linewidth}
    \centering
    \includegraphics[width=\linewidth]{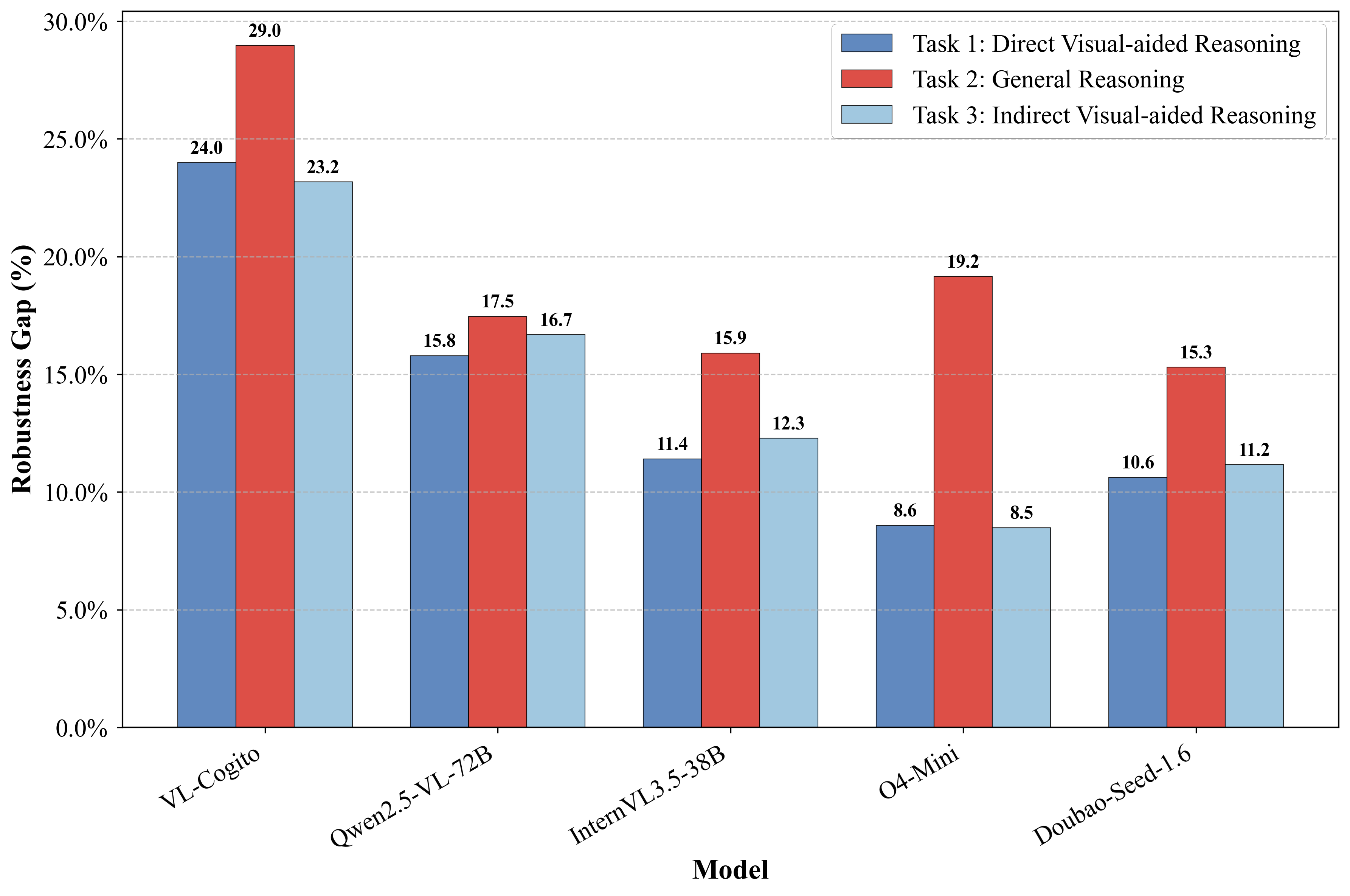}
    \caption{The \textbf{Robustness Gap} (``ACCU - SPRS'') across the three reasoning tasks.}
    \label{Figure: delta graph sprs}
  \end{subfigure}
  \caption{Performance degradation from surface accuracy (ACCU) to process-level evaluation. The Reliability Gap (a) measures the proportion of correct answers with procedurally invalid reasoning. The Robustness Gap (b) measures the total drop in solution quality. Both gaps are most pronounced in the Direct Visual-aided Reasoning (D-VAR) task, highlighting its unique challenge.}
  \label{fig:delta_graphs}
  \vspace{-0.3cm}
\end{figure}

\paragraph{Reliability and Robustness Gaps}
To objectively measure the performance drop from surface accuracy to procedural quality, we define two metrics: the \textbf{Reliability Gap} (``ACCU - PVA'') and the \textbf{Robustness Gap} (``ACCU - SPRS''). The former quantifies the proportion of correct answers derived from flawed processes, while the latter measures the overall decline in solution quality. Figure \ref{fig:delta_graphs} visualizes these two gaps across our three reasoning tasks.

\begin{figure}[htbp!]
    \centering
    \begin{subfigure}[t]{0.48\linewidth}
        \centering
        \includegraphics[width=\linewidth]{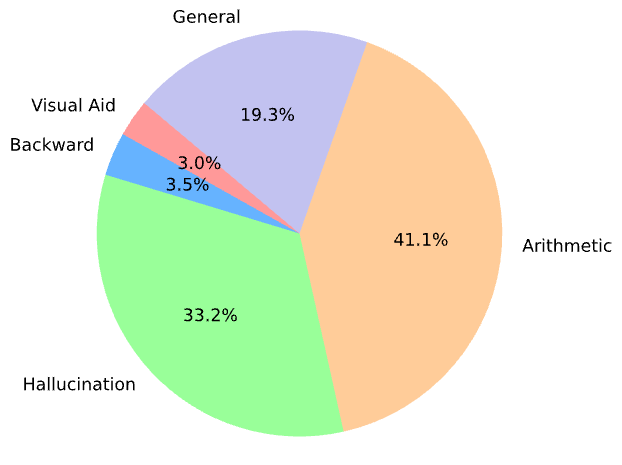}
        \caption{Distribution of reasoning strategies for correctly solved problems. The minimal use of "Visual-Aided" reasoning (3.0\%) reveals a strong model tendency to evade the intended solution path.}
        \label{Figure: Reasoning Tendency}
    \end{subfigure}
    \hfill  % 自动填充间距
    \begin{subfigure}[t]{0.48\linewidth}
        \centering
        \includegraphics[width=\linewidth]{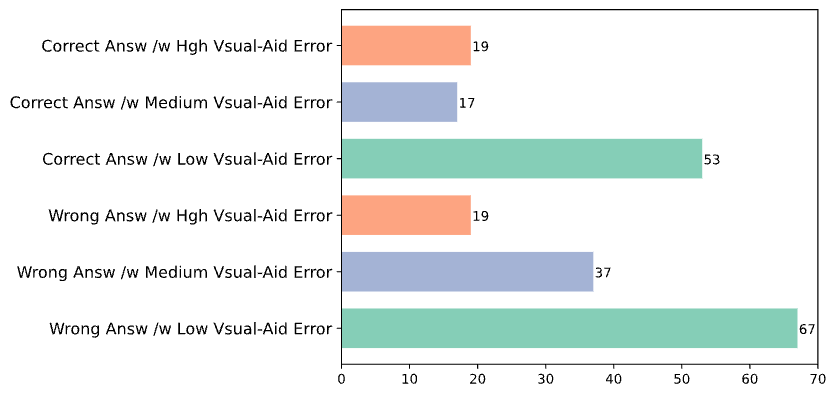}
        \caption{Correlation between visual-aid error severity and final answer correctness. High-severity errors are strongly predictive of an incorrect final answer.}
        \label{Figure: Correlation Analysis of Visual-Aid Error Rate and Answer Correctness}
    \end{subfigure}
    \caption{Qualitative diagnosis of the reasoning gap.}
    \label{fig:qualitativeAnalysis}
    \vspace{-0.3cm}
\end{figure}

The analysis reveals a critical finding: the \textbf{Direct Visual-aided Reasoning (D-VAR) task induces a dramatically larger gap} in both reliability and robustness compared to the other tasks. This indicates that while models may seem competent on the surface, their reasoning processes are particularly fragile when required to directly engage with visual information. For instance, InternVL3.5-38B on the D-VAR task exhibits a massive 43.7-point Reliability Gap and a 33.2-point Robustness Gap. This provides strong quantitative evidence that the core task of VisAidMath is uniquely effective at pressuring models to reveal their underlying reasoning deficiencies.

\paragraph{Diagnosing Failure Patterns}
The interplay between PVA and SPRS also serves as a powerful diagnostic tool for identifying distinct model failure modes. The relationship between these two metrics is not uniform across models, revealing different behavioral patterns:
\begin{itemize}
    \item \textbf{Case A: Catastrophic Failures (SPRS > PVA).} For models like InternVL3.5-38B on the D-VAR task (PVA: 20.20\%, SPRS: 30.67\%), the SPRS score is notably higher than the PVA score. This pattern suggests the model frequently makes critical, "all-or-nothing" errors that cause its solutions to be entirely invalidated by the PVA filter.
    \item \textbf{Case B: Systemic Minor Flaws (PVA > SPRS).} In contrast, models like Doubao-Seed-1.6 (PVA: 62.03\%, SPRS: 47.58\%) exhibit a higher PVA than SPRS. This indicates the model is more adept at avoiding fatal errors, but its procedurally "valid" solutions are often rife with minor inaccuracies, which are penalized by the SPRS, dragging down the overall robustness score.
\end{itemize}
This ability to distinguish between models prone to catastrophic accidents and those exhibiting systemic sloppiness underscores the diagnostic depth of our framework.

\subsection{Qualitative Diagnosis}
\paragraph{Evasion of Visual Reasoning}\label{Reasoning Tendency}
A primary cause for the reasoning gap is that models strongly tend to \textit{evade} the intended visual-aided path. Our manual analysis of 200 correctly answered D-VAR samples shows that a staggering majority relied on non-visual shortcuts (Figure \ref{Figure: Reasoning Tendency}). 
Only 3.0\% of correct solutions were achieved by generating and using visual aids as intended. The dominant strategies were pure arithmetic (41.1\%) or general reasoning that ignored the visual context (19.3\%). This circumvention of the core task is a key factor explaining how high ACCU scores can mask poor underlying reasoning processes.

\paragraph{The High Cost of Flawed Attempts}
Furthermore, when models do attempt the visual reasoning path, errors in this intermediate step are often fatal to the final outcome. As demonstrated in Figure \ref{Figure: Correlation Analysis of Visual-Aid Error Rate and Answer Correctness}, there is a strong correlation between the severity of visual-aid errors and the correctness of the final answer. Solutions with significant visual-aid errors are overwhelmingly more likely to result in an incorrect final answer. This high cost of flawed attempts further explains the significant performance gaps observed in the D-VAR task, as engaging with the task incorrectly is highly detrimental. A more detailed breakdown of these error types can be found in Appendix \ref{Figure: Correlation Analysis of Visual-Aid Error Reason and Answer Correctness}.
\section{Conclusion}
In this paper, we lay the groundwork for mathematical problem solving using multi-modal reasoning steps. We introduce VisAidMath, a benchmark designed to investigate the visual-aided reasoning capabilities of both large language models and large multi-modal models. Experiments on mainstream models demonstrate deficiencies in deducing visual aids and the corresponding textual reasoning steps.
We also conducted fine-grained quantitative and qualitative analyses to reveal disparities in visual-aid reasoning.This exposes a divide between passive perception and active, visually grounded deduction in current LMMs. VisAidMath thus stands as both a challenging benchmark and a guiding paradigm for advancing visual grounded reasoning.

\bibliographystyle{unsrt}
\bibliography{custom}

@inproceedings{wu2020knowledge,
  title={A knowledge-aware sequence-to-tree network for math word problem solving},
  author={Wu, Qinzhuo and Zhang, Qi and Fu, Jinlan and Huang, Xuan-Jing},
  booktitle={Proceedings of the 2020 conference on empirical methods in natural language processing (EMNLP)},
  pages={7137--7146},
  year={2020}
}

@article{lewkowycz2022solving,
  title={Solving quantitative reasoning problems with language models},
  author={Lewkowycz, Aitor and Andreassen, Anders and Dohan, David and Dyer, Ethan and Michalewski, Henryk and Ramasesh, Vinay and Slone, Ambrose and Anil, Cem and Schlag, Imanol and Gutman-Solo, Theo and others},
  journal={Advances in Neural Information Processing Systems},
  volume={35},
  pages={3843--3857},
  year={2022}
}

@article{hendrycks2021measuring,
  title={Measuring mathematical problem solving with the math dataset},
  author={Hendrycks, Dan and Burns, Collin and Kadavath, Saurav and Arora, Akul and Basart, Steven and Tang, Eric and Song, Dawn and Steinhardt, Jacob},
  journal={arXiv preprint arXiv:2103.03874},
  year={2021}
}

@inproceedings{wu2021lime,
  title={Lime: Learning inductive bias for primitives of mathematical reasoning},
  author={Wu, Yuhuai and Rabe, Markus N and Li, Wenda and Ba, Jimmy and Grosse, Roger B and Szegedy, Christian},
  booktitle={International Conference on Machine Learning},
  pages={11251--11262},
  year={2021},
  organization={PMLR}
}

@article{lu2023mathvista,
  title={Mathvista: Evaluating mathematical reasoning of foundation models in visual contexts},
  author={Lu, Pan and Bansal, Hritik and Xia, Tony and Liu, Jiacheng and Li, Chunyuan and Hajishirzi, Hannaneh and Cheng, Hao and Chang, Kai-Wei and Galley, Michel and Gao, Jianfeng},
  journal={arXiv preprint arXiv:2310.02255},
  year={2023}
}

@article{lindstrom2022clevr,
  title={Clevr-math: A dataset for compositional language, visual and mathematical reasoning},
  author={Lindstr{\"o}m, Adam Dahlgren and Abraham, Savitha Sam},
  journal={arXiv preprint arXiv:2208.05358},
  year={2022}
}

@article{masry2022chartqa,
  title={Chartqa: A benchmark for question answering about charts with visual and logical reasoning},
  author={Masry, Ahmed and Long, Do Xuan and Tan, Jia Qing and Joty, Shafiq and Hoque, Enamul},
  journal={arXiv preprint arXiv:2203.10244},
  year={2022}
}

@article{rae2021scaling,
  title={Scaling language models: Methods, analysis \& insights from training gopher},
  author={Rae, Jack W and Borgeaud, Sebastian and Cai, Trevor and Millican, Katie and Hoffmann, Jordan and Song, Francis and Aslanides, John and Henderson, Sarah and Ring, Roman and Young, Susannah and others},
  journal={arXiv preprint arXiv:2112.11446},
  year={2021}
}

@inproceedings{gao2023pal,
  title={Pal: Program-aided language models},
  author={Gao, Luyu and Madaan, Aman and Zhou, Shuyan and Alon, Uri and Liu, Pengfei and Yang, Yiming and Callan, Jamie and Neubig, Graham},
  booktitle={International Conference on Machine Learning},
  pages={10764--10799},
  year={2023},
  organization={PMLR}
}

@article{llama2,
  title={Llama 2: Open foundation and fine-tuned chat models},
  author={Touvron, Hugo and Martin, Louis and Stone, Kevin and Albert, Peter and Almahairi, Amjad and Babaei, Yasmine and Bashlykov, Nikolay and Batra, Soumya and Bhargava, Prajjwal and Bhosale, Shruti and others},
  journal={arXiv preprint arXiv:2307.09288},
  year={2023}
}

@inproceedings{mistral,
  title={Mistral: Dynamically managing power, performance, and adaptation cost in cloud infrastructures},
  author={Jung, Gueyoung and Hiltunen, Matti A and Joshi, Kaustubh R and Schlichting, Richard D and Pu, Calton},
  booktitle={2010 IEEE 30th International Conference on Distributed Computing Systems},
  pages={62--73},
  year={2010},
  organization={IEEE}
}

@inproceedings{gpt35,
  year={2023},
  title={https://platform.openai.com/docs/models/gpt-3-5-turbo},
  organization={OpenAI},
  author={OpenAI},
  booktitle={OpenAI}
}

@inproceedings{gpt4,
  year={2023},
  title={https://platform.openai.com/docs/models/gpt-4},
  organization={OpenAI},
  author={OpenAI},
  booktitle={OpenAI}
}

@inproceedings{gpt4v,
  year={2023},
  title={https://openai.com/index/gpt-4v-system-card/},
  organization={OpenAI},
  author={OpenAI},
  booktitle={OpenAI}
}

@inproceedings{gpt4.1,
  year={2023},
  title={https://platform.openai.com/docs/models/gpt-4.1},
  organization={OpenAI},
  author={OpenAI},
  booktitle={OpenAI}
}

@inproceedings{o4mini,
  year={2023},
  title={https://platform.openai.com/docs/models/o4-mini},
  organization={OpenAI},
  author={OpenAI},
  booktitle={OpenAI}
}

@article{guo2025seed1,
  title={Seed1. 5-vl technical report},
  author={Guo, Dong and Wu, Faming and Zhu, Feida and Leng, Fuxing and Shi, Guang and Chen, Haobin and Fan, Haoqi and Wang, Jian and Jiang, Jianyu and Wang, Jiawei and others},
  journal={arXiv preprint arXiv:2505.07062},
  year={2025}
}

@misc{liu2024llava,
  title={Llava-next: Improved reasoning, ocr, and world knowledge},
  author={Liu, Haotian and Li, Chunyuan and Li, Yuheng and Li, Bo and Zhang, Yuanhan and Shen, Sheng and Lee, Yong Jae},
  year={2024}
}

@article{internlm-xcomposer2,
  title={InternLM-XComposer2: Mastering free-form text-image composition and comprehension in vision-language large model},
  author={Dong, Xiaoyi and Zhang, Pan and Zang, Yuhang and Cao, Yuhang and Wang, Bin and Ouyang, Linke and Wei, Xilin and Zhang, Songyang and Duan, Haodong and Cao, Maosong and others},
  journal={arXiv preprint arXiv:2401.16420},
  year={2024}
}

@article{bai2023qwen,
  title={Qwen-vl: A versatile vision-language model for understanding, localization, text reading, and beyond},
  author={Bai, Jinze and Bai, Shuai and Yang, Shusheng and Wang, Shijie and Tan, Sinan and Wang, Peng and Lin, Junyang and Zhou, Chang and Zhou, Jingren},
  journal={arXiv preprint arXiv:2308.12966},
  volume={1},
  number={2},
  pages={3},
  year={2023}
}

@inproceedings{geminiprov,
  year={2023},
  title={https://labelbox.com/product/model/foundry-models/google-gemini-pro-vision/},
  organization={Google},
  author={Google},
  booktitle={Google}
}

@article{anthropic2024claude,
  title={The claude 3 model family: Opus, sonnet, haiku},
  author={Anthropic, AI},
  journal={Claude-3 Model Card},
  year={2024}
}

@article{yuan2023well,
  title={How well do Large Language Models perform in Arithmetic tasks?},
  author={Yuan, Zheng and Yuan, Hongyi and Tan, Chuanqi and Wang, Wei and Huang, Songfang},
  journal={arXiv preprint arXiv:2304.02015},
  year={2023}
}

@article{yue2023mammoth,
  title={Mammoth: Building math generalist models through hybrid instruction tuning},
  author={Yue, Xiang and Qu, Xingwei and Zhang, Ge and Fu, Yao and Huang, Wenhao and Sun, Huan and Su, Yu and Chen, Wenhu},
  journal={arXiv preprint arXiv:2309.05653},
  year={2023}
}

@article{chen2022unigeo,
  title={Unigeo: Unifying geometry logical reasoning via reformulating mathematical expression},
  author={Chen, Jiaqi and Li, Tong and Qin, Jinghui and Lu, Pan and Lin, Liang and Chen, Chongyu and Liang, Xiaodan},
  journal={arXiv preprint arXiv:2212.02746},
  year={2022}
}

@article{lu2021inter,
  title={Inter-GPS: Interpretable geometry problem solving with formal language and symbolic reasoning},
  author={Lu, Pan and Gong, Ran and Jiang, Shibiao and Qiu, Liang and Huang, Siyuan and Liang, Xiaodan and Zhu, Song-Chun},
  journal={arXiv preprint arXiv:2105.04165},
  year={2021}
}

@article{kojima2022large,
  title={Large language models are zero-shot reasoners},
  author={Kojima, Takeshi and Gu, Shixiang Shane and Reid, Machel and Matsuo, Yutaka and Iwasawa, Yusuke},
  journal={Advances in neural information processing systems},
  volume={35},
  pages={22199--22213},
  year={2022}
}

@article{liu2024visual,
  title={Visual instruction tuning},
  author={Liu, Haotian and Li, Chunyuan and Wu, Qingyang and Lee, Yong Jae},
  journal={Advances in neural information processing systems},
  volume={36},
  year={2024}
}

@inproceedings{Wolf_Transformers_State-of-the-Art_Natural_2020,
    title = "Transformers: State-of-the-Art Natural Language Processing",
    author = "Wolf, Thomas  and
      Debut, Lysandre  and
      Sanh, Victor  and
      Chaumond, Julien  and
      Delangue, Clement  and
      Moi, Anthony  and
      Cistac, Pierric  and
      Rault, Tim  and
      Louf, Remi  and
      Funtowicz, Morgan  and
      Davison, Joe  and
      Shleifer, Sam  and
      von Platen, Patrick  and
      Ma, Clara  and
      Jernite, Yacine  and
      Plu, Julien  and
      Xu, Canwen  and
      Le Scao, Teven  and
      Gugger, Sylvain  and
      Drame, Mariama  and
      Lhoest, Quentin  and
      Rush, Alexander",
    editor = "Liu, Qun  and
      Schlangen, David",
    booktitle = "Proceedings of the 2020 Conference on Empirical Methods in Natural Language Processing: System Demonstrations",
    month = oct,
    year = "2020",
    address = "Online",
    publisher = "Association for Computational Linguistics",
    url = "https://aclanthology.org/2020.emnlp-demos.6",
    doi = "10.18653/v1/2020.emnlp-demos.6",
    pages = "38--45",
}

@inproceedings{image_concat,
  year={2023},
  title={https://github.com/larryflynt/image-concat},
  author={larryflynt},
  booktitle={Github}
}

@article{bai2025qwen2,
  title={Qwen2. 5-vl technical report},
  author={Bai, Shuai and Chen, Keqin and Liu, Xuejing and Wang, Jialin and Ge, Wenbin and Song, Sibo and Dang, Kai and Wang, Peng and Wang, Shijie and Tang, Jun and others},
  journal={arXiv preprint arXiv:2502.13923},
  year={2025}
}

@article{wang2025internvl3,
  title={Internvl3. 5: Advancing open-source multimodal models in versatility, reasoning, and efficiency},
  author={Wang, Weiyun and Gao, Zhangwei and Gu, Lixin and Pu, Hengjun and Cui, Long and Wei, Xingguang and Liu, Zhaoyang and Jing, Linglin and Ye, Shenglong and Shao, Jie and others},
  journal={arXiv preprint arXiv:2508.18265},
  year={2025}
}

@article{yuan2025vl,
  title={Vl-cogito: Progressive curriculum reinforcement learning for advanced multimodal reasoning},
  author={Yuan, Ruifeng and Xiao, Chenghao and Leng, Sicong and Wang, Jianyu and Li, Long and Xu, Weiwen and Chan, Hou Pong and Zhao, Deli and Xu, Tingyang and Wei, Zhongyu and others},
  journal={arXiv preprint arXiv:2507.22607},
  year={2025}
}

@article{su2025thinking,
  title={Thinking with images for multimodal reasoning: Foundations, methods, and future frontiers},
  author={Su, Zhaochen and Xia, Peng and Guo, Hangyu and Liu, Zhenhua and Ma, Yan and Qu, Xiaoye and Liu, Jiaqi and Li, Yanshu and Zeng, Kaide and Yang, Zhengyuan and others},
  journal={arXiv preprint arXiv:2506.23918},
  year={2025}
}

@article{hu2022promptcap,
  title={Promptcap: Prompt-guided task-aware image captioning},
  author={Hu, Yushi and Hua, Hang and Yang, Zhengyuan and Shi, Weijia and Smith, Noah A and Luo, Jiebo},
  journal={arXiv preprint arXiv:2211.09699},
  year={2022}
}

@article{wu2024mind,
  title={Mind's eye of LLMs: visualization-of-thought elicits spatial reasoning in large language models},
  author={Wu, Wenshan and Mao, Shaoguang and Zhang, Yadong and Xia, Yan and Dong, Li and Cui, Lei and Wei, Furu},
  journal={Advances in Neural Information Processing Systems},
  volume={37},
  pages={90277--90317},
  year={2024}
}

@article{liu2025visual,
  title={Visual Abstract Thinking Empowers Multimodal Reasoning},
  author={Liu, Dairu and Wang, Ziyue and Ruan, Minyuan and Luo, Fuwen and Chen, Chi and Li, Peng and Liu, Yang},
  journal={arXiv preprint arXiv:2505.20164},
  year={2025}
}

@article{qi2024cogcom,
  title={Cogcom: A visual language model with chain-of-manipulations reasoning},
  author={Qi, Ji and Ding, Ming and Wang, Weihan and Bai, Yushi and Lv, Qingsong and Hong, Wenyi and Xu, Bin and Hou, Lei and Li, Juanzi and Dong, Yuxiao and others},
  journal={arXiv preprint arXiv:2402.04236},
  year={2024}
}

@inproceedings{gupta2023visual,
  title={Visual programming: Compositional visual reasoning without training},
  author={Gupta, Tanmay and Kembhavi, Aniruddha},
  booktitle={Proceedings of the IEEE/CVF conference on computer vision and pattern recognition},
  pages={14953--14962},
  year={2023}
}

@article{fu2025refocus,
  title={Refocus: Visual editing as a chain of thought for structured image understanding},
  author={Fu, Xingyu and Liu, Minqian and Yang, Zhengyuan and Corring, John and Lu, Yijuan and Yang, Jianwei and Roth, Dan and Florencio, Dinei and Zhang, Cha},
  journal={arXiv preprint arXiv:2501.05452},
  year={2025}
}

@article{chervonyi2025gold,
  title={Gold-medalist performance in solving olympiad geometry with alphageometry2},
  author={Chervonyi, Yuri and Trinh, Trieu H and Ol{\v{s}}{\'a}k, Miroslav and Yang, Xiaomeng and Nguyen, Hoang and Menegali, Marcelo and Jung, Junehyuk and Verma, Vikas and Le, Quoc V and Luong, Thang},
  journal={arXiv preprint arXiv:2502.03544},
  year={2025}
}

@article{weng2025geosketch,
  title={GeoSketch: A Neural-Symbolic Approach to Geometric Multimodal Reasoning with Auxiliary Line Construction and Affine Transformation},
  author={Weng, Shichao and Wang, Zhiqiang and Zhou, Yuhua and Lu, Rui and Liu, Ting and Teng, Zhiyang and Liu, Xiaozhang and Liu, Hanmeng},
  journal={arXiv preprint arXiv:2509.22460},
  year={2025}
}

@article{team2024chameleon,
  title={Chameleon: Mixed-modal early-fusion foundation models},
  author={Team, Chameleon},
  journal={arXiv preprint arXiv:2405.09818},
  year={2024}
}

@article{guo2025can,
  title={Can We Generate Images with CoT? Let's Verify and Reinforce Image Generation Step by Step},
  author={Guo, Ziyu and Zhang, Renrui and Tong, Chengzhuo and Zhao, Zhizheng and Huang, Rui and Zhang, Haoquan and Zhang, Manyuan and Liu, Jiaming and Zhang, Shanghang and Gao, Peng and others},
  journal={arXiv preprint arXiv:2501.13926},
  year={2025}
}

@inproceedings{zhao2025cot,
  title={Cot-vla: Visual chain-of-thought reasoning for vision-language-action models},
  author={Zhao, Qingqing and Lu, Yao and Kim, Moo Jin and Fu, Zipeng and Zhang, Zhuoyang and Wu, Yecheng and Li, Zhaoshuo and Ma, Qianli and Han, Song and Finn, Chelsea and others},
  booktitle={Proceedings of the Computer Vision and Pattern Recognition Conference},
  pages={1702--1713},
  year={2025}
}

@article{li2025visiomath,
  title={Visiomath: Benchmarking figure-based mathematical reasoning in lmms},
  author={Li, Can and Liu, Ying and Zhang, Ting and Wang, Mei and Huang, Hua},
  journal={arXiv preprint arXiv:2506.06727},
  year={2025}
}

@article{park2025explain,
  title={Explain with Visual Keypoints Like a Real Mentor! A Benchmark for Multimodal Solution Explanation},
  author={Park, Jaewoo and Park, Jungyang and Jang, Dongju and Chung, Jiwan and Yoo, Byungwoo and Shin, Jaewoo and Park, Seonjoon and Kim, Taehyeong and Yu, Youngjae},
  journal={arXiv preprint arXiv:2504.03197},
  year={2025}
}

@article{fu2025trustgeogen,
  title={Trustgeogen: Scalable and formal-verified data engine for trustworthy multi-modal geometric problem solving},
  author={Fu, Daocheng and Chen, Zijun and Xia, Renqiu and Liu, Qi and Feng, Yuan and Zhou, Hongbin and Zhang, Renrui and Feng, Shiyang and Gao, Peng and Yan, Junchi and others},
  journal={arXiv preprint arXiv:2504.15780},
  year={2025}
}

%%%%%%%%%%%%%%%%%%%%%%%%%%%%%%%%%%%%%%%%%%%%%%%%%%%%%%%%%%%%

\appendix

%%%%%%%%%%%%%%%%%%%%%%%%%%%%%%%%%%%%%%%%%%%%%%%%%%%%%%%%%%%%

\newpage
\section*{NeurIPS Paper Checklist}

\begin{enumerate}

\item {\bf Claims}
    \item[] Question: Do the main claims made in the abstract and introduction accurately reflect the paper's contributions and scope?
    \item[] Answer: \answerYes{} % Replace by \answerYes{}, \answerNo{}, or \answerNA{}.
    \item[] Justification: The main claims has been made in the abstract and introduction accurately.

\item {\bf Limitations}
    \item[] Question: Does the paper discuss the limitations of the work performed by the authors?
    \item[] Answer: \answerYes{} % Replace by \answerYes{}, \answerNo{}, or \answerNA{}.
    \item[] Justification: The limitation is discueseed in appendix \ref{limitation and social impact}.

\item {\bf Theory assumptions and proofs}
    \item[] Question: For each theoretical result, does the paper provide the full set of assumptions and a complete (and correct) proof?
    \item[] Answer: \answerNA{} % Replace by \answerYes{}, \answerNo{}, or \answerNA{}.
    \item[] Justification: There is no theoretical result provided in this work.

    \item {\bf Experimental result reproducibility}
    \item[] Question: Does the paper fully disclose all the information needed to reproduce the main experimental results of the paper to the extent that it affects the main claims and/or conclusions of the paper (regardless of whether the code and data are provided or not)?
    \item[] Answer: \answerYes{} % Replace by \answerYes{}, \answerNo{}, or \answerNA{}.
    \item[] Justification: Detailed information for reproducing the main experimental results is provided in section \ref{sec:Dataset}, and appendix \ref{Appendix: metadata} and \ref{dataset collection detail}.

\item {\bf Open access to data and code}
    \item[] Question: Does the paper provide open access to the data and code, with sufficient instructions to faithfully reproduce the main experimental results, as described in supplemental material?
    \item[] Answer: \answerYes{} % Replace by \answerYes{}, \answerNo{}, or \answerNA{}.
    \item[] Justification: Our data will be hosted on an evaluation platform at CodaBench for testing publicly.

\item {\bf Experimental setting/details}
    \item[] Question: Does the paper specify all the training and test details (e.g., data splits, hyperparameters, how they were chosen, type of optimizer, etc.) necessary to understand the results?
    \item[] Answer: \answerYes{} % Replace by \answerYes{}, \answerNo{}, or \answerNA{}.
    \item[] Justification: Training and test details are provided in appendix \ref{append: detail expr setting}.

\item {\bf Experiment statistical significance}
    \item[] Question: Does the paper report error bars suitably and correctly defined or other appropriate information about the statistical significance of the experiments?
    \item[] Answer: \answerNo{} % Replace by \answerYes{}, \answerNo{}, or \answerNA{}.
    \item[] Justification: The experiments were each conducted only a single time due to expenses limit. As a result, it was not possible to calculate error bars or determine the statistical significance of the results.

\item {\bf Experiments compute resources}
    \item[] Question: For each experiment, does the paper provide sufficient information on the computer resources (type of compute workers, memory, time of execution) needed to reproduce the experiments?
    \item[] Answer: \answerNo{} % Replace by \answerYes{}, \answerNo{}, or \answerNA{}.
    \item[] Justification: Information of computer resources used in this work is hard to record.
    
\item {\bf Code of ethics}
    \item[] Question: Does the research conducted in the paper conform, in every respect, with the NeurIPS Code of Ethics \url{https://neurips.cc/public/EthicsGuidelines}?
    \item[] Answer: \answerYes{} % Replace by \answerYes{}, \answerNo{}, or \answerNA{}.
    \item[] Justification: This work is conducted conform with the NeurIPS Code of Ethics.

\item {\bf Broader impacts}
    \item[] Question: Does the paper discuss both potential positive societal impacts and negative societal impacts of the work performed?
    \item[] Answer: \answerYes{} % Replace by \answerYes{}, \answerNo{}, or \answerNA{}.
    \item[] Justification: Social impact is discussed in appendix \ref{limitation and social impact}.
    
\item {\bf Safeguards}
    \item[] Question: Does the paper describe safeguards that have been put in place for responsible release of data or models that have a high risk for misuse (e.g., pretrained language models, image generators, or scraped datasets)?
    \item[] Answer: \answerNo{} % Replace by \answerYes{}, \answerNo{}, or \answerNA{}.
    \item[] Justification: 

\item {\bf Licenses for existing assets}
    \item[] Question: Are the creators or original owners of assets (e.g., code, data, models), used in the paper, properly credited and are the license and terms of use explicitly mentioned and properly respected?
    \item[] Answer: \answerYes{} % Replace by \answerYes{}, \answerNo{}, or \answerNA{}.
    \item[] Justification: This paper properly cited and credited the paper of code, data, and models used in this work.

\item {\bf New assets}
    \item[] Question: Are new assets introduced in the paper well documented and is the documentation provided alongside the assets?
    \item[] Answer: \answerYes{} % Replace by \answerYes{}, \answerNo{}, or \answerNA{}.
    \item[] Justification: Detail of data liscence is provided in appendix \ref{Appendix: Data Process - Formalization} 

\item {\bf Crowdsourcing and research with human subjects}
    \item[] Question: For crowdsourcing experiments and research with human subjects, does the paper include the full text of instructions given to participants and screenshots, if applicable, as well as details about compensation (if any)? 
    \item[] Answer: \answerYes{} % Replace by \answerYes{}, \answerNo{}, or \answerNA{}.
    \item[] Justification: The detail of compensation and screenshots are provided in appendix \ref{dataset collection detail}.

\item {\bf Institutional review board (IRB) approvals or equivalent for research with human subjects}
    \item[] Question: Does the paper describe potential risks incurred by study participants, whether such risks were disclosed to the subjects, and whether Institutional Review Board (IRB) approvals (or an equivalent approval/review based on the requirements of your country or institution) were obtained?
    \item[] Answer: \answerYes{} % Replace by \answerYes{}, \answerNo{}, or \answerNA{}.
    \item[] Justification: Refer to appendix \ref{annotation details}.

\item {\bf Declaration of LLM usage}
    \item[] Question: Does the paper describe the usage of LLMs if it is an important, original, or non-standard component of the core methods in this research? Note that if the LLM is used only for writing, editing, or formatting purposes and does not impact the core methodology, scientific rigorousness, or originality of the research, declaration is not required.
    %this research? 
    \item[] Answer: \answerYes{} % Replace by \answerYes{}, \answerNo{}, or \answerNA{}.
    \item[] Justification: Refer to appendix \ref{llm usage}.

\end{enumerate}

%%%%%%%%%%%%%%%%%%%%%%%%%%%%%%%%%%%%%%%%%%%%%%%%%%%%%%%%%%%%%%%%%%%

% Reset depth to add sections and subsections to ToC
\addtocontents{toc}{\protect\setcounter{tocdepth}{3}}

% Setting colorlinks=black just for the table of contents
\hypersetup{linkcolor=black}

\clearpage
\section*{Appendix}
\tableofcontents % Lists only the appendix sections and subsections

\clearpage
% \addtocontents{toc}{\protect\setcounter{tocdepth}{3}}

% Setting colorlinks=black just for the table of contents
\hypersetup{linkcolor=black}

% Examples for Different Categorization
\section{Future Direction}\label{Future Direction}
\paragraph{Spatial Capability}
Despite the predominant emphasis on the construction and fitting of extensive datasets, mainstream works are confined to inference tasks within textual dimension. LLM exhibits exceedingly poor performance in providing visual reasoning step, revealing deficiencies in spatial understanding, imagination, and more other aspects. To address mathematical reasoning with visual-aid inference, future investigation could be directed to specifically enhance LLM's adaptation to visual reasoning tasks, devise new methods for better integration of textual and visual reasoning, or design specific architectures for multimodal reasoning steps inference.
\paragraph{Mathematical Image Generation}
Preliminary experiments find mainstream models exhibit poor mathematical image generation performance, thus further captioning each mathematical images to explore visual reasoning step inference. Primary model deficiencies fall in: mathematical image caption comprehension, spatial relationships apprehension, lack of numerical precision, significant stylization discrepancies in the images, and more. Generate image drawing code can increase the overall drawing precision, while suffering from plenty of code errors. There lies a long research road in mathematical image generation before fully exploration of textual-visual interconnected inference.
\paragraph{Evaluation Metrics}
Reasoning non-uniqueness enhances evaluation complexity of visual aids generation. Different viewing angle, relative element size, and styles can alter perceptual features instead of semantic feature. Visual-aid can be captioned by multiple correct expressions with semantic remains stable. Therefore, future evaluation metrics research for visual-aid should be directed toward semantic-based method.
\section{Dataset Preparation}

\subsection{Machine Translation}
\label{Appendix: Data Process - Machine Translation}
Since most of our data and their corresponding captions were in Chinese, we first translated all text into English. Open-source machine-translation (MT) models often exhibit deficiencies in semantic fidelity and numerical accuracy within mathematical contexts. Through manual sampling and comparison, we identified Baidu Translate\footnote{\url{https://fanyi.baidu.com/}} and DeepL\footnote{\url{https://www.deepl.com/translator}} as high-quality services for our specific needs. Therefore, we employed both platforms, splitting each sample's translation between the two to mitigate potential data leakage.

\subsection{Data Processing}
\label{Appendix: Data Process - Data Process}
We further process the annotated data to match the model‐input format and the requirements of our evaluation protocol. Because many models accept only a single image per generation round, we use the tool described in~\cite{image_concat} to merge either multiple visual‐context images into one when necessary. Decimal answers are rounded to three decimal places, and fractional answers are rewritten in the form “numerator/denominator.” Problems with free-form answers are reformulated as multiple-choice or true/false questions, with the correct option marked as the answer. Finally, we perform a manual validation pass, revising the captions of both visual-context and visual-aid images to ensure completeness and accuracy.

\subsection{Formalization}
\label{Appendix: Data Process - Formalization}
Each sample is stored in its own directory containing two subfolders—one for visual-context images and one for visual-aid images—and a \texttt{data.json} file. Images are named in the order in which they appear in the problem or rationale (e.g., \texttt{1.png}, \texttt{2.png}). Any merged image is saved as \texttt{concatenate.png}. The \texttt{data.json} file holds all text and metadata, including the original (untranslated) text when available. The dataset is released under the CC-BY-SA-4.0 license, and VisAidMath is intended for research use only.
% \clearpage
\begin{table*}[htbp!]
% \small
\caption{Detail metadata for \textbf{VisAidMath}}
\centering
\begin{tabular}{p{0.28\linewidth} p{0.62\linewidth}}
\toprule
\textbf{Category} & \textbf{Detail} \\
\midrule 
question & Text of mathematical problem \\
 
visual context: image path & Relative path of visual context image \\
 
visual context: caption & Caption of visual context image \\
 
visual aid: image path & Relative path of visual aids image \\
 
visual aid: caption & Caption of visual aids image \\
 
choices & Options for multiple choice problems \\
 
question form & Question form includes: multiple choice, true/false, free form \\
 
answer form & Integer, decimal, fraction, and choice (for multiple choice and true/false problems) \\
 
answer & Answer of this mathematical problem \\
 
metadata: language & Original language of this mathematical problem. \\
 
metadata: source & Data source \\
 
metadata: math branch & Mathematical branch \\
 
metadata: drawing type & type of visual aids \\
\bottomrule
\rule{0pt}{10pt} \\
\end{tabular}
\label{Table: metadata}
\end{table*}

\section{Dataset Analysis}\label{Appendix: Dataset Analysis}
\subsection{Metadata}\label{Appendix: metadata}
We list the manually annotated metadata for each sample in Table \ref{Table: metadata}. ``visual context'' is optional depending on whether image is provided along with the problem. ``choices'' is not empty when question form belongs to multiple choices or true/false. ``language'' stamp the original language of each problem. We also record the Chinese version text before machine translation with "zh\_" prefix at the bottom of each data sample.

\subsection{Data Source}\label{Appendix: data source}
In accordance with the aforementioned principles, the VisAidMath benchmark has been manually collected and annotated using a diverse and balanced set of data sources.
Through an extensive search and careful examination on a case-by-case basis, we discovered that the Chinese community offers a larger pool of mathematical problems with visual aids across various complexity levels and mathematical branches compared to other communities. 
As a result, we primarily collected data from Chinese sources and subsequently performed machine translation. To ensure formula consistency, we replace LaTeX formulas with placeholders before translation and refill afterward. We also provide human validation of all samples to further ensure translation quality.
To ensure a range of difficulty levels, we categorized the data samples based on their sources into the following categories: Easy, Medium and Hard.
Additionally, metadata has been included for further in-depth analysis, discuessed in section \ref{Appendix: metadata}.

We analyze the problem complexity of 16 data source following three difficulty levels: 1) Easy: Chinese High school entrance examination 2) Medium: Chinese College Entrance Examination 3) High: Chinese Mathematical Olympiad. The complete complexity categorization of each source is listed in \ref{Table: data sources detail}. Particularly, since ``AP Calculus'' consists of both easy and medium level mathematical problems, we consider questions satisfying one of the following conditions as medium level: 1) involve coordinate axes rotation 2) cannot be resolved in one step leveraging Newton-Leibniz formula.
\begin{table*}[htbp!]
% \small
\centering
\caption{Detail of data sources}
\begin{tabular}{p{0.4\linewidth}  p{0.5\linewidth}}
\toprule
\textbf{Data Source} & \textbf{Detail} \\
\midrule 
High Textbook & Chinese high school textbook \\
Middle Practice & Chinese high school practice sheet \\
AP Easy & AP calculus (categorized into \texttt{Easy} category) \\
Middle\ Simulate & Chinese middle school simulated examination \\
AP Middle & AP calculus (categorized into \texttt{Medium} category) \\
High Practice & Chinese high school practice sheet \\
DSE Final & HKDSE final examination \\
High Final & Chinese high school final examination \\
High Simulate & Chinese high school simulated examination \\
Math Analysis Demidovich Textbook & Demidovich Problems in Mathematical Analysis \\
Analytic Geometry Lv Textbook & Analytic geometry textbook written by Lingen Lv \\
CMO Final & Chinese Mathematical Olympiad \\
CMO Practice & Chinese Mathematical Olympiad practice sheet \\
AIME Final & American Invitational Mathematics Examination (AIME) \\
AMC 8 Practice & American Mathematics Competition 8 (AMC 8) \\
AMC 10 Final & American Mathematics Competition 10 (AMC 10) \\
\bottomrule
\rule{0pt}{10pt} \\
\end{tabular}
\label{Table: data sources detail}
\end{table*}

% \section{Detail Dataset Analysis}
\section{Examples for Different Categorizations}\label{Appendix: examples for different categorizations}
\subsection{Math Branch}\label{Appendix: Graph - Math Branch Distribution}
As shown in Table \ref{tab:tab4}, this section presents examples of mathematical problems from different branches: plane geometry, solid geometry, analytic geometry, and calculus. Each example includes a question and visual aids to help illustrate the concepts.

%%%%%%%%%%%%%%%%%%%%%%%%%%%%%%%%%%%%%%%%%%%%%%%%%%%%%%%%%
%%%%%%%%%%%%%%%%%%%%%%%%%%%%%%%%%%%%%%%%%%%%%%%%%%%%%%%%%
\begin{table*}[htbp!]
\fontsize{9.0pt}{\baselineskip}\selectfont
\linespread{0.9}\selectfont
\caption{Example of mathematical problems of plane geometry.}
\begin{mybody}
\begin{minipage}[s][]{1\columnwidth}
\textbf{ID:} 1114 \\\\
\textbf{Question:} determine that the sum of the squares of the lengths of the three medians of a triangle is equal to the sum of the squares of the lengths of the three sides  $\frac{3}{4}$ . \\\\
\textbf{Visual Aids:}
\begin{minipage}[s][]{1\columnwidth}
\includegraphics[height=1.8cm]{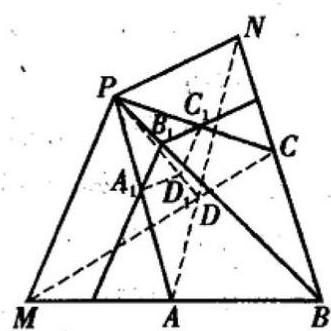}
\end{minipage}
\hspace{1mm}
\textbf{Caption of Visual Aids:} Let the three vertices of a triangle be the midpoint of $A , B , C, A B$ , the midpoint of $F, A C$ , the midpoint of $E, B C$ , and the midpoint of $D$ , connecting $A D , B E , C F$  \\\\
\end{minipage}
\end{mybody}
\label{tab:tab4}
\end{table*}
%%%%%%%%%%%%%%%%%%%%%%%%%%%%%%%%%%%%%%%%%%%%%%%%%%%%%%%%%
%%%%%%%%%%%%%%%%%%%%%%%%%%%%%%%%%%%%%%%%%%%%%%%%%%%%%%%%%

%%%%%%%%%%%%%%%%%%%%%%%%%%%%%%%%%%%%%%%%%%%%%%%%%%%%%%%%%
%%%%%%%%%%%%%%%%%%%%%%%%%%%%%%%%%%%%%%%%%%%%%%%%%%%%%%%%%
\begin{table*}[htbp!]
\fontsize{9.0pt}{\baselineskip}\selectfont
\linespread{0.9}\selectfont
\caption{Example of mathematical problems of solid geometry.}
\begin{mybody}
\textbf{ID:} 1006 \\\\
\textbf{Visual Context:}
\begin{minipage}[s][]{1\columnwidth}
\includegraphics[height=1.8cm]{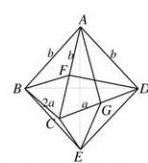}
\end{minipage}
\hspace{1mm}
\begin{minipage}[s][]{1\columnwidth}
\textbf{Caption of Visual Context:} Two congruent triangles are glued together to obtain a hexahedron with all the dihedral angles equal, and the length of the shortest prong of the hexahedron is 2. \\\\
\textbf{Question:} Given that two congruent triangular pheons are glued together to obtain a hexahedron with all the dihedral angles equal, and that the shortest prong of the hexahedron is 2, the distance between the two farthest vertices is \\\\
\textbf{Visual Aids:}
\begin{minipage}[s][]{1\columnwidth}
\includegraphics[height=1.8cm]{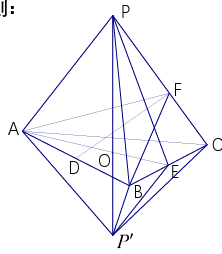}
\end{minipage}
\hspace{1mm}
\textbf{Caption of Visual Aids:} Make a plane $B D F \perp$ with edges $A C$ intersecting $A C$ at $F$ \\\\
\end{minipage}
\end{mybody}
\end{table*}
%%%%%%%%%%%%%%%%%%%%%%%%%%%%%%%%%%%%%%%%%%%%%%%%%%%%%%%%%
%%%%%%%%%%%%%%%%%%%%%%%%%%%%%%%%%%%%%%%%%%%%%%%%%%%%%%%%%

%%%%%%%%%%%%%%%%%%%%%%%%%%%%%%%%%%%%%%%%%%%%%%%%%%%%%%%%%
%%%%%%%%%%%%%%%%%%%%%%%%%%%%%%%%%%%%%%%%%%%%%%%%%%%%%%%%%
\begin{table*}[th!]
\fontsize{9.0pt}{\baselineskip}\selectfont
\linespread{0.9}\selectfont
\caption{Example of mathematical problems of analytic geometry.}
\begin{mybody}
\textbf{ID:} 619 \\\\
\textbf{Visual Context:}
\begin{minipage}[s][]{1\columnwidth}
\includegraphics[height=1.8cm]{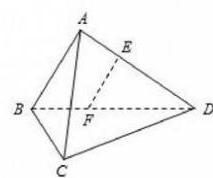}
\end{minipage}
\hspace{1mm}
\textbf{Caption of Visual Context:} In the trigonometry $\mathrm{A}-\mathrm{BCD}$, $\mathrm{AB} \perp \mathrm{AD}, \mathrm{BC} \perp \mathrm{BD}$, the plane $\mathrm{ABD} \perp$ the plane $\mathrm{BCD}$, points $\mathrm{E}, \mathrm{~F}(\mathrm{E}$ do not coincide with $\mathrm{A}, \mathrm{D}$) are on the prisms $\mathrm{AD}, \mathrm{BD}$ and $\mathrm{EF}$ $\mathrm{BD}$ respectively. $\perp \mathrm{AD}$ . $\mathrm{EF}$ / /$Plane$ $\mathrm{ABC}$ 
\begin{minipage}[s][]{1\columnwidth}
\textbf{Question:} In the trigonal pheasant  $\mathrm{A}-\mathrm{BCD}$ ,  $\mathrm{AB} \perp \mathrm{AD}, \mathrm{BC} \perp \mathrm{BD}$ , plane  $\mathrm{ABD} \perp$  plane  $\mathrm{BCD}$ , points  $\mathrm{E} , \mathrm{~F}(\mathrm{E}$  and  $\mathrm{A} , \mathrm{D}$  do not coincide) on the prong  $\mathrm{AD}, \mathrm{BD}$  respectively, and  $\mathrm{EF} \perp \mathrm{AD} . \mathrm{EF} / /$  plane  $\mathrm{ABC}$  determine whether $\mathrm{AD} \perp \mathrm{AC}$ .  \\\\
\textbf{Visual Aids:}
\begin{minipage}[s][]{1\columnwidth}
\includegraphics[height=1.8cm]{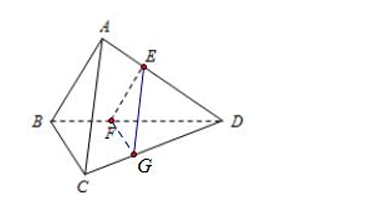}
\end{minipage}
\hspace{1mm}
\textbf{Caption of Visual Aids:} Take a point $\mathrm {G} $on the line segment $\mathrm {CD} $, connect $\mathrm {FG}, \mathrm {EG} $so that $\mathrm {FG}//\mathrm {BC}$ \\\\
\end{minipage}
\end{mybody}
\end{table*}
%%%%%%%%%%%%%%%%%%%%%%%%%%%%%%%%%%%%%%%%%%%%%%%%%%%%%%%%%
%%%%%%%%%%%%%%%%%%%%%%%%%%%%%%%%%%%%%%%%%%%%%%%%%%%%%%%%%

%%%%%%%%%%%%%%%%%%%%%%%%%%%%%%%%%%%%%%%%%%%%%%%%%%%%%%%%%
%%%%%%%%%%%%%%%%%%%%%%%%%%%%%%%%%%%%%%%%%%%%%%%%%%%%%%%%%
\begin{table*}[th!]
\fontsize{9.0pt}{\baselineskip}\selectfont
\linespread{0.9}\selectfont
\caption{Example of mathematical problems of calculus and functions.}
\begin{mybody}
\textbf{ID:} 939 \\\\
\begin{minipage}[s][]{1\columnwidth}
\textbf{Question:} Can the surface of a container formed by rotating the curve $y=C x^{4}$about the vertical axis $O y$cause the liquid surface to fall uniformly as it flows from the bottom of the container? \\\\
\textbf{Visual Aids:}
\begin{minipage}[s][]{1\columnwidth}
\includegraphics[height=1.8cm]{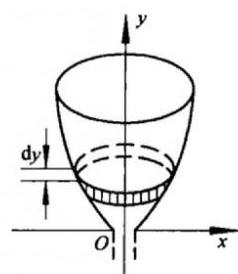}
\end{minipage}
\hspace{1mm}
\textbf{Caption of Visual Aids:} In a geometric body obtained by rotating an upward parabola along the $\mathrm {x} $axis, $\mathrm {dx} $is a line segment parallel to the $\mathrm{x} $axis, and the volume enclosed by it is drawn \\\\
\end{minipage}
\end{mybody}
\end{table*}
%%%%%%%%%%%%%%%%%%%%%%%%%%%%%%%%%%%%%%%%%%%%%%%%%%%%%%%%%
%%%%%%%%%%%%%%%%%%%%%%%%%%%%%%%%%%%%%%%%%%%%%%%%%%%%%%%%%

\subsection{Visual Aid Type}\label{Appendix: Graph - Visual Aid Type Distribution}
%%%%%%%%%%%%%%%%%%%%%%%%%%%%%%%%%%%%%%%%%%%%%%%%%%%%%%%%%
%%%%%%%%%%%%%%%%%%%%%%%%%%%%%%%%%%%%%%%%%%%%%%%%%%%%%%%%%
\begin{table*}[htbp!]
\fontsize{9.0pt}{\baselineskip}\selectfont
\linespread{0.9}\selectfont
\caption{Example of mathematical problem with auxiliary line as visual aid.}
\begin{mybody}
\textbf{ID:} 10 \\\\
\textbf{Visual Context:}
\begin{minipage}[s][]{1\columnwidth}
\includegraphics[height=1.8cm]{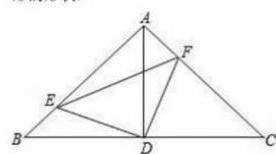}
\end{minipage}
\hspace{1mm}
\begin{minipage}[s][]{1\columnwidth}
\textbf{Caption of Visual Context:} In Rt $\triangle A B C$, $\triangle B A C=90^{\circ}$, point $D$ is the midpoint of $B C$, points $E, F$ are points on $A B, A C$ respectively, and $E D \perp F D$. \\\\
\textbf{Question:} As shown in the figure, in Rt  $\triangle A B C$ ,  $\angle B A C=90^{\circ}$ , the point  $D$  is the midpoint of  $B C$ , the point  $E , F$  is the point on  $A B , A C$ , and  $E D \perp F D$ . Can a triangle be formed with the segment  $B E , E F , F C$  as its side? If so, determine the shape of the triangle. \\\\
\textbf{Visual Aids:} 
\begin{minipage}[s][]{1\columnwidth}
\includegraphics[height=1.8cm]{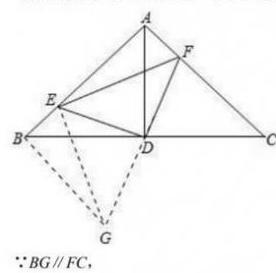}
\end{minipage}
\hspace{1mm}
\textbf{Caption of Visual Aids:} Make $B G / / F C$ , intersect with $F D$ extension line at $G$ , connect $E G$ \\\\
\end{minipage}
\end{mybody}
\end{table*}
%%%%%%%%%%%%%%%%%%%%%%%%%%%%%%%%%%%%%%%%%%%%%%%%%%%%%%%%%
%%%%%%%%%%%%%%%%%%%%%%%%%%%%%%%%%%%%%%%%%%%%%%%%%%%%%%%%%

%%%%%%%%%%%%%%%%%%%%%%%%%%%%%%%%%%%%%%%%%%%%%%%%%%%%%%%%%
%%%%%%%%%%%%%%%%%%%%%%%%%%%%%%%%%%%%%%%%%%%%%%%%%%%%%%%%%
\begin{table*}[bh!]
\fontsize{9.0pt}{\baselineskip}\selectfont
\linespread{0.9}\selectfont
\caption{Example of mathematical problem with rectangular coordinate system as visual aid.}
\begin{mybody}
\textbf{ID:} 719 \\\\
\begin{minipage}[s][]{1\columnwidth}
\textbf{Question:} Given that the vector  $\overrightarrow{O A}, \overrightarrow{O B}$  satisfies  $|\overrightarrow{O A}|=|\overrightarrow{O B}|=1, \overrightarrow{O A} \perp \overrightarrow{O B}, \overrightarrow{O C}=\lambda \overrightarrow{O A}+\mu \overrightarrow{O B}(\lambda, \mu \in R)$ , if  $M$  is the midpoint of  $A B$  and  $|\overrightarrow{M C}|=1$ then the maximum value of  $\lambda+\mu$  is ( ) \\\\
\textbf{Choices:} A. $1-\sqrt{3}$ B. $1+\sqrt{2}$ C. $\sqrt{5}$ D. $1+\sqrt{3}$
\textbf{Visual Aids:}
\begin{minipage}[s][]{1\columnwidth}
\includegraphics[height=1.8cm]{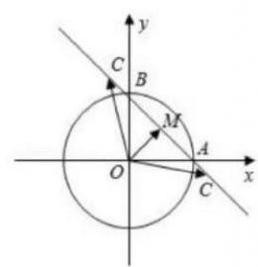}
\end{minipage}
\hspace{1mm}
\textbf{Caption of Visual Aids:} Establish a Cartesian coordinate systemDraw the vector $\overrightarrow{O A}, \overrightarrow{O B}$ M is the midpoint of AB \\\\
\end{minipage}
\end{mybody}
\end{table*}
%%%%%%%%%%%%%%%%%%%%%%%%%%%%%%%%%%%%%%%%%%%%%%%%%%%%%%%%%
%%%%%%%%%%%%%%%%%%%%%%%%%%%%%%%%%%%%%%%%%%%%%%%%%%%%%%%%%

%%%%%%%%%%%%%%%%%%%%%%%%%%%%%%%%%%%%%%%%%%%%%%%%%%%%%%%%%
%%%%%%%%%%%%%%%%%%%%%%%%%%%%%%%%%%%%%%%%%%%%%%%%%%%%%%%%%
\begin{table*}[th!]

\fontsize{9.0pt}{\baselineskip}\selectfont
\linespread{0.9}\selectfont
\caption{Example of mathematical problem with rectangular three-dimensional coordinate system as visual aid.}
\begin{mybody}
\textbf{ID:} 1099 \\\\
\textbf{Visual Context:}
\begin{minipage}[s][]{1\columnwidth}
\includegraphics[height=1.8cm]{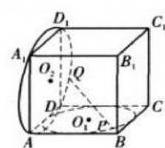}
\end{minipage}
\hspace{1mm}
\begin{minipage}[s][]{1\columnwidth}
\textbf{Caption of Visual Context:} A square $A B C D-A_{1} B_{1} C_{1} D_{1}$ has prism length $1, \odot O_{1}$ is the incircle of the square $A B C D$, $\odot O_{2}$ is the outer circle of the square $A D D_{1} A_{1}$, and $P, Q$ are the points on $\odot O_{1}, \odot O_{2}$ respectively. \\\\
\textbf{Question:} As shown in the figure, the prisms of the square  $A B C D-A_{1} B_{1} C_{1} D_{1}$  have the lengths  $1, \odot O_{1}$  as the tangent circle of the square  $A B C D$ ,  $\odot O_{2}$  as the outer circle of the square  $A D D_{1} A_{1}$ , and  $P , Q$  as the point on  $\odot O_{1} , \odot O_{2}$ . Find the range of the length of  $P Q$ . \\\\
\textbf{Visual Aids:}
\begin{minipage}[s][]{1\columnwidth}
\includegraphics[height=1.8cm]{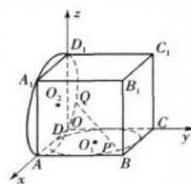}
\end{minipage}
\hspace{1mm}
\textbf{Caption of Visual Aids:} Let the origin $\mathrm{O} $and $\mathrm{D} $points coincide, along $D D_ {1}$ Establish the $\mathrm{Z} $axis in the $\mathrm{DC} $direction, establish the $\mathrm{y} $axis in the $\mathrm{DA} $direction, and establish the $\mathrm{X} $axis in the $\mathrm{DC} $direction \\\\
\end{minipage}
\end{mybody}
\end{table*}
%%%%%%%%%%%%%%%%%%%%%%%%%%%%%%%%%%%%%%%%%%%%%%%%%%%%%%%%%
%%%%%%%%%%%%%%%%%%%%%%%%%%%%%%%%%%%%%%%%%%%%%%%%%%%%%%%%%

%%%%%%%%%%%%%%%%%%%%%%%%%%%%%%%%%%%%%%%%%%%%%%%%%%%%%%%%%
%%%%%%%%%%%%%%%%%%%%%%%%%%%%%%%%%%%%%%%%%%%%%%%%%%%%%%%%%
\begin{table*}[th!]
\fontsize{9.0pt}{\baselineskip}\selectfont
\linespread{0.9}\selectfont
\caption{Example of mathematical problem with geometry graph as visual aid.}
\begin{mybody}
\textbf{ID:} 723 \\\\
\begin{minipage}[s][]{1\columnwidth}
\textbf{Question:} Given that the plane vector  $\vec{a}, \vec{b}, \vec{c}$  satisfies the angle between  $|\vec{a}|=1,|\vec{b}|=\sqrt{3}, \vec{a} \cdot \vec{b}=0, \vec{c}-\vec{a}$  and  $\vec{c}$   $-\vec{b}$  is  $\frac{\pi}{6}$ , then the maximum value of  $\vec{c} \cdot(\vec{b}-\vec{a})$  is \\\\
\textbf{Visual Aids:}
\begin{minipage}[s][]{1\columnwidth}
\includegraphics[height=1.8cm]{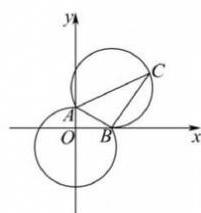}
\end{minipage}
\hspace{1mm}
\textbf{Caption of Visual Aids:} Make the tangent between circle o and circle c $I_{1}, I_{2}, I_{3}$  \\\\
\end{minipage}
\end{mybody}
\end{table*}
%%%%%%%%%%%%%%%%%%%%%%%%%%%%%%%%%%%%%%%%%%%%%%%%%%%%%%%%%
%%%%%%%%%%%%%%%%%%%%%%%%%%%%%%%%%%%%%%%%%%%%%%%%%%%%%%%%%

%%%%%%%%%%%%%%%%%%%%%%%%%%%%%%%%%%%%%%%%%%%%%%%%%%%%%%%%%
%%%%%%%%%%%%%%%%%%%%%%%%%%%%%%%%%%%%%%%%%%%%%%%%%%%%%%%%%
\begin{table*}[th!]
\fontsize{9.0pt}{\baselineskip}\selectfont
\linespread{0.9}\selectfont
\caption{Example of mathematical problem with solid geometry as visual aid.}
\begin{mybody}
\textbf{ID:} 220 \\\\
\textbf{Visual Context:}
\begin{minipage}[s][]{1\columnwidth}
\includegraphics[height=1.8cm]{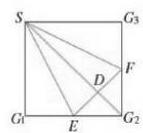}
\end{minipage}
\hspace{1mm}
\begin{minipage}[s][]{1\columnwidth}
\textbf{Caption of Visual Context:} In the square $S G_{1} G_{2} G_{3}$, $E, F$ are the midpoints of $G_{1} G_{2}, G_{2} G_{3}$ respectively, and $D$ is the midpoint of $E F$, connecting $EF$, $SE, SF$, $S G_{2}$. \\\\
\textbf{Question:} As shown in the figure, in the square  $S G_{1} G_{2} G_{3}$ ,  $E, F$  is the midpoint of  $G_{1} G_{2}, G_{2} G_{3}$  and  $D$  is the midpoint of  $E F$  respectively. If this square is folded into a tetrahedron along  $S E, S F$  and  $E F$  so that the three points of  $G_{1}, G_{2}, G_{3}$  coincide, and the point of coincidence is labeled  $G$ , then which of the edges of the tetrahedron  $S E F G$  are perpendicular to the faces of the tetrahedron? \\\\
\textbf{Visual Aids:}
\begin{minipage}[s][]{1\columnwidth}
\includegraphics[height=1.8cm]{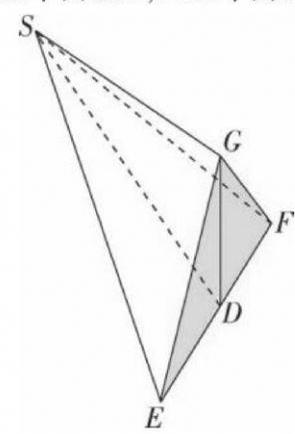}
\end{minipage}
\hspace{1mm}
\textbf{Caption of Visual Aids:} Make a four prism S-EFG and unfold it with SG GE GF to form the original square $S G_{1} G_{2} G_{3}$ \\\\
\end{minipage}
\end{mybody}
\end{table*}
%%%%%%%%%%%%%%%%%%%%%%%%%%%%%%%%%%%%%%%%%%%%%%%%%%%%%%%%%
%%%%%%%%%%%%%%%%%%%%%%%%%%%%%%%%%%%%%%%%%%%%%%%%%%%%%%%%%

%%%%%%%%%%%%%%%%%%%%%%%%%%%%%%%%%%%%%%%%%%%%%%%%%%%%%%%%%
%%%%%%%%%%%%%%%%%%%%%%%%%%%%%%%%%%%%%%%%%%%%%%%%%%%%%%%%%
\begin{table*}[th!]
\fontsize{9.0pt}{\baselineskip}\selectfont
\linespread{0.9}\selectfont
\caption{Example of mathematical problem with function graph as visual aid.}
\begin{mybody}
\textbf{ID:} 394 \\\\
\begin{minipage}[s][]{1\columnwidth}
\textbf{Question:} Through the parabola $C: y^{2}=2 p x(p\&gt; 0)$ The focus of $F$, as the slope of $2 \sqrt{2}$line $l$and parabola $C$intersect at two points $A, B$, If $\overrightarrow{A F}=\lambda \overrightarrow{F B}$, $\lambda=(\quad)$ \\\\
\textbf{Visual Aids:}
\begin{minipage}[s][]{1\columnwidth}
\includegraphics[height=1.8cm]{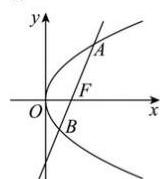}
\end{minipage}
\hspace{1mm}
\textbf{Caption of Visual Aids:} Let me draw the parabola $C: y^{2}=2 p x(p\&gt; 0)$, focus $F$, as the slope of $2 \sqrt{2}$line $l$and parabola $C$intersect at two points $A, B$ \\\\
\end{minipage}
\end{mybody}
\end{table*}
%%%%%%%%%%%%%%%%%%%%%%%%%%%%%%%%%%%%%%%%%%%%%%%%%%%%%%%%%
%%%%%%%%%%%%%%%%%%%%%%%%%%%%%%%%%%%%%%%%%%%%%%%%%%%%%%%%%
\subsection{Complexity}\label{Appendix: Graph - Complexity Distribution}
%%%%%%%%%%%%%%%%%%%%%%%%%%%%%%%%%%%%%%%%%%%%%%%%%%%%%%%%%
%%%%%%%%%%%%%%%%%%%%%%%%%%%%%%%%%%%%%%%%%%%%%%%%%%%%%%%%%
\begin{table*}[bh!]
\linespread{0.9}\selectfont
\caption{Example of mathematical problem classified into ``Easy'' category}
\begin{mybody}
\textbf{ID:} 30 \\\\
\textbf{Data Source:} Chinese Middle School Practice \\\\
\textbf{Visual Context:}
\begin{minipage}[s][]{1\columnwidth}
\includegraphics[height=1.8cm]{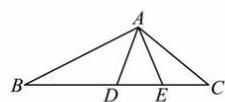}
\end{minipage}
\hspace{1mm}
\begin{minipage}[s][]{1\columnwidth}
\textbf{Caption of Visual Context:} Figure: In $\triangle A B C$, $B D=D C=A C, A E$ is the center line of $\triangle A D C$. \\\\
\textbf{Question}: It is known that, as shown in the figure:  $\triangle A B C$ ,  $B D=D C=A C, A E$  is the center line of  $\triangle A D C$ : judge if:  $A B=3 A E$ . \\\\
\textbf{Visual Aids:} 
\begin{minipage}[s][]{1\columnwidth}
\includegraphics[height=1.8cm]{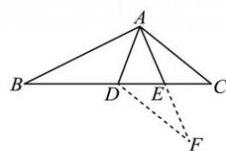}
\end{minipage}
\hspace{1mm}
\textbf{Caption of Visual Aids:} Extend $A E$ to $F$ , connect $E F=A E$ to $D F$ \\\\
\end{minipage}
\end{mybody}
\fontsize{9.0pt}{\baselineskip}\selectfont
\end{table*}
%%%%%%%%%%%%%%%%%%%%%%%%%%%%%%%%%%%%%%%%%%%%%%%%%%%%%%%%%
%%%%%%%%%%%%%%%%%%%%%%%%%%%%%%%%%%%%%%%%%%%%%%%%%%%%%%%%%

%%%%%%%%%%%%%%%%%%%%%%%%%%%%%%%%%%%%%%%%%%%%%%%%%%%%%%%%%
%%%%%%%%%%%%%%%%%%%%%%%%%%%%%%%%%%%%%%%%%%%%%%%%%%%%%%%%%
\begin{table*}[bh!]
\fontsize{9.0pt}{\baselineskip}\selectfont
\linespread{0.9}\selectfont
\caption{Example of mathematical problem classified into ``Medium'' category}
\begin{mybody}
\textbf{ID:} 650 \\\\
\textbf{Data Source:} HKDSE Examination \\\\
\textbf{Visual Context:}
\begin{minipage}[s][]{1\columnwidth}
\includegraphics[height=1.8cm]{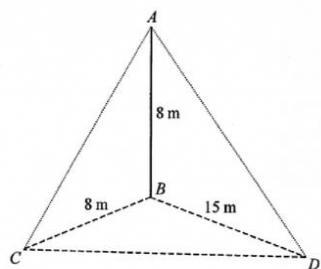}
\end{minipage}
\hspace{1mm}
\begin{minipage}[s][]{1\columnwidth}
\textbf{Caption of Visual Context:} $A B$ is a vertical pole standing on the horizontal ground $B C D$, where $\angle C B D=90^{\circ}$ \\\\
\textbf{Question}: In the figure, $A B$ is a vertical pole standing on the horizontal ground $B C D$, where $\angle C B D=90^{\circ}$. If the angle between the plane $A C D$ and the horizontal ground is $\theta$, then $\tan \theta=$ \\\\
\textbf{Visual Aids:} 
\begin{minipage}[s][]{1\columnwidth}
\includegraphics[height=1.8cm]{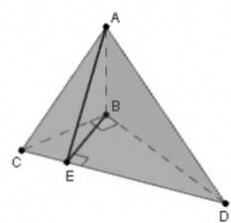}
\end{minipage}
\hspace{1mm}
\textbf{Caption of Visual Aids:} From point $B$ to point $C D$ , the perpendicular intersects point $C D$ at point $E$ , connecting $A E$ , $BE$  \\\\
\end{minipage}
\end{mybody}
\end{table*}
%%%%%%%%%%%%%%%%%%%%%%%%%%%%%%%%%%%%%%%%%%%%%%%%%%%%%%%%%
%%%%%%%%%%%%%%%%%%%%%%%%%%%%%%%%%%%%%%%%%%%%%%%%%%%%%%%%%

%%%%%%%%%%%%%%%%%%%%%%%%%%%%%%%%%%%%%%%%%%%%%%%%%%%%%%%%%
%%%%%%%%%%%%%%%%%%%%%%%%%%%%%%%%%%%%%%%%%%%%%%%%%%%%%%%%%
\begin{table*}[th!]
\fontsize{9.0pt}{\baselineskip}\selectfont
\linespread{0.9}\selectfont
\caption{Example of mathematical problem classified into ``Hard'' category}
\begin{mybody}
\textbf{ID:} 1024 \\\\
\textbf{Data Source:} Chinese Mathematical Olympiad (CMO) \\\\
\textbf{Visual Context:}
\begin{minipage}[s][]{1\columnwidth}
\includegraphics[height=1.8cm]{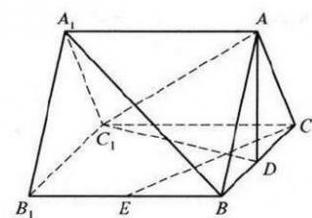}
\end{minipage}
\hspace{1mm}
\begin{minipage}[s][]{1\columnwidth}
\textbf{Caption of Visual Context:} In the right triangular prism $A B C-A_{1} B_{1} C_{1}$, $A B=A C=5, D, E$ are the midpoints of $B C, B B_{1}$ respectively, and the quadrilateral $B_{1} B C C_{1}$ is a square with side length 6. \\\\
\textbf{Question}: As shown in the figure, in the rectangular triangular prism  $A B C-A_{1} B_{1} C_{1}$ ,  $A B=A C=5, D , E$  are the midpoints of  $B C , B B_{1}$  and the quadrilateral  $B_{1} B C C_{1}$  is a square of side 6. determine that:  $A_{1} B / /$  is in the plane of  $A C_{1} D$ . \\\\
\textbf{Visual Aids:} 
\begin{minipage}[s][]{1\columnwidth}
\includegraphics[height=1.8cm]{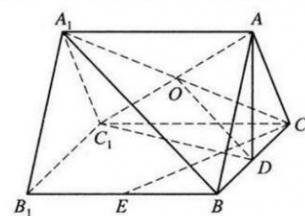}
\end{minipage}
\hspace{1mm}
\textbf{Caption of Visual Aids:} Connect $A_{1} C$ , intersect with $A C_{1}$ at $O$ point, connect $O D$ \\\\
\end{minipage}
\end{mybody}
\end{table*}
%%%%%%%%%%%%%%%%%%%%%%%%%%%%%%%%%%%%%%%%%%%%%%%%%%%%%%%%%
%%%%%%%%%%%%%%%%%%%%%%%%%%%%%%%%%%%%%%%%%%%%%%%%%%%%%%%%%

% Dataset Collection Detail
\section{Dataset Collection Detail}\label{dataset collection detail}
% \subsection{Human Labeling}
\subsection{Annotation Details}\label{annotation details}
The quality of annotation plays a crucial role in ensuring the reliability of the benchmark, and the annotation of metadata significantly impacts the depth and breadth of analysis.
In the annotation process, we instructed the annotators to label various metadata, including the \textit{math category}, \textit{source}, \textit{visual aids' type}, \textit{question form}, \textit{choices}, \textit{answer form}, \textit{answer}, and \textit{language}.
Given that the formalization of mathematical problems can vary significantly across different data sources, we specifically asked the annotators to manually extract the question and answer text from the \LaTeX{} file to ensure completeness and accuracy. Following the approach used in MathVista \cite{lu2023mathvista}, we transformed free-form questions without a purely numerical answer into multiple-choice or True/False questions. This deterministic transformation allows for a more robust evaluation.
To create precise and comprehensive descriptions of the visual context and visual aids, we provided the annotators with caption writing templates that were designed to simplify the complexity of caption writing. For more details, please refer to section \ref{Appendix: Caption Writing Templates}.
Four Chinese undergraduate students who majored in STEM during their high school period were carefully selected to form our annotation team. The collectors and annotators were compensated salary aligns with average price in local data crowdsourcing market for each data sample they collected or annotated. Furthermore, a graduate student specializing in mathematics was appointed as the verifier, offering professional annotation guidance, validation, and refinement throughout the process. All participants were volunteers who provided informed consent. The task was described as minimal risk.

\subsection{Annotation Roles}\label{key role detail}
The dataset creation pipeline involves four key roles:
\begin{itemize}%[itemsep=2pt,topsep=0pt,parsep=0pt]
\item \texttt{Administrator}: This role assigns daily collection tasks based on the progress and previous annotation feedback.
\item \texttt{Collector}: The collector searches for data that satisfies the assigned collection tasks. The collected data should be in PDF format and later transformed into \LaTeX{} files using OCR.
\item \texttt{Annotator}: The annotator first validates and refines the \LaTeX{} files by comparing the original PDF files provided by the collector with the transformed \LaTeX{} files. Then, the annotator performs interactive labeling using our designed tool. To ensure a balanced distribution across different categories, the annotator regularly analyzes the data distribution and provides feedback on the current progress and any annotation issues to the collector and administrator.
\item \texttt{Verifier}: The verifier is responsible for validating the categorization and data quality. If labels are not appropriate, they adjust the annotated captions of the context and visual aids.
\end{itemize}

\subsection{Caption Writing Templates}\label{Appendix: Caption Writing Templates}
Mathematical graphs are consists of shapes and elements bound with specific relation or theorem. To reduce manual annotation work and enhance caption consistency, we standardize the caption writing for visual context and visual aids by defining templates for certain visual elements. The annotators should caption image referring to these templates as listed in \ref{Table: caption template}.
\begin{table*}[th!]

% \small
\centering
\caption{Caption writing templates of various visual aid types for annotators' reference.}
\begin{tabular}{p{0.4\linewidth} p{0.5\linewidth}}
\toprule
\textbf{Category} & \textbf{\hspace{2em}Detail} \\
\midrule 
\multirow{3}{*}{Auxiliary Line} & 
\begin{itemize}
\item[1.1] Connect the points a,b to make the line segment
\item[1.2] Connect points a,b and extend them to intersect the CD line at point E.
\item[1.3] Make a vertical line AB through point a.
\end{itemize} \\
\midrule
\multirow{3}{*}{\parbox[c]{5cm}{Rectangular Three-Dimensional Coordinate System}} & 
\begin{itemize}
\item[2.1] Establish a three-dimensional rectangular coordinate system with point o as the origin and oa as the x-axis positive direction and ob as the y-axis and oc as the z-axis positive direction. 
\end{itemize} \\
\midrule
\multirow{2}{*}{Rectangular Coordinate System} &
\begin{itemize}
\item[3.1] Take point o as the origin oa as the x-axis positive direction ob as the y-axis Establish a right-angle coordinate system
\item[3.2] With point o as origin oa as x-axis positive direction y-axis positive direction as x-axis rotated 90 degrees counterclockwise/clockwise
\end{itemize} \\
\midrule
\multirow{2}{*}{Function Graph} &
\begin{itemize}
\item[4.1] Draw the $y^2=ax$ image in the real coordinate system with 0 as the origin.
\item[4.2] Draw the $y^2=ax$ image in the real coordinate system with 0 as the origin and the y=ax image intersecting at point k in the first quadrant and at point D in the second quadrant.
\end{itemize} \\
\midrule
Riemann integral problem &
\begin{itemize}
\item[5.1] Draw the graph of [\texttt{function}] function in interval [\texttt{interval}] and draw the Riemann integral graph in units of [\texttt{unit}] with values on the right/left/middle side.
\end{itemize} \\
\bottomrule
\rule{0pt}{10pt} \\
\end{tabular}
\label{Table: caption template}
\end{table*}

\subsection{Dataset Creation Pipeline}
The pipeline is shown in the Figure 9.
\begin{figure*}[htbp!]
  \begin{center}
  % \centering
    \includegraphics[width=0.9\textwidth]{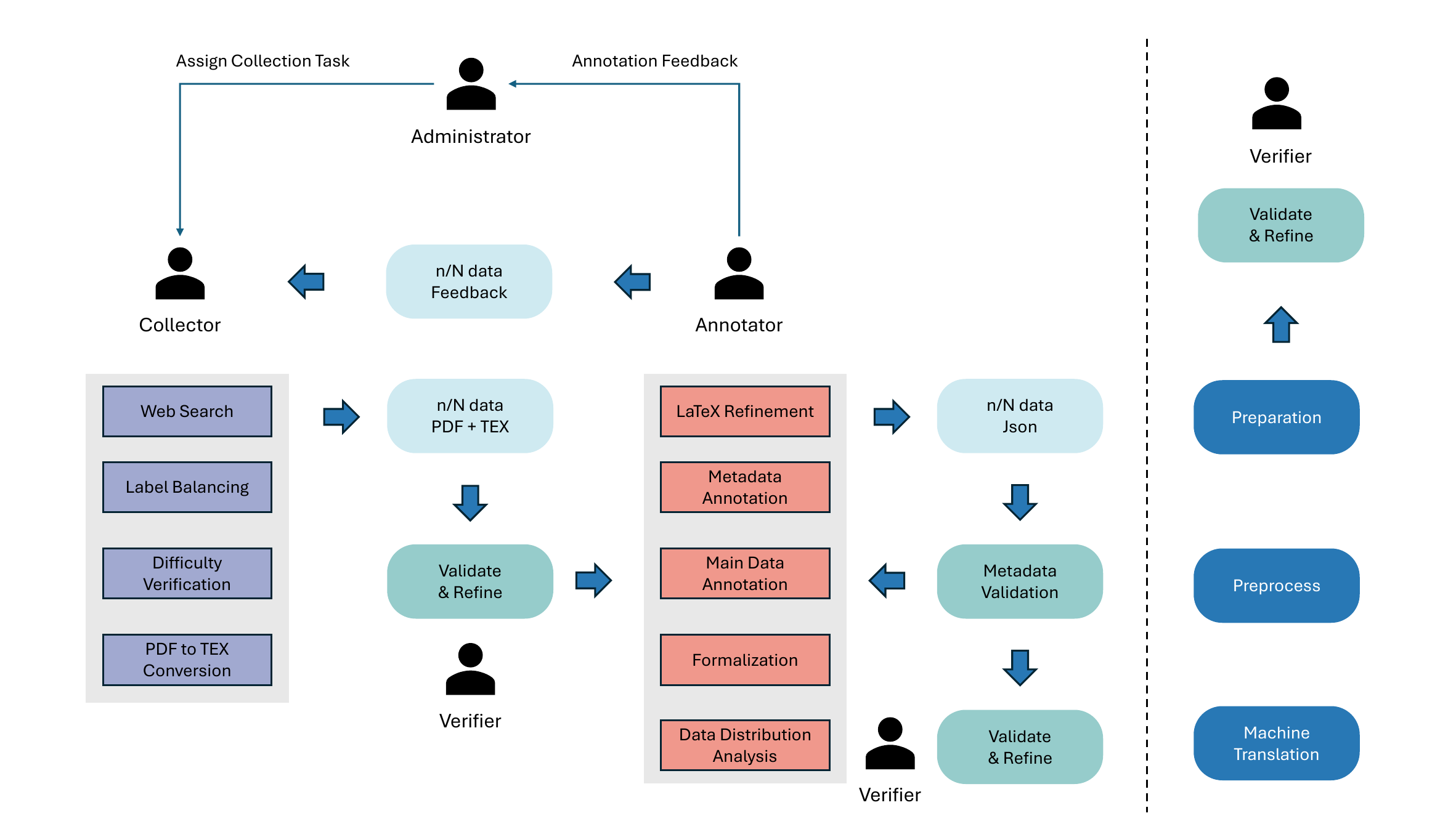}
    \caption{Pipeline invloving data collection, annotation and verification.}
    \label{Figure: Data Collection And Annotation Pipeline}
  \end{center}
\end{figure*}

\subsection{Human Annotation Interface}
Shown in the Figure 10.
\begin{figure*}[!hb]
  \begin{center}
  % \centering
    \includegraphics[width=0.8\textwidth]{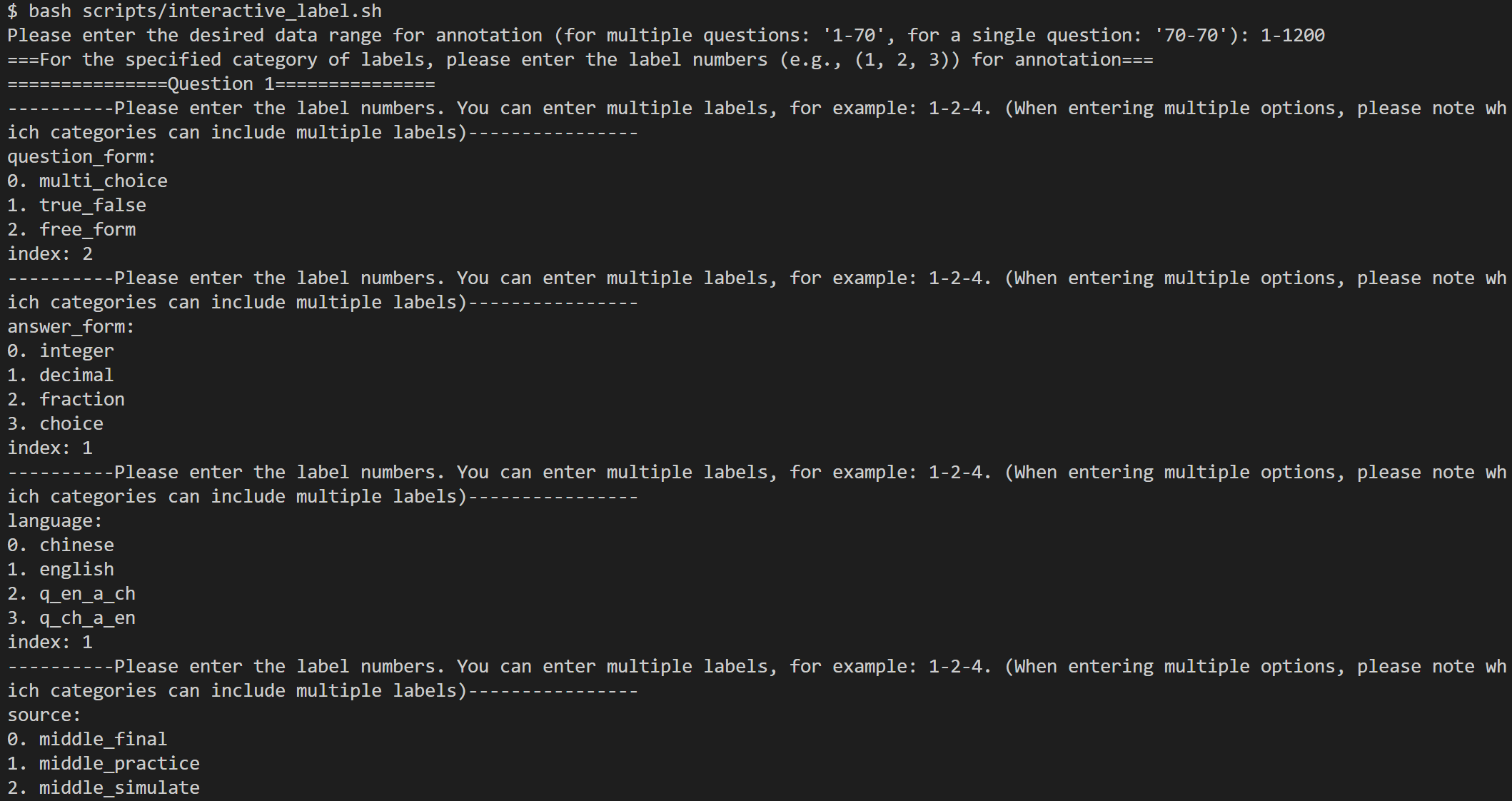}
    \caption{Annotation tool for interactive labeling}
    \label{Figure: Annotation tool for interactive labeling}
  \end{center}
\end{figure*}
\clearpage

\begin{figure}[htbp]
    \centering
    \includegraphics[width=0.6\textwidth]{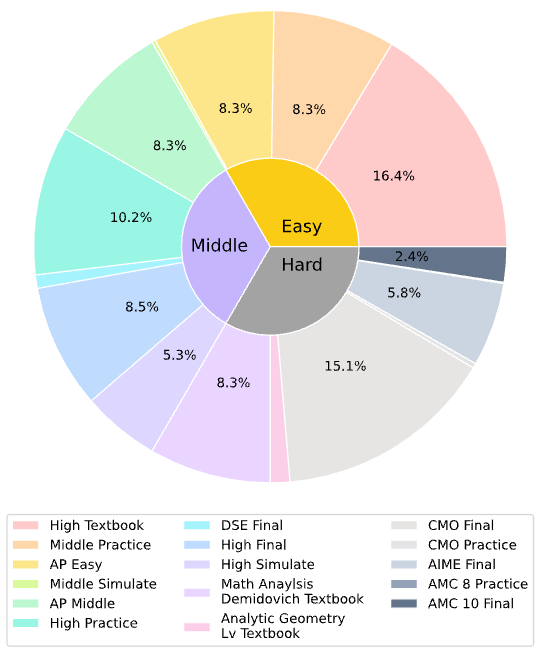}
    \caption{Distribution of data sources and difficulty levels.}
    \label{Data Source and Complexity Distribution}
    % \vspace{-1cm}
\end{figure}

%%%%
\clearpage
%%%%%%%%%%%%%%%%%%%%%%%%%%%%%%%%%%%%%%%%%%%%%%%%%%%%%%%%%
%%%%%%%%%%%%%%%%%%%%%%%%%%%%%%%%%%%%%%%%%%%%%%%%%%%%%%%%%
\begin{table*}[th!]
\fontsize{9.0pt}{\baselineskip}\selectfont
\linespread{0.9}\selectfont
\caption{An MPS example with visual context from MathVista and reasoning provided by GPT4V.}
\begin{mybody}
\textbf{Visual Context:} \\\\   
\begin{minipage}[s][]{1\columnwidth}
\includegraphics[height=1.8cm]{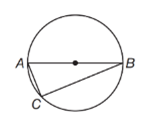}
\end{minipage}
\hspace{1mm}
\begin{minipage}[s][]{1\columnwidth}
\textbf{Question:} $\overline{AB}$ is a diameter, $AC=8$ inches, and $BC=15$ inches. Find the radius of the circle.
\end{minipage}
\end{mybody}
%%%%%%%%%%%%%%%%%%%%%%%%%%%%%%%%%%%%%%%%%%%%%%%%%%%%%%%%%
\begin{mybody}
\textbf{GPT4 Reasoning:} To solve for the radius of the circle, we can use the fact that triangle ACB is a right triangle because AB is the diameter of the circle (Thales' theorem).
Given:
•	AC=8 inches
•	BC=15 inches
Using the Pythagorean theorem in triangle ACB:
AB2=AC2+BC2
Substitute the given values:
AB2=82+152=64+225=289
Thus,
AB=289=17
Since AB is the diameter, the radius r is half of AB:
r=172=8.5
The correct answer is (C) 8.5.

\end{mybody}
\label{table: 1}
\end{table*}
%%%%%%%%%%%%%%%%%%%%%%%%%%%%%%%%%%%%%%%%%%%%%%%%%%%%%%%%%
%%%%%%%%%%%%%%%%%%%%%%%%%%%%%%%%%%%%%%%%%%%%%%%%%%%%%%%%%

%%%%%%%%%%%%%%%%%%%%%%%%%%%%%%%%%%%%%%%%%%%%%%%%%%%%%%%%%
%%%%%%%%%%%%%%%%%%%%%%%%%%%%%%%%%%%%%%%%%%%%%%%%%%%%%%%%%
\begin{table*}[th!]
\fontsize{9.0pt}{\baselineskip}\selectfont
\linespread{0.9}\selectfont
\begin{mybody}
\textbf{Visual Context:} \\\\   
\begin{minipage}[s][]{1\columnwidth}
\includegraphics[height=1.8cm]{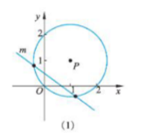}
\end{minipage}
\hspace{1mm}
\begin{minipage}[s][]{1\columnwidth}
\textbf{Question:} It is known that the line $m: 3x+4y-2=0$ and the circle $P: x^{2}-y^{2}-2x-2y=0$. (2) From the graph drawn in (1), determine the relationship between the line $m$ and the circle $P$. If they intersect, find the length of the chord of the line $m$ intercepted by the circle $P$. If they are tangent or disjoint, give a proof.
\end{minipage}
\caption{An example of using only arithmetic methods to solve mathematical problems}
\end{mybody}
%%%%%%%%%%%%%%%%%%%%%%%%%%%%%%%%%%%%%%%%%%%%%%%%%%%%%%%%%
\begin{mybody}
\textbf{GPT4 Reasoning:} Step 1: Solve $m$ for $y$ to find the slope and intercept: ... \red{$3x+4y-2=0 \Rightarrow y=-\frac{3}{4}x+\frac{1}{2}.$} \\\\
Step 2: Complete the square for circle $P$ to find its center and radius:\red{$x^2-2x+y^2-2y=0 \Rightarrow (x-1)^2+(y-1)^2=2.$} \\\\
Step 3: \red{Substitute $y = -\frac{3}{4}x + \frac{1}{2}$ into the circle's equation and solve for $x$.}\\\\
**Step 4:** If two points of intersection exist, $A(x_1, y_1)$ and $B(x_2, y_2)$, use the distance formula to find the chord length: \\\\
\red{$\text{Chord Length} = \sqrt{(x_2-x_1)^2 + (y_2-y_1)^2}.$}
\\\\The final answer will be the length of the chord, provided as an integer in LaTeX format. Let’s carry out the calculations to find the result ...
\end{mybody}
\label{table: detail motivation arithmetic sample}
\end{table*}
%%%%%%%%%%%%%%%%%%%%%%%%%%%%%%%%%%%%%%%%%%%%%%%%%%%%%%%%%
%%%%%%%%%%%%%%%%%%%%%%%%%%%%%%%%%%%%%%%%%%%%%%%%%%%%%%%%%

%%%%%%%%%%%%%%%%%%%%%%%%%%%%%%%%%%%%%%%%%%%%%%%%%%%%%%%%%
%%%%%%%%%%%%%%%%%%%%%%%%%%%%%%%%%%%%%%%%%%%%%%%%%%%%%%%%%
\begin{table*}[th!]
\fontsize{9.0pt}{\baselineskip}\selectfont
\linespread{0.9}\selectfont
\begin{mybody}
\textbf{Visual Context:} \\\\   
\begin{minipage}[s][]{1\columnwidth}
\includegraphics[height=1.8cm]{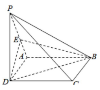}
\end{minipage}
\hspace{1mm}
\begin{minipage}[s][]{1\columnwidth}
\textbf{Question:} Given that $P D \perp$plane $ABCD, ABCD$ is A square, and $E$ is the midpoint of $PA$, judge :(1) $PC/ /$plane $EBD$
\end{minipage}
\caption{A successful direct visual-aided reasoning example}
\end{mybody}
%%%%%%%%%%%%%%%%%%%%%%%%%%%%%%%%%%%%%%%%%%%%%%%%%%%%%%%%%
\begin{mybody}
\textbf{GPT4 Constructed Visual Aids:} Connect $\mathrm{PC}$ and $\mathrm{AC}$, which intersect with $\mathrm{BD}$ at point $\mathrm{O}$. Draw a line $\mathrm{EO}$ from point $\mathrm{E}$ to $\mathrm{O}$. \\\\
\begin{minipage}[s][]{1\columnwidth}
\includegraphics[height=1.8cm]{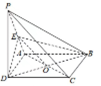}
\end{minipage}
\hspace{1mm}
\textbf{GPT4 Reasoning:} Since $\mathrm{E}$ is the midpoint of $\mathrm{PA}$ and \red{$\mathrm{O}$ is the midpoint of diagonal $\mathrm{AC}$ of square $\mathrm{ABCD}$:$\therefore \mathrm{EO}$ is parallel to $\mathrm{PC}$} (as midlines in a trapezoid).$\because \mathrm{PC} \not \subset$ plane $\mathrm{EBD}$ and \red{$\mathrm{EO} \subset$ plane $\mathrm{EBD}$}$\therefore \mathrm{PC}$ is parallel to plane $\\mathrm{EBD}$.So the answer is True.
\end{mybody}
\label{table: detail motivation visual-aided success sample}
\end{table*}
%%%%%%%%%%%%%%%%%%%%%%%%%%%%%%%%%%%%%%%%%%%%%%%%%%%%%%%%%
%%%%%%%%%%%%%%%%%%%%%%%%%%%%%%%%%%%%%%%%%%%%%%%%%%%%%%%%%

\section{Detail Experiment Settings}\label{append: detail expr setting}
\subsection{Hyperparameters}\label{append: hyperparameters}
We utilize the default inference settings for each LLMs and LMMs in our experiments. Only specific hyperparameters that are necessary to clarify are listed in Table \ref{Table: close_source_mdoel_hyperparameters} and \ref{Table: open_source_mdoel_hyperparameters}. We conduct open source model inference based on \cite{Wolf_Transformers_State-of-the-Art_Natural_2020}. We report a single run score for each experiment.
\begin{table*}[th!]
% \small
\centering
\caption{Hyperparameter for close LLMs and LMMs in experiments.}
\begin{tabular}{p{0.3\linewidth}  p{0.60\linewidth}}
\toprule
\textbf{Model} & \textbf{Hyperparameters} \\
\midrule 
GPT3.5 & model = \texttt{gpt-3.5-turbo}, temperature = 0, max\_tokens = 4000 \\
\midrule
GPT-4 &  model = \texttt{gpt-4-turbo-2024-04-09}, temperature = 0, max\_tokens = 4000 \\
\midrule
Qwen-VL-Plus & model = \texttt{qwen-vl-plus}, temperature = 0.7, max\_tokens = 4000 \\
\midrule
Gemini-Pro-Vision & model = \texttt{gemini-pro-vision}, temperature = 0.4, max\_tokens = 4000 \\
\midrule
Claude-3-Sonnet & model = \texttt{claude-3-sonnet}, temperature = 1, max\_tokens = 4000 \\
\midrule
GPT4V & model = \texttt{gpt-4-vision-preview}, temperature = 0, max\_tokens = 4000 \\
\midrule
GPT4.1 & model = \texttt{gpt-4-1-2025-04-14}, temperature = 0, max\_tokens = 4000 \\
\midrule
O4-Mini & model = \texttt{o4-mini-2025-04-16}, temperature = 0, max\_tokens = 4000 \\
\midrule
Doubao-Seed-1.6 & model = \texttt{doubao-seed-1-6-250615}, temperature = 0, max\_tokens = 4000 \\
\bottomrule
\midrule
\midrule
\rule{0pt}{10pt} \\
\end{tabular}
\label{Table: close_source_mdoel_hyperparameters}
\end{table*}
\begin{table*}[th!]
% \small
\centering
\caption{Hyperparameter for open LLMs and LMMs in experiments.}
\begin{tabular}{p{0.30\linewidth}  p{0.60\linewidth}}
\toprule
\textbf{Model} & \textbf{Hyperparameters} \\
\midrule 
Llama2-7B & model = \texttt{Llama-2-7b-chat-hf}, precision = bfloat16, temperature = 1.0, max\_tokens = 4000 \\
\midrule
Mistral-7b-Instruct-v0.2 & model = \texttt{mistral-7b-instruct-v0.2}, precision = bfloat16, temperature = 1.0, max\_tokens = 4000 \\
\midrule
LLaVA-Next-Mistral-7B & model = \texttt{llava-v1.6-mistral-7b-hf}, precision = float16, temperature = 1.0, max\_tokens = 4000 \\
\midrule
InternLM-XComposer2-VL & model = \texttt{internlm-xcomposer2-vl-7b}, precision = float16, temperature = 1.0, max\_tokens = 4000 \\
\midrule
VL-Cogito & model = \texttt{VL-Cogito}, precision = float16, temperature = 0, max\_tokens = 4000 \\
\midrule
Qwen2.5-VL & model = \texttt{Qwen2.5-VL-72B-Instruct}, precision = float16, temperature = 0, max\_tokens = 4000 \\
\midrule
InternVL3 & model = \texttt{InternVL3-78B}, precision = float16, temperature = 0, max\_tokens = 4000 \\
\bottomrule
\rule{0pt}{10pt} \\
\end{tabular}
\label{Table: open_source_mdoel_hyperparameters}
\end{table*}
\subsection{Reasoning Prompter}
We list the ICL prompts for assigning different models to perform reasoning under six task settings in mathematical domain: 1) CQ2A 2) CQ2VA 3) CQpV2A 4) pCQ2A 5) pCQ2VA 6) pCQpV2A.
Task instructions for each task are listed in Table \ref{Table: task_instruction}. ICL examples can be found at our code open sourced after acceptance. %Table \ref{Table: icl examples}.
\begin{table*}[th!]\tiny
\small
\centering
\caption{Instructions for different mathematical problem solving tasks with visual context inside input.}
\begin{tabular}{p{0.2\linewidth} p{0.7\linewidth}}
\toprule
\textbf{Answer Type} & \textbf{With Visual Context} \\
\midrule 
\fontsize{6}{11} CQ2A & Please read the following math problem and the related image. After that, \\
\midrule
CQ2VA & Please read the following math problem and the related image, then conceive the additional mathematical diagram or visual aid upon provided image to help illustrate the problem, clarify the problem or assist in finding solution. The drawing shape includes auxiliary line, plane geometry graph, solid geometry graph, function graph, plane coordinate and three-dimensional coordinate. These additional drawings can enhance understanding of the problem and potentially find new insights or approaches to solving it. Write drawing description of these additional diagram in text, and express math formula with tex gramma. After that, reason based on the drawing description, \\
\midrule
CQpV2A & Please read the following math problem and the related image. Read the provided drawing description of additional mathematical diagram or visual aid upon provided image in latex format, which helps illustrate the problem, clarify the problem or assist in finding solution. The drawing shape includes auxiliary line, plane geometry graph, solid geometry graph, function graph, plane coordinate and three-dimensional coordinate. These additional drawings can enhance understanding of the problem and potentially find new insights or approaches to solving it. After that, based on the drawing description, \\
\midrule
pCQ2A & Please read the following math problem and captions of related visual context. After that, \\
\midrule
pCQ2VA & Please read the following math problem and captions of related visual context, then conceive the additional mathematical diagram or visual aid upon provided image to help illustrate the problem, clarify the problem or assist in finding solution. The drawing shape includes auxiliary line, plane geometry graph, solid geometry graph, function graph, plane coordinate and three-dimensional coordinate. These additional drawings can enhance understanding of the problem and potentially find new insights or approaches to solving it. Write drawing description of these additional diagram in text, and express math formula with tex gramma. After that, reason based on the drawing description, \\
\midrule
pCQpV2A & Please read the following math problem and captions of related visual context. Read the provided drawing description of additional mathematical diagram or visual aid upon provided image in latex format, which helps illustrate the problem, clarify the problem or assist in finding solution. The drawing shape includes auxiliary line, plane geometry graph, solid geometry graph, function graph, plane coordinate and three-dimensional coordinate. These additional drawings can enhance understanding of the problem and potentially find new insights or approaches to solving it. After that, based on the drawing description, \\

\bottomrule
\rule{0pt}{10pt} \\
\end{tabular}
\label{Table: task_instruction}
\end{table*}

\subsection{Instruction across Answer Types}
To facilitate accuracy evaluation, each sample is bound with with non-ambiguous result across integer, fraction, decimal and choice. We define choice as answer type for multiple choice and true/false problems. The specific task instruction for each output answer type is shown in Table \ref{Table: answer_type_instruction_table}.
\begin{table*}[th!]
% \small
\centering
\caption{Instructions for mathematical problem solving toward different answer types.}
\begin{tabular}{p{0.30\linewidth}  p{0.60\linewidth}}
\toprule
\textbf{Answer Type} & \textbf{Task Instruction} \\
\midrule 
Integer & Answer the question requiring a integer answer in latex format and provide the final value at the end (e.g., -1, 0, 1, 2, 3). \\
\midrule
Decimal & Aswer the question requiring a decimal answer in latex format and provide the final value at the end. Make sure the decimal answer is a floating-point number with three decimal place (e.g., 1.234, 2.345, 3.456). \\
\midrule
Fraction & Answer the question requiring an fraction answer in latex format and provide the final value at the end. Make sure the fraction answer use '/' as fraction bar and both numerator and denominator are integers (e.g., 1/2, 2/3, 3/4). \\
\midrule
Choice (Multiple-Choice) & Answer the question in latex format and provide the correct option at the end. Make sure the option answer can only be single capital letter (e.g., A, B, C, D). \\
\midrule
Choice (True/False) &  Answer the question in latex format and provide the correct option at the end. Make sure the option answer can only be word `True' or `False'.\\
\bottomrule
\rule{0pt}{10pt} \\
\end{tabular}
\label{Table: answer_type_instruction_table}
\end{table*}

\subsection{Visual Aid Extraction}
Reasoning prompter instruct models to provide visual aids, reasoning and answer following certain format. We can directly extract visual aids and followup reasoning from output generated by LLM and LMMs with sufficient zero-shot reasoning capability (eg. GPT-4, GPT-4V). However, few weaker models (eg. llama2, LLaVA-Next-Mistral-7B) fail to provide standardized output. We regard these output as both visual-aid and followup reasoning for later answer extraction and similarity evaluation respectively.
\subsection{Answer Extraction Prompter}\label{Appendix: Answer Extractor}
Answer extraction is conducted upon the extracted reasoning text from model output. We provide five ICL examples instructing GPT-4o mini to extract answer from varied model outputs. Examples are shown in Table \ref{Table: examples_answer_extraction}.
\begin{table*}[th!]
\centering
\caption{3 ICL examples leveraged to prompt GPT-4 for answer extraction. Full 5 examples will be listed in code open sourced after acceptance.}
\small
\begin{tabular}{p{0.30\linewidth}  p{0.60\linewidth}}
 \toprule
 \textbf{Element} & \textbf{Prompt} \\
 \midrule
 Example 1 & 
 \begin{minipage}[s][]{0.6\textwidth}
 \textbf{Instruction:} Answer the question requiring a integer answer and provide the final value at the end (e.g., -1, 0, 1, 2, 3). \\
 \textbf{Question:} Acute Angle $\triangle A B C, B E$vertical $A C$at $E, C D$vertical $A B$at $\mathrm{D}$, given acute Angle $\triangle A B C, B E$vertical $A C$at $E, B C D$ vertical $A $in $\mathrm{D}$, $\mathrm{BC} = 25$, $\mathrm {CE} = 7$, $\mathrm{BD} = 15$, $\mathrm{BE}, \mathrm{CD} $to $\mathrm{H} $, connection $\mathrm{DE}$, $\mathrm{DE} $for circle diameter, and $\mathrm {AC} $to another $\mathrm {F}$, Then the length of $\mathrm{AF}$ is\\
 
 \textbf{Model response:} Connect $\mathrm{DF}$, then $\angle \mathrm{DFE}=\angle \mathrm{BEC}=90^{\circ}$, so $\mathrm{DF} / / \mathrm{BE}$. In Rt $\triangle \mathrm{BEC}$, $\mathrm{BC}=25, \mathrm{CE}=7$, so $\mathrm{BE}=24, \cos \angle \mathrm{ACB} = \frac{7}{25}$ , $\sin \angle A C B=\frac{24}{25}$, in Rt $\triangle B C D$, $B C=25, B D=15$, so $\cos \angle A B C=\frac{3}{5}, \sin \angle A B C=\frac{4}{5}$. So $\sin A=\sin (\angle A B C+\angle A C B)=\sin \angle A B C \cos \angle A C B+\cos \angle A B C \sin \angle A C B=\frac{4}{5}$, in Rt $\triangle A B E$, $\sin$ $A= \frac{B E}{A B}=\frac{24}{A B}=\frac{4}{5}$, we get $A B=30$, and $B D=15$, so the point $D$ is the midpoint of $A B$, and $D F$ is the median of $\triangle A B E$, so $A F=\frac{1}{2} A E=$ $\frac{1}{2 } \sqrt{30^{2}-24^{2}}=9$. So the final answer is 9\\
 
 \textbf{Extracted answer:} \green{\textbf{9}}\\
 \end{minipage}
 \\
 \midrule
 Example 2 & 
 \begin{minipage}[s][]{0.6\textwidth}
 \textbf{Instruction:} Answer the question requiring a decimal answer and provide the final value at the end. Make sure the decimal answer is a floating-point number with three decimal place (e.g., 1.234, 2.345, 3.456) \\
 \textbf{Question:} In a right angle $\triangle A B C$, $\angle D A C=45^{\circ}, A D=13, A B=12$. Calculate the length of $C D$. \\
 \textbf{Model response:} Let $D E=x$, then $C E=A E=13+x$, $\frac{12}{13+x}=\frac{5}{x} \quad x=\frac{65}{7}$ , $D E=\frac{65}{7} \quad C E=\frac{156}{7}$ , $C D=sqrt{D E^{3}+C E^{2}}= \frac{169}{7}=24.1$ \\
 
 \textbf{Extracted answer:} \green{\textbf{24.143}}
 \end{minipage}
 \\
 \midrule
 Example 3 & 
 \begin{minipage}[s][]{0.6\textwidth}
 \textbf{Instruction:} Answer the question requiring an fraction answer and provide the final value at the end. Make sure the fraction answer use '/' as fraction bar and both numerator and denominator are integers (e.g., 1/2, 2/3, 3/4). \\
 \textbf{Question:} Linear $x + y = \sqrt{3} $a and round $x ^ + y ^ {2} {2} = a ^ {2} + (a - 1) ^ {2} $intersect at point of $a, $B, $O $is the origin of coordinates, if $\\triangle is a O B $is triangle, the real value of $a $ \\
 
 \textbf{Model response:} From the question, the radius of the circle $r=\sqrt{a^{2}+(a-1)^{2}}$; the distance from the centre of the circle to the line $x+y-\sqrt{3} a=0$ $d=\frac{|\sqrt{3} a|}{\sqrt{2}}$; because $\triangle A O B$ is a positive triangle, so $d=\frac{\sqrt{3}}{2} r$, i.e. $\frac{|\sqrt{3} a|}{\sqrt{2}}=\frac{\sqrt{3}}{2} \sqrt{a^{2}+(a-1)^{2}}$, which solves $a=\frac{1}{2}$ \\
 
 \textbf{Extracted answer:} \green{\textbf{1/2}}
 \end{minipage}
 \\

 \bottomrule
 \rule{0pt}{10pt} \\
\end{tabular}
\label{Table: examples_answer_extraction}
\end{table*}

.

% More Experimental Results
% Result of Other Tasks
\section{More Experimental Results}
\begin{figure*}[htbp!]
  % \begin{center}
  % \centering
    \includegraphics[width=1\textwidth]{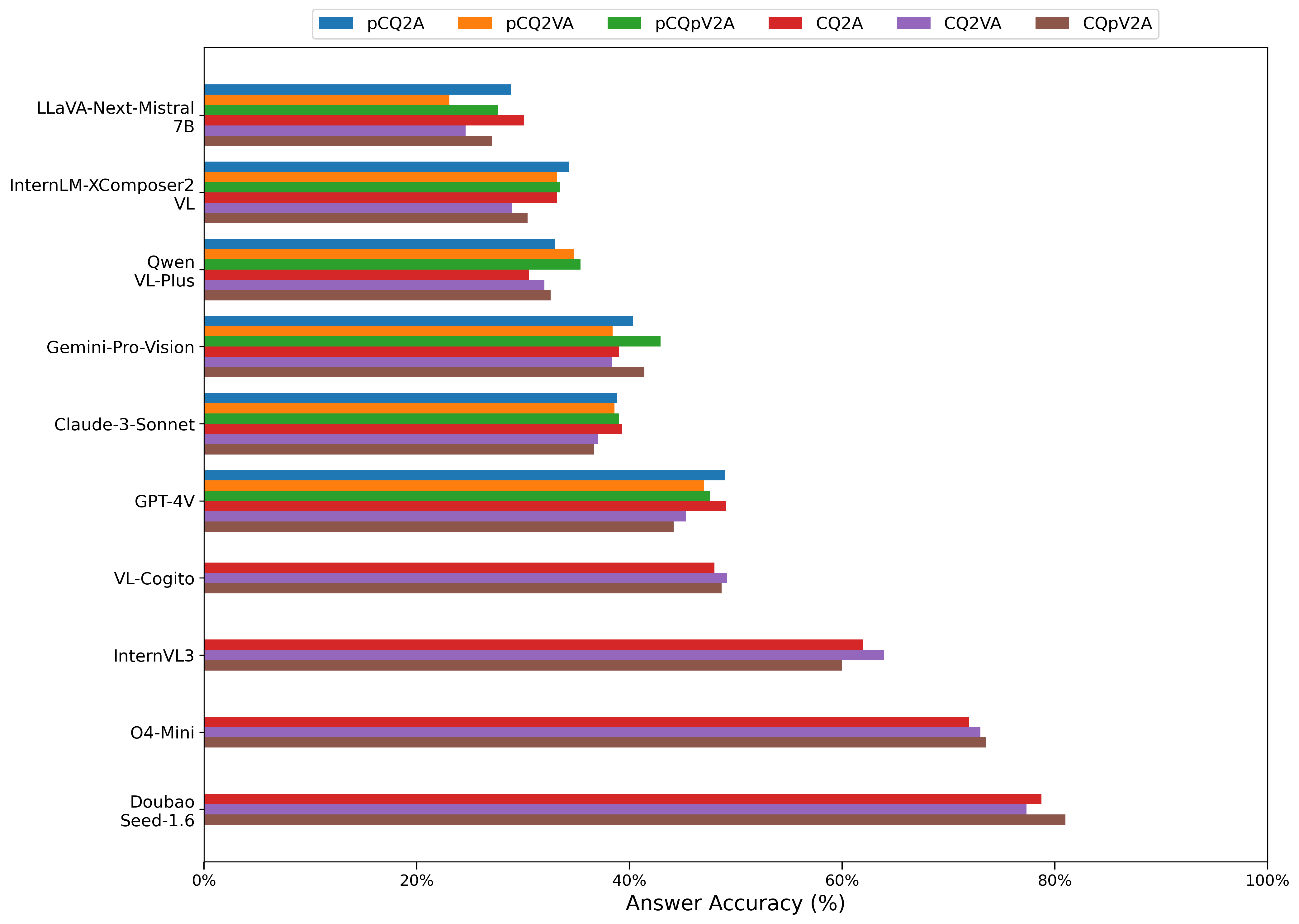}
    \caption{CQ2A is corresponding to GR, CQ2VA is corresponding to D-VAR, and CQpV2A is corresponding to I-VAR.}
    \label{Figure: result comparision of different tasks}
  % \end{center}
\end{figure*}

More experimental results are in Figure \ref{Figure: result comparision of different tasks}.
\subsection{Results of other tasks}
\begin{table*}[t!]
\vspace{-3mm}
\caption{Accuracy scores on \textbf{General Reasoning task (GR)} task upon \textbf{VisAidMath}. Meanings of all abbreviations are listed below. ALL $\rightarrow$ overall accuracy. For Mathematical Branch: PLG $\rightarrow$ plane geometry, SDG $\rightarrow$ solid geometry, AYG$\rightarrow$ analytic geometry, CAL: calculus and functions. Visual Aids Types: AXL $\rightarrow$ auxiliary line, RTC $\rightarrow$ rectangular coordinate, THC $\rightarrow$ rectangular three-dimensional coordinate, PLG $\rightarrow$ plane geometry graph, SDG $\rightarrow$ solid geometry graph, FUG $\rightarrow$ function graph. The highest scores in chunk and in general have been emphasized with purple and green to facilitate comparison respectively.}
\centering
 % \small
 \renewcommand\tabcolsep{5pt} % column space
 \resizebox{1.0\linewidth}{!}{
    \begin{tabular}{l|c|cccc|cccccc}
    \toprule
    \header{Model} & \header{ALL} & \header{PLG} & \header{SDG} & \header{AYG} & \header{CAL} & \header{AXL} & \header{RTC} & \header{THC} & \header{PLG} & \header{SDG} & \header{FUG}  \\ 
    \midrule
    \multicolumn{12}{l}{\hfill \textit{Heuristics  Baselines}} \\
    \midrule
    % \rowcolor[rgb]{0.93,0.93,0.93} \multicolumn{11}{l}{\textit{Heuristic baselines}} \\
    Random Answer & 24.42 & 21.54 & 34.31 & 21.45 & 20.07 & 24.44 & 20.87 & 35.16 & 10.53 & 32.89 & 21.50 \\
    Frequent Answer & 40.83 & 28.92 & 50.65 & 40.36 & 44.22 & 32.79 & 47.25 & 74.73 & 20.00 & 47.73 & 44.53 \\
    \midrule
    \multicolumn{12}{l}{\hfill \textit{Large Language Models (LLMs): Text-Only Input} } \\
    \midrule
    Llama2-7B & 23.25 & 22.77 & 29.74 & 17.82 & 22.11 & 22.80 & 19.72 & 28.57 & 8.42 & 28.29 & 21.11 \\
    Mistral-7b-Instruct-v0.2 & 25.58 & 24.31 & 29.41 & 25.09 & 23.47 & 24.59 & 25.46 & 25.27 & 6.32 & 26.32 & 25.91 \\
    GPT3.5 & 37.75 & 32.62 & 44.44 & 33.82 & 40.14 & 37.85 & 38.30 & 40.66 & 17.89 & 43.42 & 38.20 \\
    GPT4 &  \high{51.17} & \high{41.54} & \high{47.39} & \high{50.91} & \high{65.99} & \high{45.45} & \high{55.73} & \high{59.34} & \high{22.11} & \high{49.34} & \high{61.80} \\
    \midrule
    \multicolumn{12}{l}{\hfill \textit{Large Multimodal Models (LMMs): Text-Only Input} } \\
    \midrule
    LLaVA-Next-Mistral-7B & 28.83 & 26.15 & 35.29 & 24.36 & 29.25 & 27.72 & 28.67 & 30.77 & 10.53 & 35.53 & 28.79 \\
    InternLM-XComposer2-VL & 34.33 & 28.00 & 45.75 & 28.36 & 35.03 & 32.64 & 33.49 & 53.85 & 13.68 & 36.18 & 33.40 \\
    Qwen-VL-Plus & 33.00 & 34.15 & 39.54 & 29.09 & 28.57 & 34.87 & 30.05 & 34.07 & 13.68 & 43.42 & 30.52 \\
    Gemini-Pro-Vision & 40.33 & 31.38 & \high{50.98} & 35.27 & 43.88 & 36.66 & 41.28 & 53.85 & 17.89 & 49.34 & 41.84 \\
    Claude-3-Sonnet & 38.83 & 27.08 & 47.06 & 38.18 & 43.88 & 33.38 & 39.68 & \high{57.14} & 16.84 & 48.03 & 42.80 \\
    GPT4V & \high{49.00} & \high{42.46} & 46.08 & \high{48.73} & \high{59.52} & \high{43.82} & \high{52.75} & 52.75 & \high{22.11} & \high{50.00} & \high{57.58} \\
    \midrule
    \multicolumn{12}{l}{\hfill \textit{Large Multimodal Models (LMMs): Multimodal Input}} \\
    \midrule
    LLaVA-Next-Mistral-7B & 30.08 & 27.08 & 37.91 & 25.82 & 29.25 & 28.32 & 29.13 & 41.76 & 9.47 & 34.87 & 28.98 \\
    InternLM-XComposer2-VL & 33.17 & 26.77 & 39.87 & 32.00 & 34.35 & 30.85 & 32.80 & 43.96 & 11.58 & 36.84 & 34.93 \\
    Qwen-VL-Plus & 30.58 & 29.23 & 35.62 & 28.73 & 28.57 & 31.15 & 29.82 & 27.47 & 13.68 & 41.45 & 30.52 \\
    Gemini-Pro-Vision & 39.00 & 27.38 & 49.02 & 36.36 & 43.88 & 35.32 & 40.37 & 52.75 & 14.74 & 48.68 & 42.03 \\
    Claude-3-Sonnet & 39.33 & 30.15 & 46.41 & 37.45 & 43.88 & 34.72 & 38.99 & 56.04 & 16.84 & 47.37 & 42.42 \\
    GPT4V & 49.08 & 41.54 & 47.39 & 48.73 & 59.52 & 43.82 & 53.21 & 51.65 & 24.21 & 51.97 & 57.97 \\ 
    VL-Cogito & 49.17 & 40.31 & 53.92 & 53.74 & 49.45  & 45.31 & 53.85 & 52.40 & 55.26 & 50.23 & 20.00  \\
    Qwen2.5-VL-72B & 52.25  & 42.77 & 50.00 & 61.22 & 56.36  & 45.01 & 50.55 & 62.38 & 53.95 & 58.49 & 23.16  \\
    InternVL3.5-38B & 63.92  & 57.85 & 61.11 & 73.47 & 64.00  & 56.33 & 72.53 & 71.21 & 55.92 & 67.20 & 54.74  \\
    GPT-4.1 & 62.42  & 54.77 & 58.50 & 72.79 & 64.73  & 56.93 & 72.53 & 70.25 & 56.58 & 66.51 & 54.74  \\
    O4-Mini & 73.00  & 68.92 & 76.47 & \best{74.83} & 72.00  & 69.75 & 87.91 & 74.09 & 73.03 & 71.10 & 56.84  \\
    Doubao-Seed-1.6 & \best{77.33}  & \best{75.38} & \best{81.37} & 74.49 & \best{78.18}  & \best{75.26} & \best{90.11} & \best{76.97} & \best{76.32} & \best{75.92} & \best{68.42}  \\
    \bottomrule
    \end{tabular}
   }
    % \vspace{-2mm}
    
\vspace{-3mm}
\label{Table: 2Answ}
\end{table*}

\begin{table*}[t!]
\vspace{-3mm}
\caption{Accuracy scores on \textbf{Indirect Visual-Aided Reasoning (I-VAR)} task upon \textbf{VisAidMath}. Meanings of all abbreviations are listed below. ALL $\rightarrow$ overall accuracy. For Mathematical Branch: PLG $\rightarrow$ plane geometry, SDG $\rightarrow$ solid geometry, AYG$\rightarrow$ analytic geometry, CAL: calculus and functions. Visual Aids Types: AXL $\rightarrow$ auxiliary line, RTC $\rightarrow$ rectangular coordinate, THC $\rightarrow$ rectangular three-dimensional coordinate, PLG $\rightarrow$ plane geometry graph, SDG $\rightarrow$ solid geometry graph, FUG $\rightarrow$ function graph. The highest scores in chunk and in general have been emphasized with purple and green to facilitate comparison respectively.}
\centering
 % \small
 \renewcommand\tabcolsep{5pt} % column space
 \resizebox{1.0\linewidth}{!}{
    \begin{tabular}{l|c|cccc|cccccc}
    \toprule
    \header{Model} & \header{ALL} & \header{PLG} & \header{SDG} & \header{AYG} & \header{CAL} & \header{AXL} & \header{RTC} & \header{THC} & \header{PLG} & \header{SDG} & \header{FUG}  \\ 
    \midrule
    \multicolumn{12}{l}{\hfill \textit{Heuristics  Baselines}} \\
    \midrule
    % \rowcolor[rgb]{0.93,0.93,0.93} \multicolumn{11}{l}{\textit{Heuristic baselines}} \\
    Random Answer & 24.42 & 21.54 & 34.31 & 21.45 & 20.07 & 24.44 & 20.87 & 35.16 & 10.53 & 32.89 & 21.50 \\
    Frequent Answer & 40.83 & 28.92 & 50.65 & 40.36 & 44.22 & 32.79 & 47.25 & 74.73 & 20.00 & 47.73 & 44.53 \\
    \midrule
    \multicolumn{12}{l}{\hfill \textit{Large Language Models (LLMs): Text-Only Input} } \\
    \midrule
    Llama2-7B & 24.08 & 21.23 & 31.05 & 25.82 & 18.37 & 25.04 & 22.71 & 31.87 & 7.37 & 30.26 & 22.46 \\
    Mistral-7b-Instruct-v0.2 & 28.33 & 27.69 & 33.33 & 24.73 & 27.21 & 27.72 & 27.29 & 34.07 & 14.74 & 32.89 & 27.26 \\
    GPT3.5 & 36.33 & 31.08 & 39.22 & 34.91 & 40.48 & 33.08 & 37.84 & 50.55 & 14.74 & 39.47 & 39.73 \\
    GPT4 & \high{52.17} & \high{42.77} & \high{49.02} & \high{53.09} & \high{64.97} & \high{46.94} & \high{57.11} & \high{54.95} & \high{20.00} & \high{52.63} & \high{62.76} \\
    \midrule
    \multicolumn{12}{l}{\hfill \textit{Large Multimodal Models (LMMs): Text-Only Input} } \\
    \midrule
    LLaVA-Next-Mistral-7B & 27.67 & 27.38 & 33.99 & 24.36 & 24.49 & 27.42 & 25.00 & 29.67 & 11.58 & 33.55 & 25.91 \\
    InternLM-XComposer2-VL & 33.50 & 28.31 & 43.46 & 32.36 & 29.93 & 33.68 & 32.80 & 49.45 & 13.68 & 41.45 & 31.86 \\
    Qwen-VL-Plus & 35.42 & 31.69 & 40.85 & 38.18 & 31.29 & 36.51 & 39.22 & 40.66 & 15.79 & 39.47 & 34.93 \\
    Gemini-Pro-Vision & 42.92 & 32.31 & \high{51.96} & 40.73 & 47.28 & 39.79 & 43.35 & \high{57.14} & 17.89 & 47.37 & 45.87 \\
    Claude-3-Sonnet & 39.00 & 31.38 & 42.16 & 41.45 & 41.84 & 35.92 & 40.14 & 46.15 & 17.89 & 42.11 & 43.19 \\
    GPT4V & \high{47.58} & \high{40.31} & 47.71 & \high{42.55} & \high{60.20} & \high{42.32} & \high{47.94} & 50.55 & \high{21.05} & \high{55.26} & \high{53.93} \\
    \midrule
    \multicolumn{12}{l}{\hfill \textit{Large Multimodal Models (LMMs): Multimodal Input}} \\
    \midrule
    LLaVA-Next-Mistral-7B & 27.08 & 27.69 & 32.03 & 23.64 & 24.49 & 27.42 & 24.31 & 26.37 & 11.58 & 32.89 & 25.72 \\
    InternLM-XComposer2-VL & 30.42 & 20.00 & 39.54 & 33.09 & 29.93 & 26.97 & 31.88 & 40.66 & 10.53 & 34.87 & 32.25 \\
    Qwen-VL-Plus & 32.58 & 31.69 & 30.39 & 37.45 & 31.29 & 33.23 & 38.99 & 25.27 & 16.84 & 37.50 & 34.55 \\
    Gemini-Pro-Vision & 41.42 & 29.54 & 48.69 & 41.09 & 47.28 & 37.85 & 43.81 & 45.05 & 14.74 & 48.03 & 46.07 \\
    Claude-3-Sonnet & 36.67 & 24.92 & 39.22 & 42.18 & 41.84 & 32.04 & 40.37 & 41.76 & 14.74 & 43.42 & 43.76 \\
    GPT4V & 44.17 & 37.54 & 37.25 & 42.91 & 59.86 & 38.60 & 47.25 & 36.26 & 17.89 & 48.03 & 53.74 \\
    VL-Cogito & 48.67  & 36.31 & 51.96 & 53.74 & 54.18  & 44.86 & 47.25 & 54.89 & 51.32 & 50.46 & 16.84  \\
    Qwen2.5-VL-72B & 54.67  & 44.92 & 55.56 & 63.95 & 55.27  & 47.99 & 56.04 & 62.76 & 54.61 & 58.03 & 22.11  \\
    InternVL3.5-38B & 60.00  & 50.46 & 60.13 & 65.99 & 64.73  & 55.44 & 64.84 & 66.41 & 59.87 & 62.39 & 34.74  \\
    GPT-4.1 & 60.50  & 48.92 & 58.82 & 71.43 & 64.36  & 52.01 & 75.82 & 69.48 & 59.21 & 68.35 & 46.32  \\
    O4-Mini & 73.50  & 72.00 & 73.20 & 72.79 & 76.36  & 70.34 & 78.02 & 75.43 & 73.03 & 74.08 & 70.53  \\
    Doubao-Seed-1.6  & \best{81.00}  & \best{80.92} & \best{81.05} & \best{79.93} & \best{82.18}  & \best{78.84} & \best{86.81} & \best{81.19} & \best{78.95} & \best{80.28} & \best{75.79}  \\
    \bottomrule
    \end{tabular}
    }
    % \vspace{-2mm}
    
\vspace{-3mm}
\label{Table: Aid2Answ}
\end{table*}

\subsection{Quantitative Analysis}\label{Appendix: Quantitative Analysis}
As shown in Figures \ref{Appendix: performance_math_branch_CQ2A}, \ref{Appendix: performance_math_branch_CQ2VA}, and \ref{Appendix: performance_math_branch_CQpV2A}, the accuracy scores of LMMs and LLMs across math branches show that GPT4V and Gemini-Pro-Vision excel in different tasks with visual aids.
\subsubsection{Performance across Math Branches}\label{Appendix: Quantitative Analysis - Math Category}
\begin{figure}[bh!]
  \begin{center}
  % \centering
    \includegraphics[width=1\textwidth]{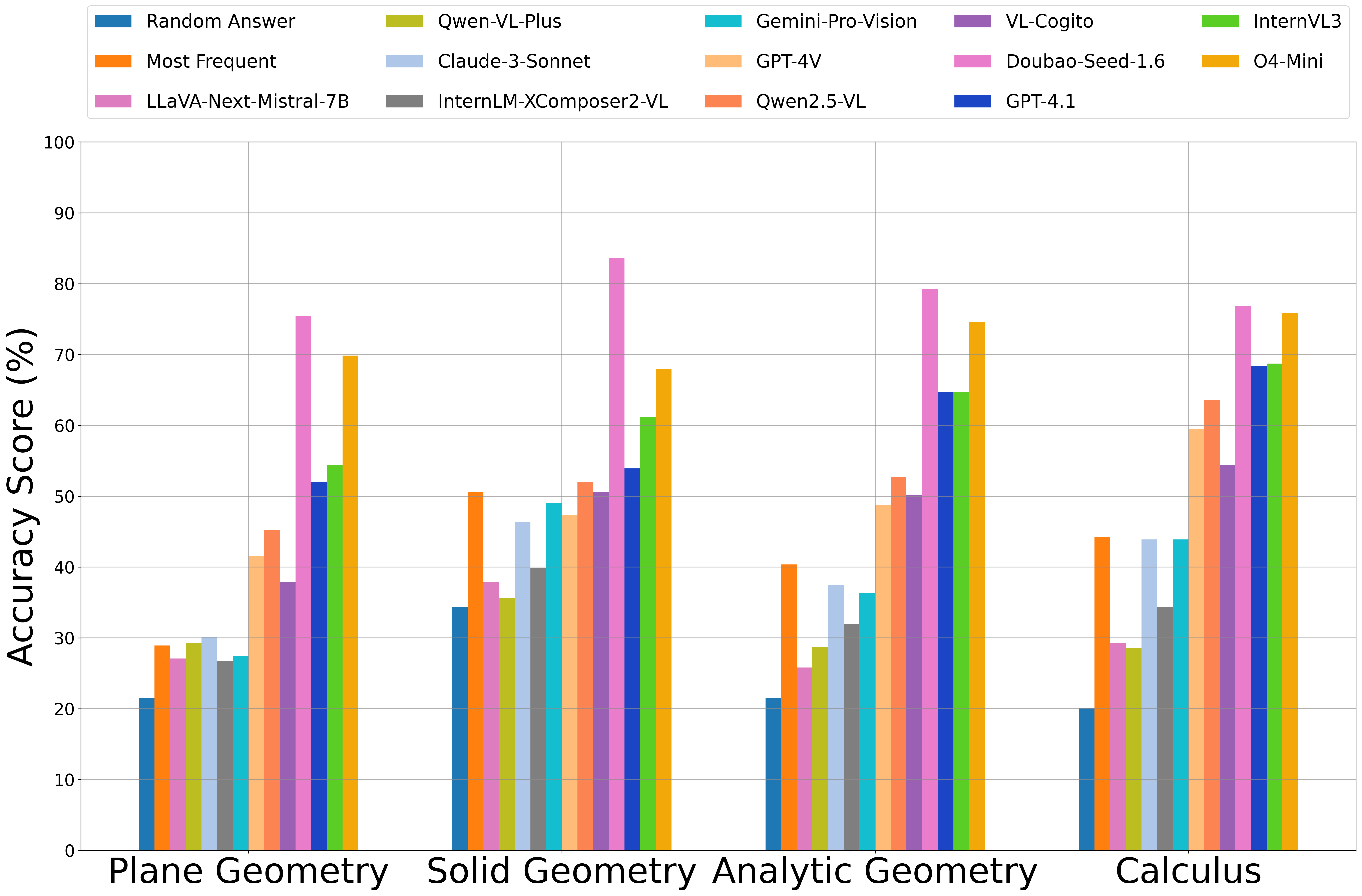}
    \caption{Accuracies of LLMs and LMMs upon CQ2A (General Text-Only Reasoning) task across math branches.}
    \label{Appendix: performance_math_branch_CQ2A}
  \end{center}
\end{figure}

\begin{figure}[bh!]
  \begin{center}
  % \centering
    \includegraphics[width=1\textwidth]{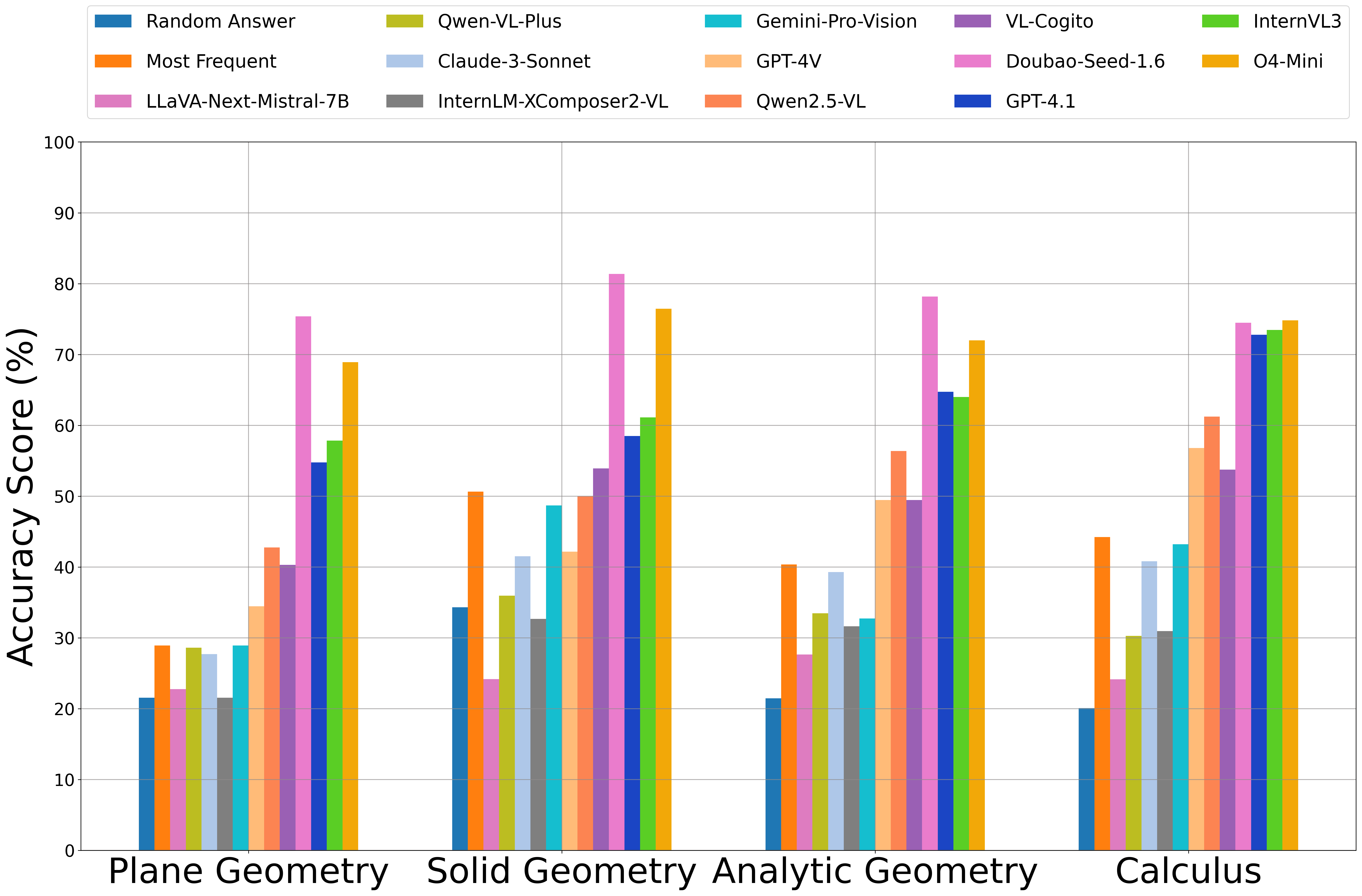}
    \caption{Accuracies of LLMs and LMMs upon CQ2VA (Direct Visual-Aided Reasoning) task across math branches.}
    \label{Appendix: performance_math_branch_CQ2VA}
  \end{center}
\end{figure}

\begin{figure}[bh!]
  \begin{center}
  % \centering
    \includegraphics[width=1\textwidth]{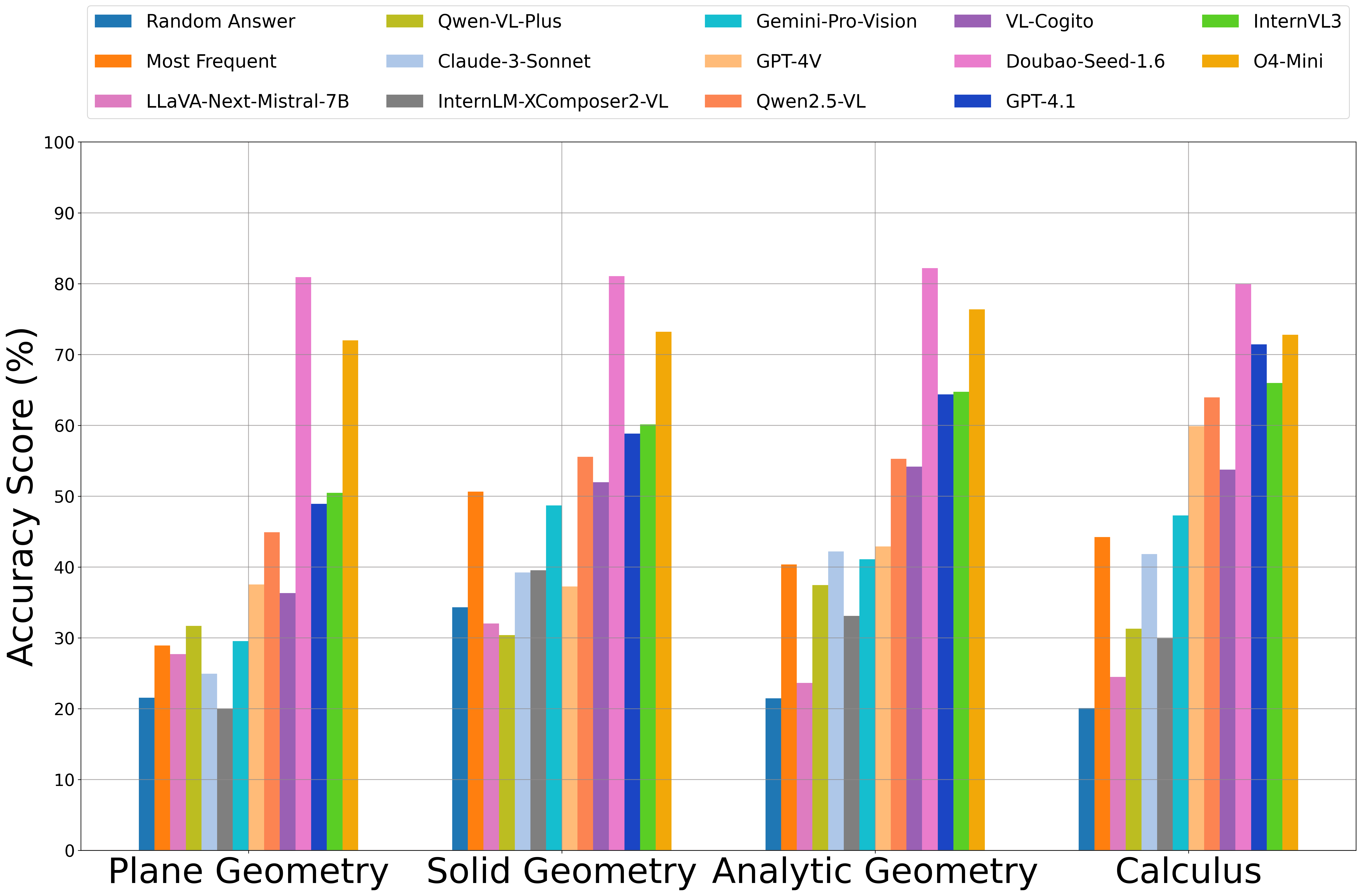}
    \caption{Accuracies of LLMs and LMMs upon CQpV2A (Indirect Reasoning) task across math branches.}
    \label{Appendix: performance_math_branch_CQpV2A}
  \end{center}
\end{figure}

Figure \ref{Appendix: performance_math_branch_CQ2A}, \ref{Appendix: performance_math_branch_CQ2VA}, and \ref{Appendix: performance_math_branch_CQpV2A} illustrate the accuracy scores of LMMs upon three tasks with image input across four math branches. GPT4V outperforms other models in problems within ``plane geometry'', ``analytic geometry'', and ``calculus and functions'' branches. Gemini-Pro-Vision achieves the highest score on solid geometry problems. Notably Claude-3-Sonnet and InternLM-XComposer2-VL both achieves comparable results toward GPT4V in ``solid geometry'' branch when reason with provided visual-aided, exhibiting robustness and enhanced capabilities in spatial understanding and visual reasoning under ``solid geometry''. GPT4V underperforms in direct visual-aided reasoning, exhibiting significant deficiency processing implicit visual information. Gemini-Pro-Vision performs better at ``analytic geometry'' and ``calculus and functions'' with provided visual aids, demonstrating better understanding of visual context within these mathematical branches.

\label{append_performance_ImageInput_math_branch}

\begin{figure}[b!]
  \begin{center}
  % \centering
    \includegraphics[width=1\textwidth]{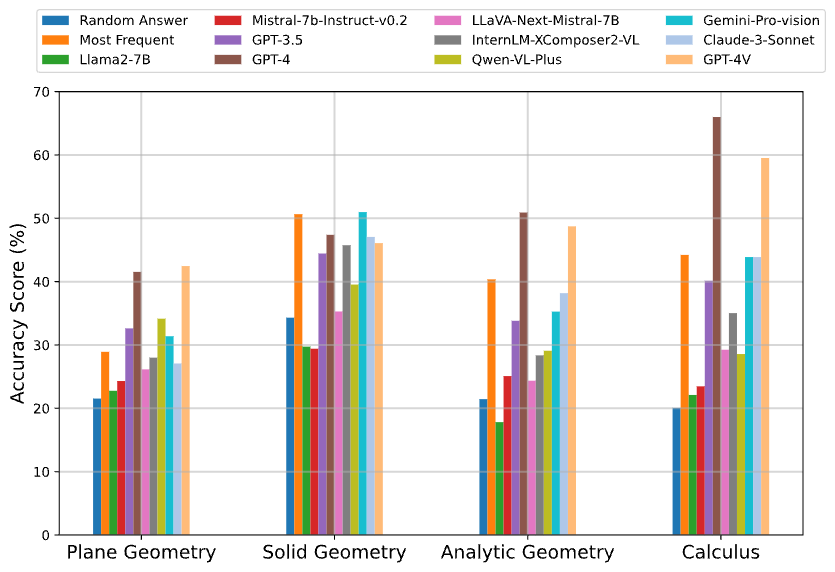}
    \caption{Accuracies of LLMs and LMMs upon pCQ2A (General Text-Only Reasoning) task across math branches.}
    \label{Appendix: performance_math_branch_pCQ2A}
  \end{center}
\end{figure}

\begin{figure}[b!]
  \begin{center}
  % \centering
    \includegraphics[width=1\textwidth]{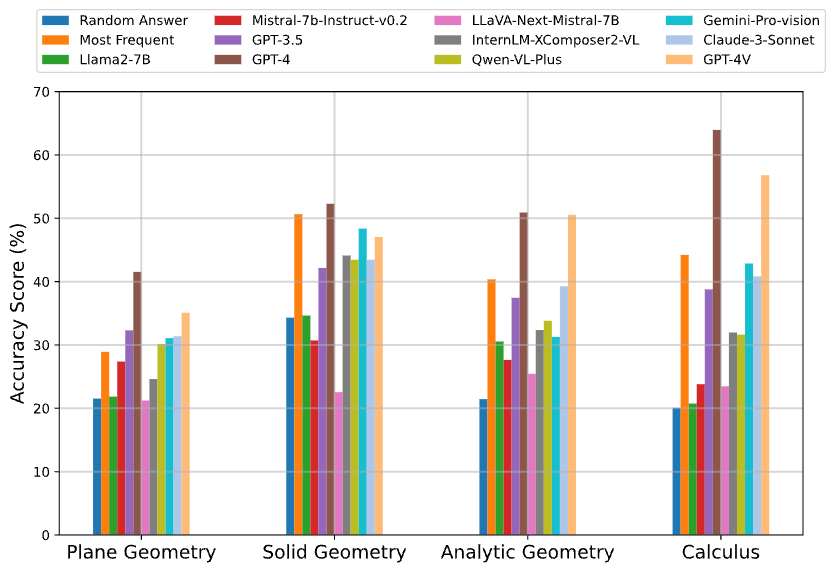}
    \caption{Accuracies of LLMs and LMMs upon pCQ2VA (Direct Visual-Aided Reasoning) task across math branches.}
    \label{Appendix: performance_math_branch_pCQ2VA}
  \end{center}
\end{figure}

\begin{figure}[b!]
  \begin{center}
  % \centering
    \includegraphics[width=1\textwidth]{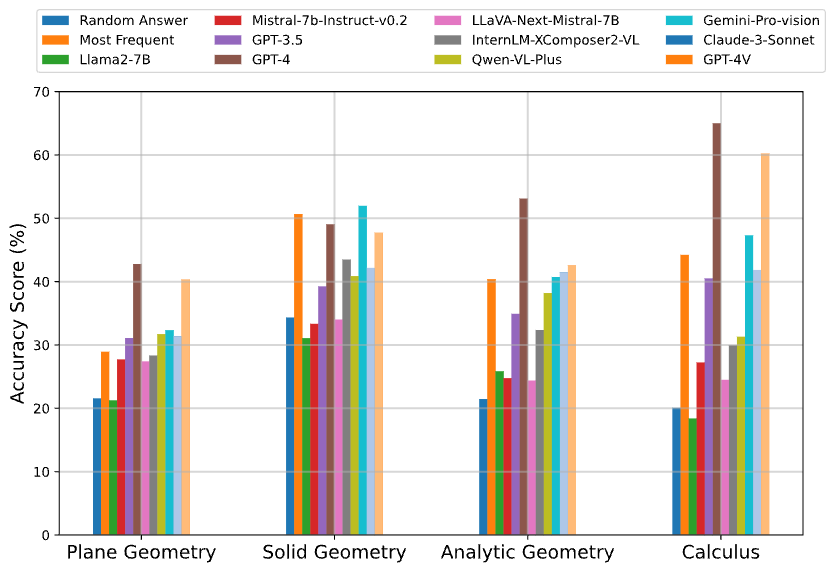}
    \caption{Accuracies of LLMs and LMMs upon pCQpV2A (Indirect Reasoning) task across math branches.}
    \label{Appendix: performance_math_branch_pCQpV2A}
  \end{center}
\end{figure}

The accuracy scores of both LLMs and LMMs on three tasks with image caption for visual context across math branches is shown in Figure \ref{Appendix: performance_math_branch_pCQ2A}, \ref{Appendix: performance_math_branch_pCQ2VA}, and \ref{Appendix: performance_math_branch_pCQpV2A}. In text-only reasoning, GPT4 outperforms other models across most branches and tasks. GPT4V and Gemini-Pro-Vision achieve the highest score at ``plane geometry'' and ``solid geometry'' respectively in pCQ2A task. Gemini-Pro-Vision also attains the best score at ``solid geometry'' branch, manifesting robust visual context utilization within ``solid geometry'' branch. GPT4V and GPT4 share similar performances over ``plane geometry'' problems. However, performance of these two models deviates when conduct indirect reasoning. However, GPT-4V achieves significantly lower accuracy on ``analytic geometry'' branch, demonstrating difficulty for GPT-4V to handle additoinal visual information on ``analytic geometry'' branch. Gemini-Pro-Vision outperform other models after introduce visual aids into reasoning at ``solid geometry'' category, indicating better three-dimensional spatial information understanding and processing. Overall, more models perform reasoning better with visual aids on ``solid geometry'' and ``analytic geometry'' problems, possibly because visual aids in these problems are bound tighter to the reasoning path. In opposite, visual aids within ``plane geometry'' problems still often leave broad decision space, thus making it harder to utilize the additonnal visual context. Since various problems of ``calculus and functions'' can often be solved by generic method, most models maintain the highest accuracy in this branch.

\
% \clearpage
\subsubsection{Performance across Complexities}\label{Appendix: Quantitative Analysis - Complexity}
\begin{figure}[b!]
  \begin{center}
  % \centering
    \includegraphics[width=1\textwidth]{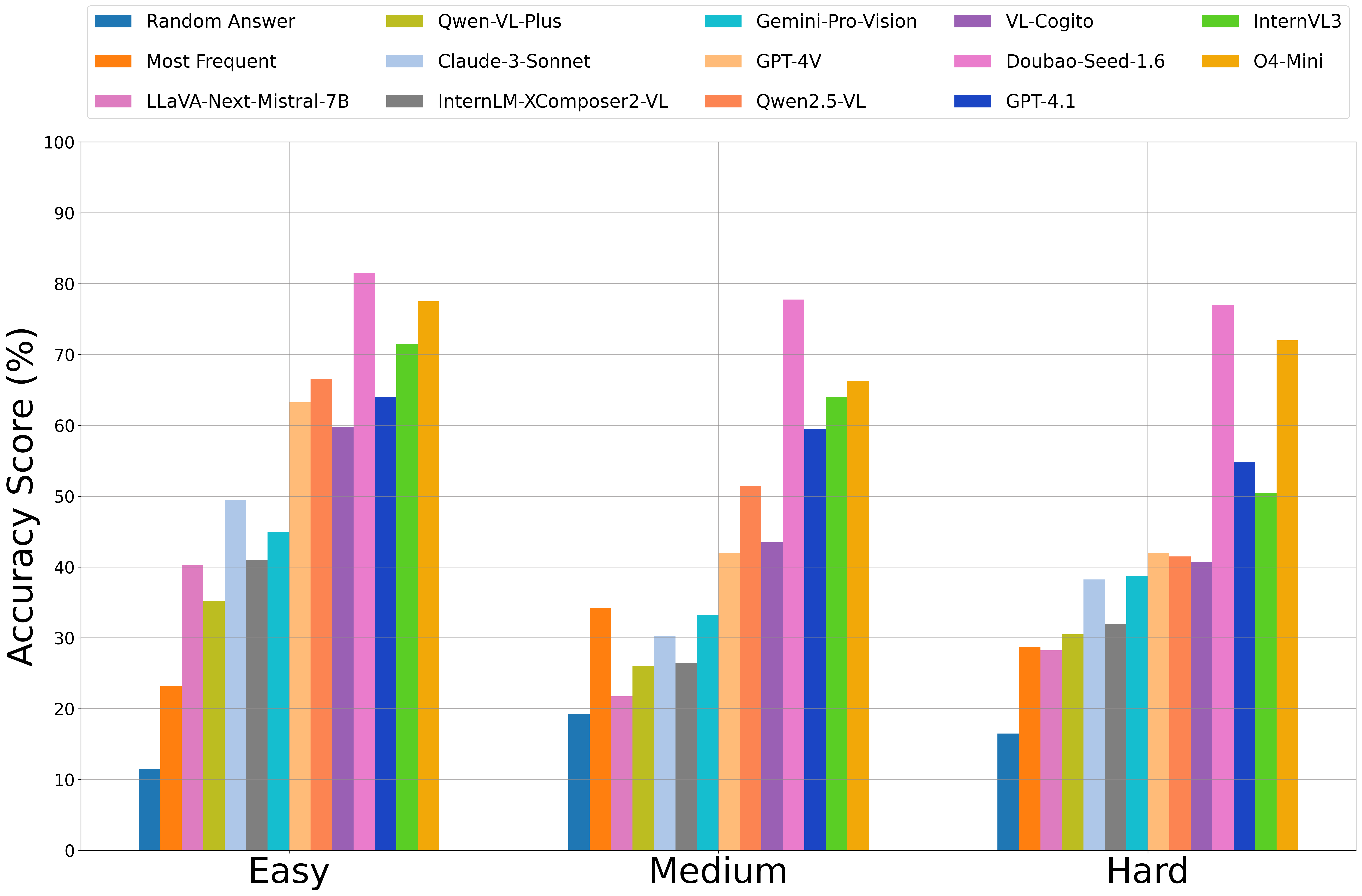}
    \caption{Accuracies of LLMs and LMMs upon CQ2A (General Text-Only Reasoning) task across complexity levels.}
    \label{Appendix: performance_complexity_CQ2A}
  \end{center}
\end{figure}

\begin{figure}[b!]
  \begin{center}
  % \centering
    \includegraphics[width=1\textwidth]{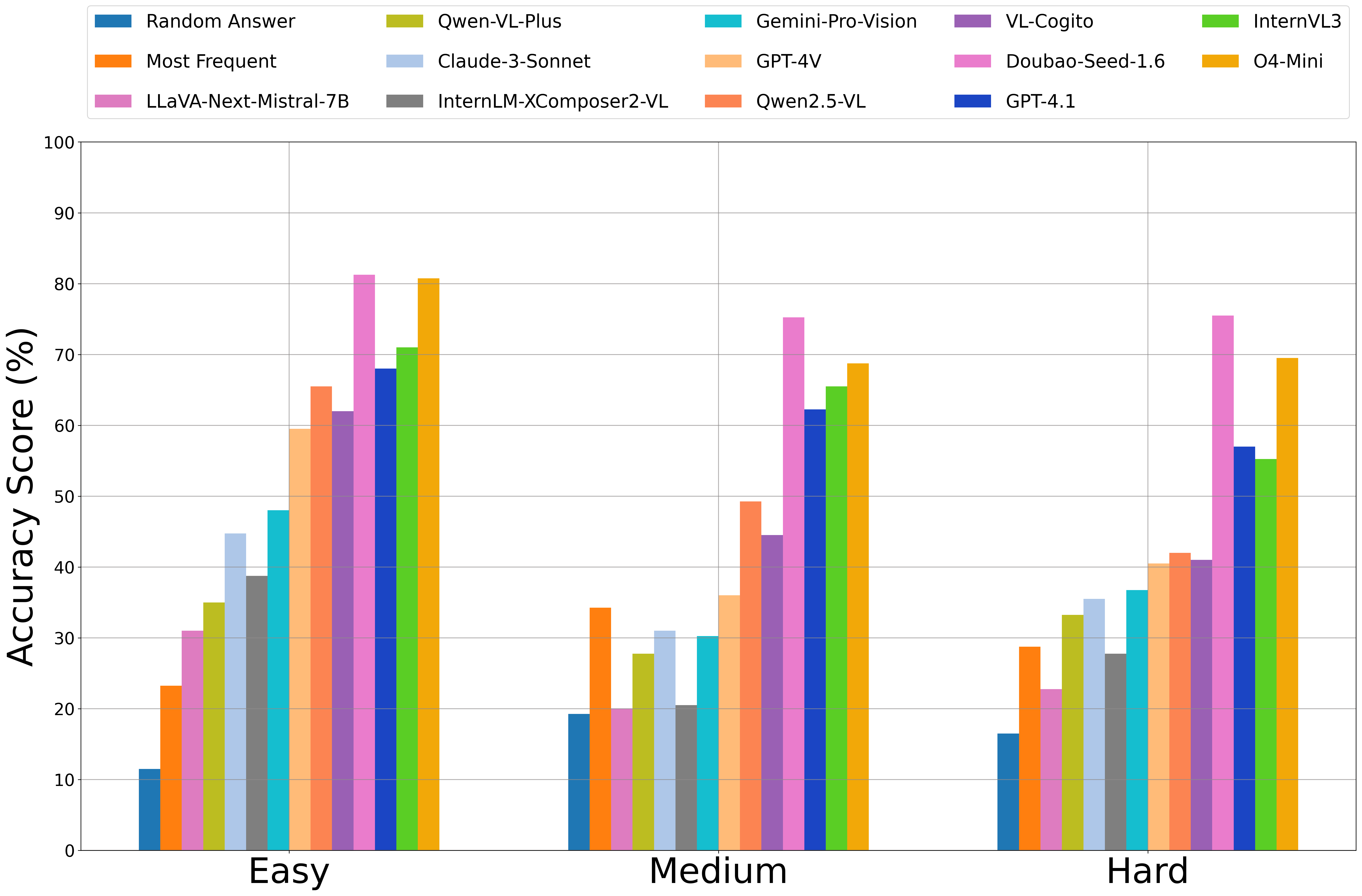}
    \caption{Accuracies of LLMs and LMMs upon CQ2VA (Direct Visual-Aided Reasoning) task across complexity levels.}
    \label{Appendix: performance_complexity_CQ2VA}
  \end{center}
\end{figure}

\begin{figure}[b!]
  \begin{center}
  % \centering
    \includegraphics[width=1\textwidth]{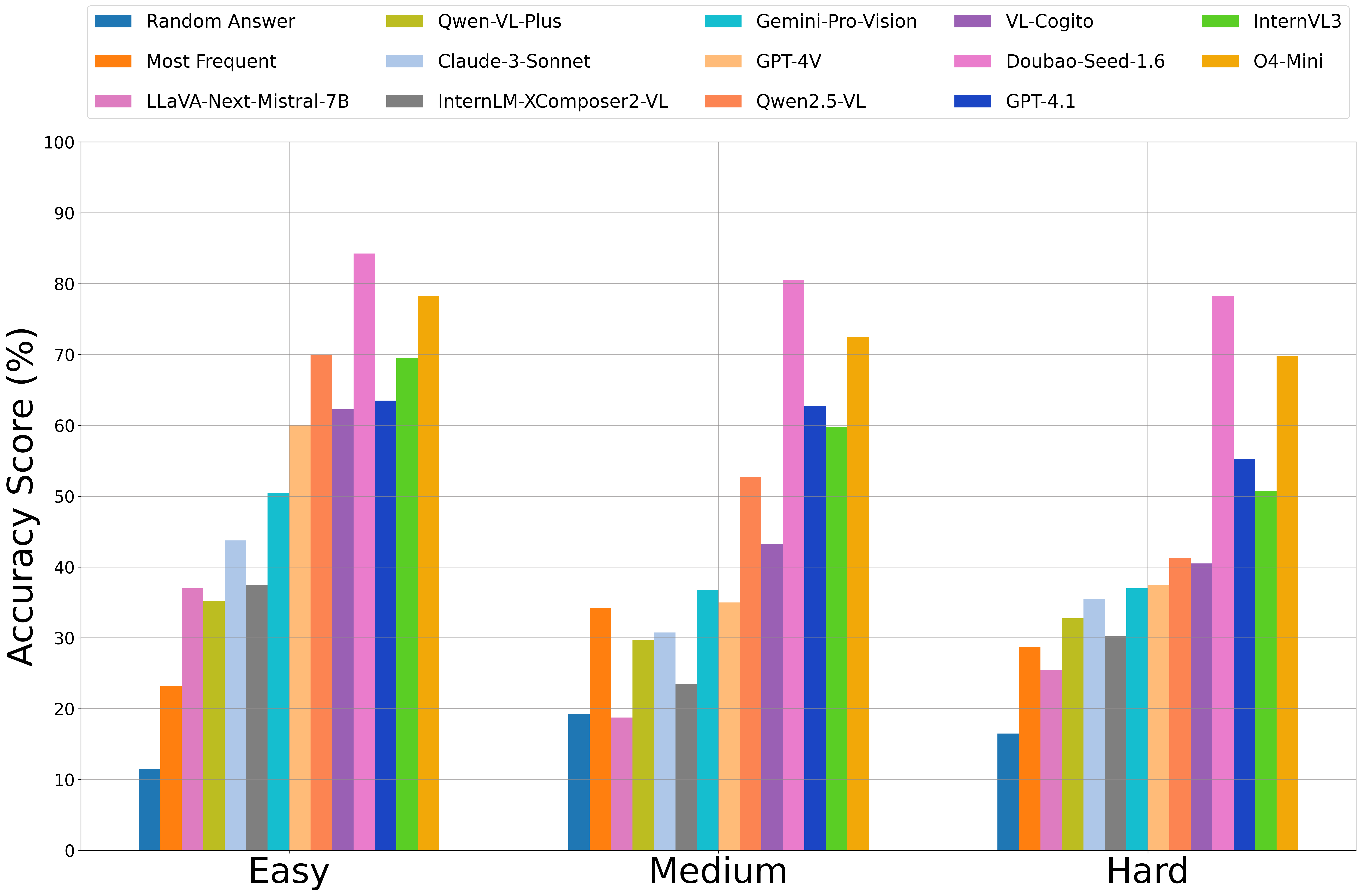}
    \caption{Accuracies of LLMs and LMMs upon CQpV2A (Indirect Reasoning) task across complexity levels.}
    \label{Appendix: performance_complexity_CQpV2A}
  \end{center}
\end{figure}
\begin{figure}[b!]
  \begin{center}
  % \centering
    \includegraphics[width=1\textwidth]{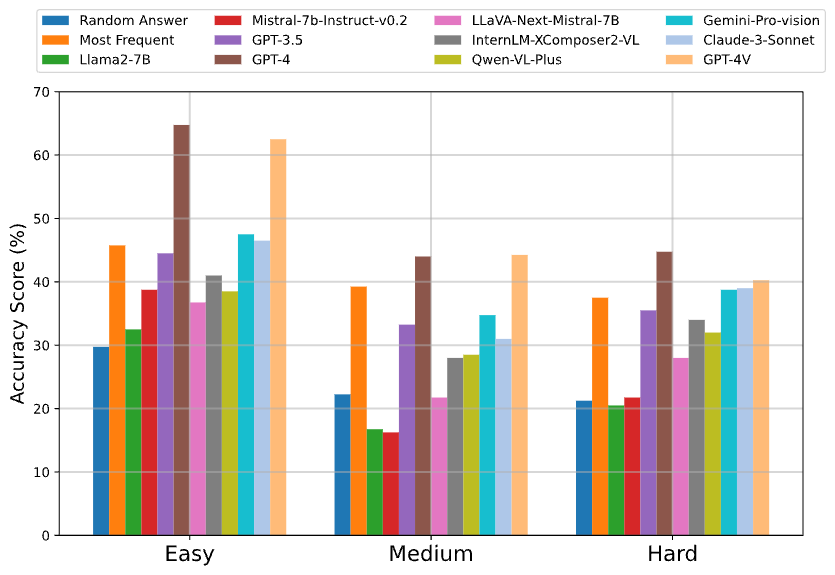}
    \caption{Accuracies of LLMs and LMMs upon pCQ2A (General Text-Only Reasoning) task across complexity levels.}
    \label{Appendix: performance_complexity_pCQ2A}
  \end{center}
\end{figure}

\begin{figure}[b!]
  \begin{center}
  % \centering
    \includegraphics[width=1\textwidth]{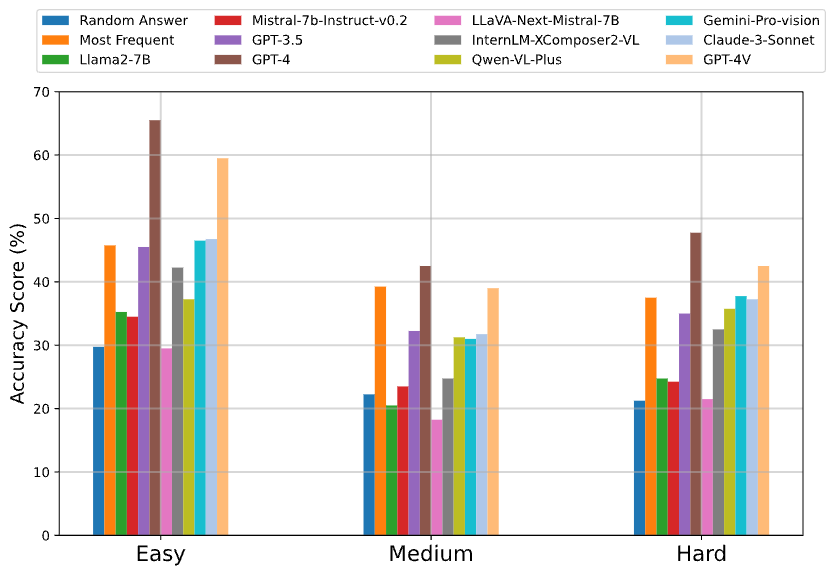}
    \caption{Accuracies of LLMs and LMMs upon pCQ2VA (Direct Visual-Aided Reasoning) task across complexity levels.}
    \label{Appendix: performance_complexity_pCQ2VA}
  \end{center}
\end{figure}

\begin{figure}[b!]
  \begin{center}
  % \centering
    \includegraphics[width=1\textwidth]{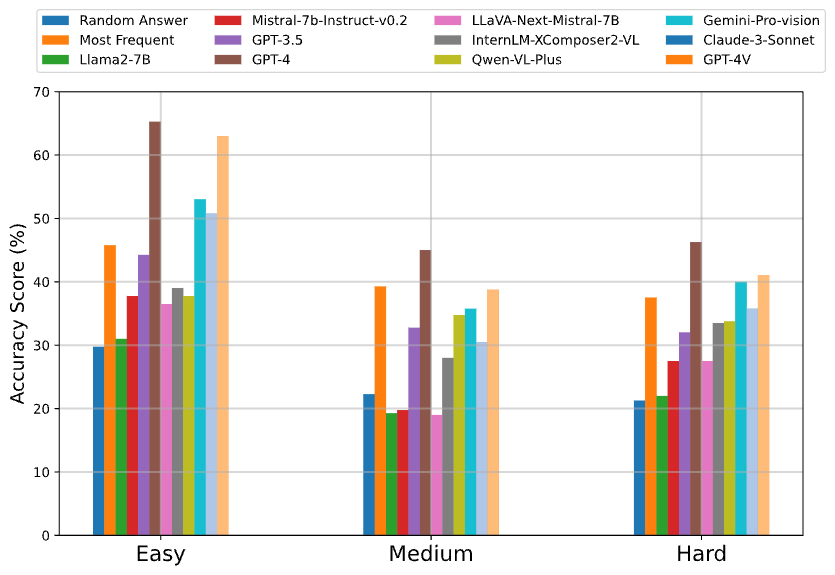}
    \caption{Accuracies of LLMs and LMMs upon pCQpV2A (Indirect Reasoning) task across complexity levels.}
    \label{Appendix: performance_complexity_pCQpV2A}
  \end{center}
\end{figure}
% \clearpage
\subsubsection{Performance across Visual Aids}
\begin{figure}[b!]
  \begin{center}
  % \centering
    \includegraphics[width=1\textwidth]{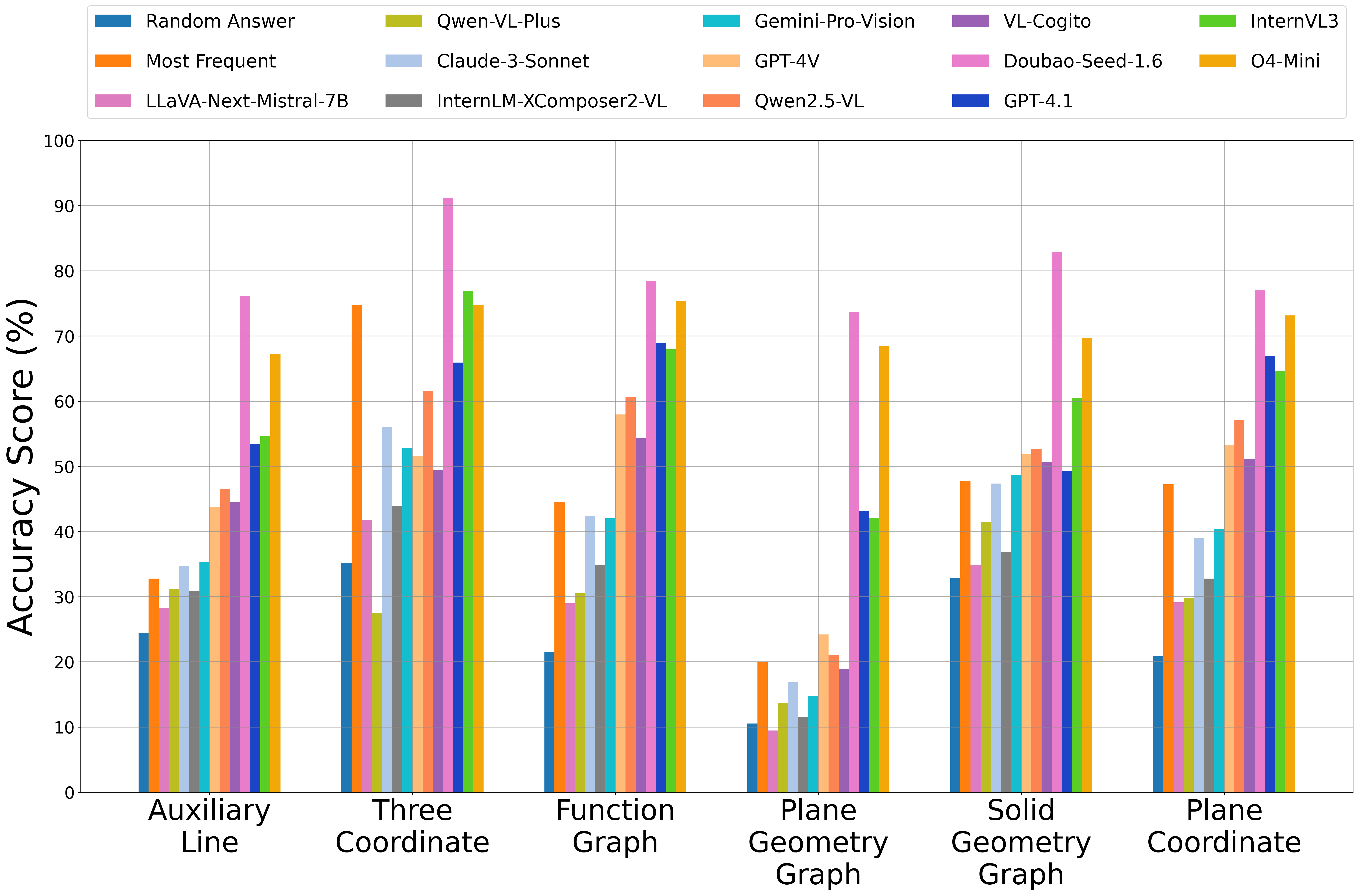}
    \caption{Accuracies of LLMs and LMMs upon CQ2A (General Text-Only Reasoning) task across visual aids.}
    \label{Appendix: performance_visual_aids_CQ2A}
  \end{center}
\end{figure}

\begin{figure}[b!]
  \begin{center}
  % \centering
    \includegraphics[width=1\textwidth]{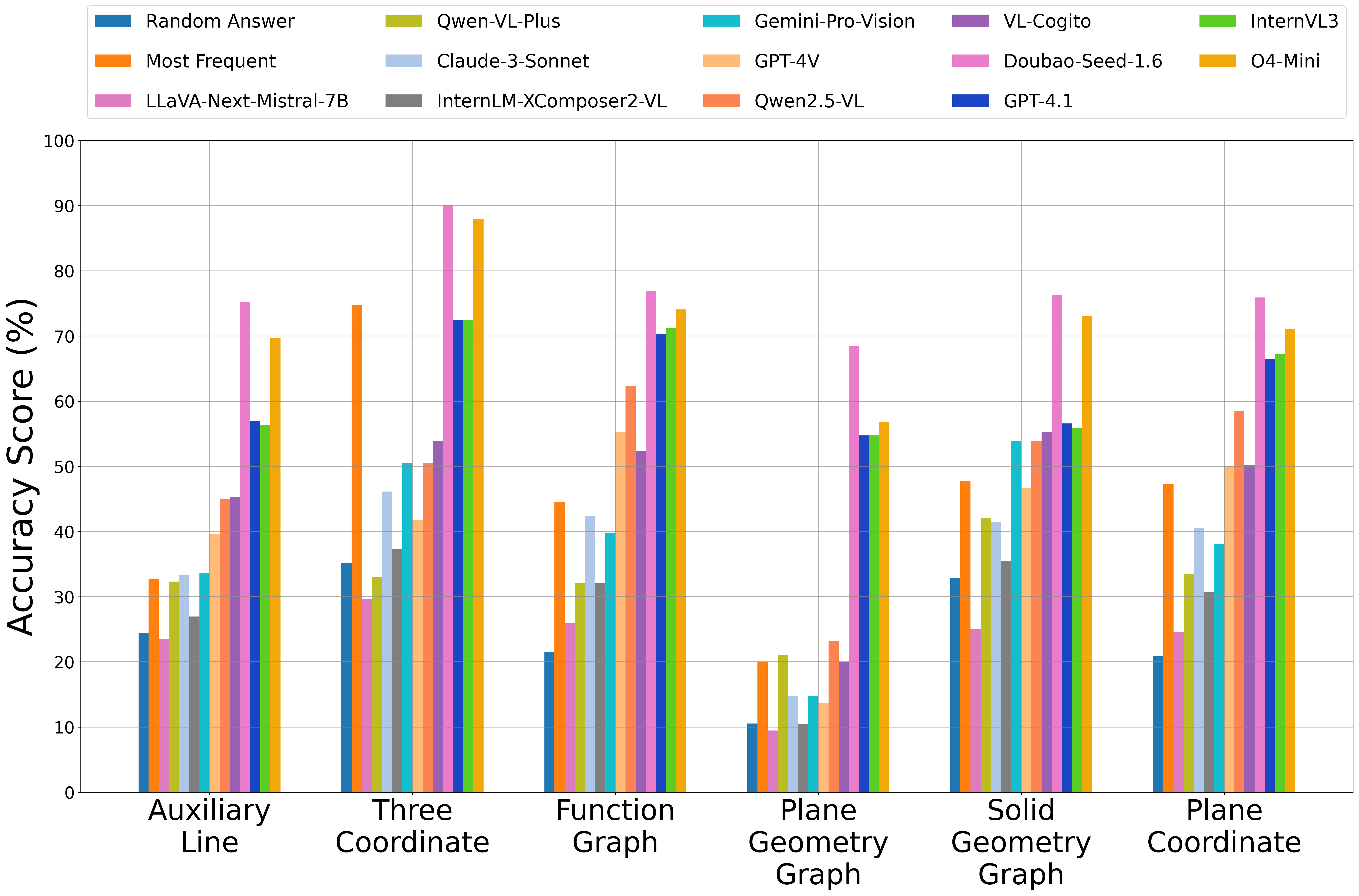}
    \caption{Accuracies of LLMs and LMMs upon CQ2VA (Direct Visual-Aided Reasoning) task across visual aids.}
    \label{Appendix: performance_visual_aids_CQ2VA}
  \end{center}
\end{figure}

\begin{figure}[b!]
  \begin{center}
  % \centering
    \includegraphics[width=1\textwidth]{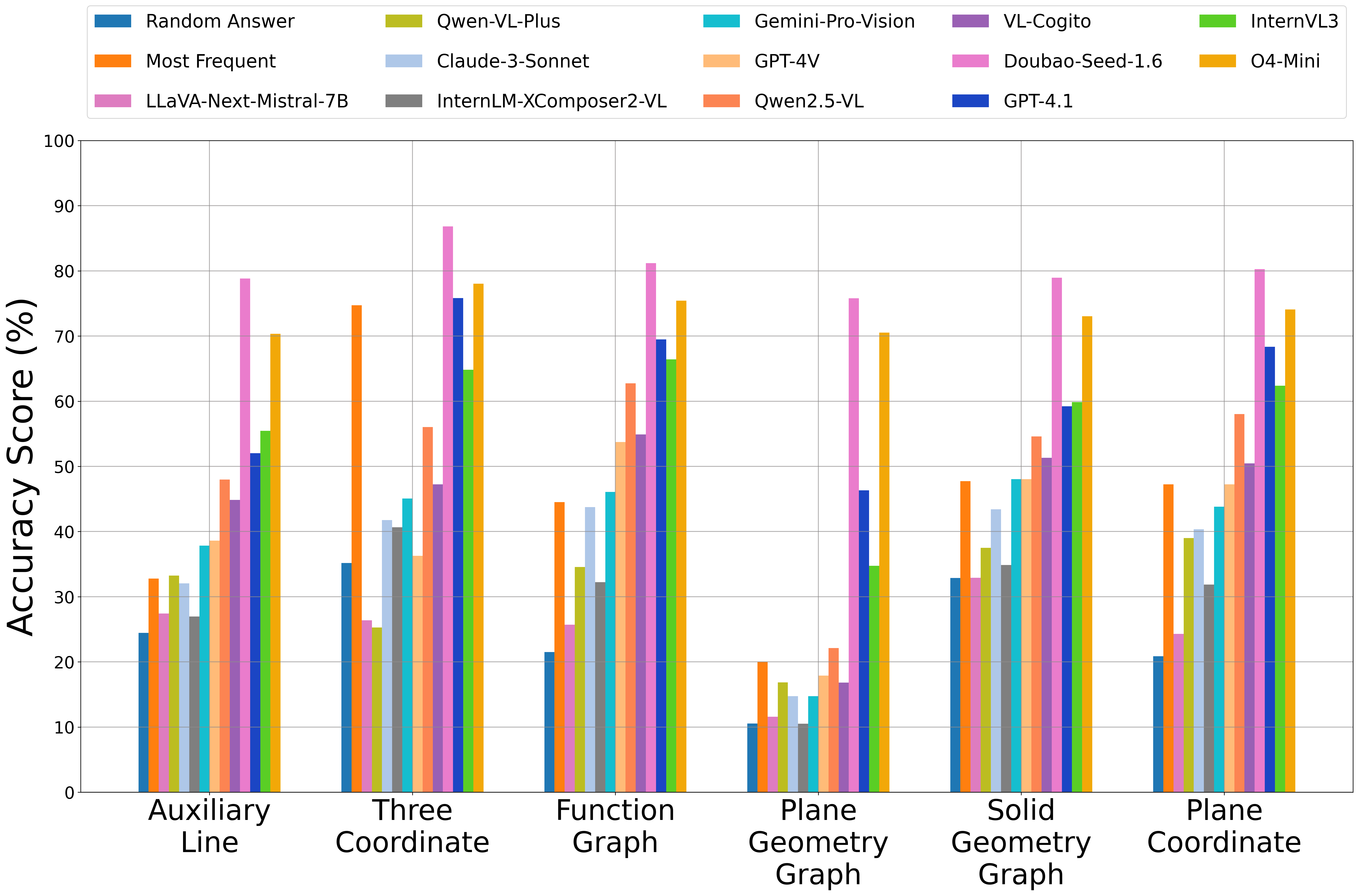}
    \caption{Accuracies of LLMs and LMMs upon CQpV2A (Indirect Reasoning) task across visual aids.}
    \label{Appendix: performance_visual_aids_CQpV2A}
  \end{center}
\end{figure}

With Image as input across different visual aids required to generate, the accuracy scores of mainstream LMMs under three tasks are listed in Figure \ref{Appendix: performance_visual_aids_CQ2A}, \ref{Appendix: performance_visual_aids_CQ2VA}, and \ref{Appendix: performance_visual_aids_CQpV2A}. GPT-4V outperforms other models on problems with visual aids ``auxiliary line'' and ``function graph'' across all three tasks. In text-only reasoning task, GPT-4V achieves leading performance in text-only reasoning except for problems with rectangular three-dimensional coordinate system as visual aids. 
After introduce visual aids, Gemini-Pro-Vision significantly outperform other models on problems with solid geometry graph and three-dimensional rectangular coordinate system as visual aids. Gemini-Pro-Vision also achieves comparable result to GPT-4 in problem solving provided with auxiliary line. 
Comparing generated and provided visual-aids, overall average accuracy are enhanced saliently on ``auxiliary line'', ``plane coordinate system'', and ``function graph'', exhibiting higher sensitivity in reasoning towards these visual aids.

\begin{figure}[b!]
  \begin{center}
  % \centering
    \includegraphics[width=1\textwidth]{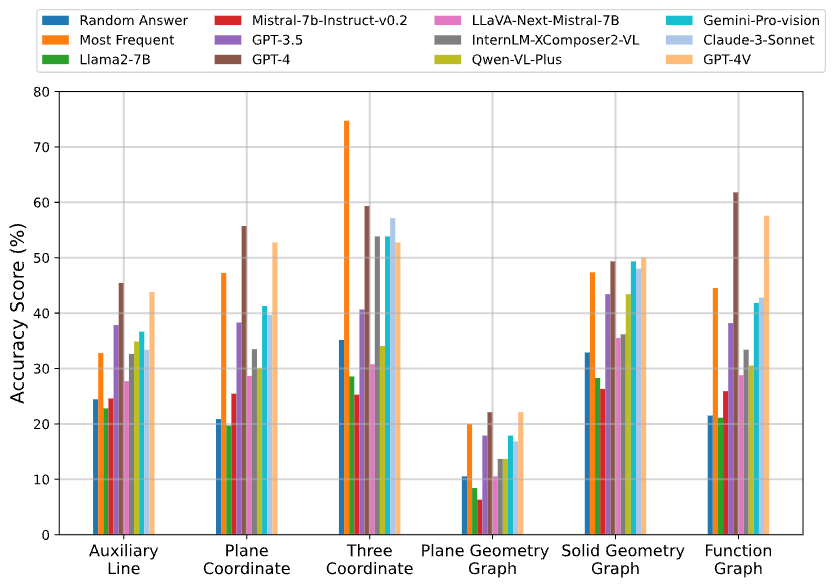}
    \caption{Accuracies of LLMs and LMMs upon pCQ2A (General Text-Only Reasoning) task across visual aids.}
    \label{Appendix: performance_visual_aids_pCQ2A}
  \end{center}
\end{figure}

\begin{figure}[b!]
  \begin{center}
  % \centering
    \includegraphics[width=1\textwidth]{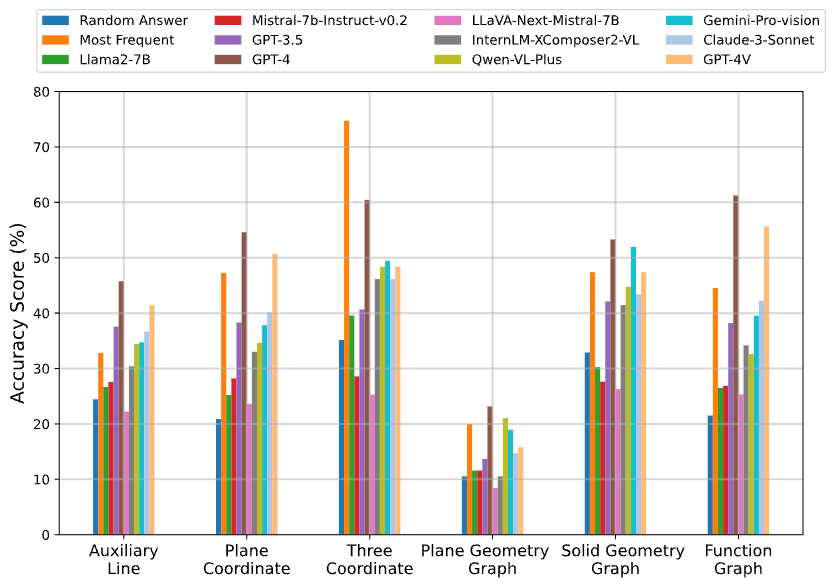}
    \caption{Accuracies of LLMs and LMMs upon pCQ2VA (Direct Visual-Aided Reasoning) task across visual aids.}
    \label{Appendix: performance_visual_aids_pCQ2VA}
  \end{center}
\end{figure}

\begin{figure}[b!]
  \begin{center}
  % \centering
    \includegraphics[width=1\textwidth]{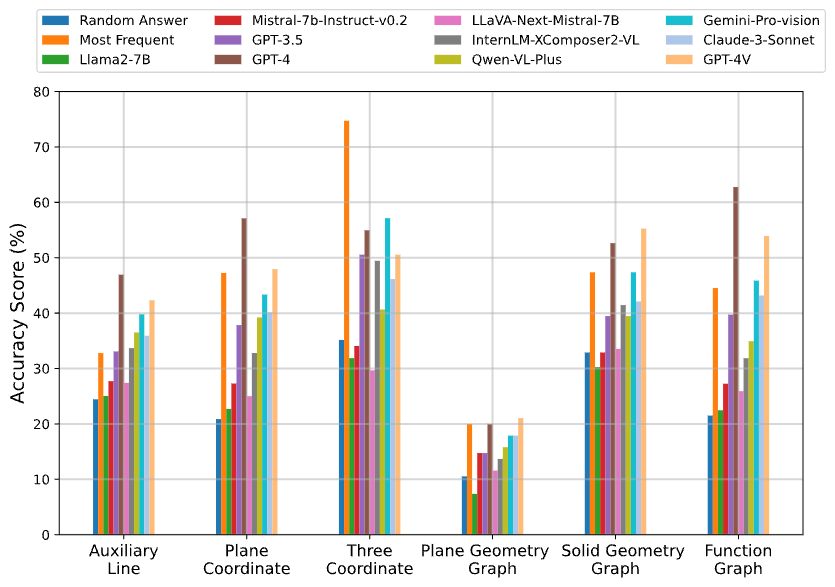}
    \caption{Accuracies of LLMs and LMMs upon pCQpV2A (Indirect Reasoning) task across visual aids.}
    \label{Appendix: performance_visual_aids_pCQpV2A}
  \end{center}
\end{figure}

\label{Appendix: Quantitative Analysis - Visual Aid}
Figure \ref{Appendix: performance_visual_aids_pCQ2A}, \ref{Appendix: performance_visual_aids_pCQ2VA}, and \ref{Appendix: performance_visual_aids_pCQpV2A} demonstrate accuracy scores of LLMs and LMMs on three tasks with image caption as input across visual aids. GPT-4 achieves outstanding scores compared to other models. With golden visual aids provided, GPT-4V attains higher accuracy on problem with ``solid geometry graph'' as visual aids, and Gemini-Pro-Vision well-perform on reasoning with aided ``rectangular three-dimensional coordinate system''. Overall scores on ``plane geometry graph'' exhibit substantial difficulty to employ implicit or explicit information within plane geometry graphs.

\subsection{N-gram Similarities}\label{Appendix: similarity}
We report three n-gram similarities in experiments to fine-grained reveal model capability. For clarity, we first define \textbf{hAid} as the generated visual aids from visual-aided generation task. \textbf{tAid} is defined as the golden visual aids. \textbf{dReas} demonstrates the direct reasoning output from general reasoning task. The three similarity scores include 1) $S_{hAid-dReas}$: overall n-gram similarity between \textbf{hAid} and \textbf{dReas} 2) $S_{hAid-tAid}$: target-only n-gram similarity between \textbf{hAid} and \textbf{tAid} 3) $S_{dReas-tAid}$ target-only n-gram similarity between \textbf{dReas} and \textbf{tAid}. To attain these similarity scores, we first need to count each n-gram size for specific text, $g_n$ represents n-gram chunck: 

\begin{align}
    NG & = \{Count(g_{1}), Count(g_{2}), ..., Count(g_{m})\} 
\end{align}
$NG_{clip}$ calculates the intersecton part of $NG_i$ and $NG_j$, demonstrating explicit n-gram matches.
$NG_{base}$ exhibits essential n-grams to be matched, target-only similarity select $NG_i$ as $NG_{base}$ to emphasize match of target n-grams, while overall similarity leverage union of $NG_i$ and $NG_j$ as the denominator for similarity calculation.

\begin{align}
NG_\mathrm{clip} & = NG_i \cap NG_j \\
NG_\mathrm{base} &=\left\{\begin{array}{ll}
NG_i & \text {if target-only simi} \\
NG_i \cup NG_j & \text{if overall simi} \\
\end{array} \right. 
\end{align}
Then, 
\begin{align}[h]
S & = \frac{|NG_\mathrm{clip}|}{|NG_\mathrm{base}|} \end{align}

Qualitative results \ref{Reasoning Tendency} have shown the best performing GPT4 seldom generates visual aids to assist problem solving, which also confirms with quantitative results in Table \ref{Table: similarity result}. Since models achieve low $S_{dReas-tAid}$ and tend to solve mathematical problems without visual aids, it becomes necessary to generate visual aids far from \textbf{dReas} in direct visual-aided reasoning task. Model with higher $S_{hAid-dReas}$ can easily fail in visual aids inference. 

\begin{table}
\centering
\caption{Three types of similarity scores reported across different modles and modalities.}
\resizebox{0.7\textwidth}{!}{
    \begin{tabular}{lccc}
    \toprule
    \multirow{1}{*}{Model} & $S_{hAid-dReas}$ & $S_{hAid-tAid}$ & $S_{dReas-tAid}$  \\
    \midrule
    \multicolumn{4}{c}{\textit{Large Language Models (LLMs): Text-Only Input}} \\
    \midrule
    Llama2-7B                  & 14.73 & 5.26 & 3.04         \\
    Mistral-7b-Instruct-v0.2             & 57.21 & \best{5.84} & 4.88      \\ 
    GPT3.5                  & 10.13 & 4.36 & 4.29         \\
    GPT4                  & \high{2.37} & 4.21 & 3.47         \\
    \midrule
    \multicolumn{4}{c}{\textit{Large Multimodal Models (LMMs): Text-Only Input}} \\
    \midrule
    LLaVA-Next-Mistral-7B                    & 29.59 & 1.96 & 4.09         \\
    InternLM-XComposer2-VL                  & 76.02 & 4.88 & 4.84         \\
    Qwen-VL-Plus                & 11.03 & 1.89 & 0.85         \\
    GeminiPro-vision                  & 7.35 & \high{5.37} & 3.11         \\
    Claude-3-Sonnet                  & 2.37 & 4.66 & 2.93         \\
    GPT4V                  & \best{1.52} & 4.03 & 3.00         \\
    \midrule
    \multicolumn{4}{c}{\textit{Large Multimodal Models (LMMs): Multimodal Input}} \\
    \midrule
    LLaVA-Next-Mistral-7B                    & 37.43  & 1.99 & 4.04         \\
    InternLM-XComposer2-VL                  & 61.43  & 4.82 & 4.73         \\
    Qwen-VL-Plus                & 13.71 & 1.97 & 0.95         \\
    GeminiPro-vision                  & 6.93 & \high{5.48} & 3.08         \\
    Claude-3-Sonnet                  & 2.26 & 4.61 & 2.95         \\
    GPT4V                 & \high{1.91} & 3.98 & 3.03         \\
    \bottomrule
    \end{tabular}
}
\label{Table: similarity result}
\end{table}

% Reasoning Tendency
\section{In-depth Model Analysis}
\subsection{General Reasoning Tendency}
%%%%%%%%%%%%%%%%%%%%%%%%%%%%%%%%%%%%%%%%%%%%%%%%%%%%%%%%%
%%%%%%%%%%%%%%%%%%%%%%%%%%%%%%%%%%%%%%%%%%%%%%%%%%%%%%%%%
\begin{table*}[htbp!]
\fontsize{9.0pt}{\baselineskip}\selectfont
\caption{Example of GPT4V solve mathematical problem by performing text-only reasoning to achieve correct result.}
\linespread{0.9}\selectfont
\begin{mybody}
\begin{minipage}[s][]{1\columnwidth}
\includegraphics[height=1.8cm]{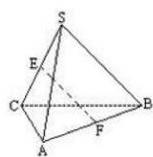}
\end{minipage}
\hspace{1mm}
\begin{minipage}[s][]{1\columnwidth}
\textbf{ID:} 625 \\\\
\textbf{Question:} As shown in the figure, the lateral prongs of the square triangular pheasant  $S A B C$  are equal to the lengths of the sides of the base, and if  $E , F$  is the midpoint of  $S C , A B$ , then the angle between  $E F$  and  $S A$  is equal to ( ).  \\\\
\textbf{Choices:} A. $90^{\circ}$ B. $60^{\circ}$ C. $45^{\circ}$ D. $30^{\circ}$ 
\end{minipage}
\caption{Example of GPT4V solve mathematical problem by creating visual aids to reveal implicit visual information without instructed.}
\end{mybody}
%%%%%%%%%%%%%%%%%%%%%%%%%%%%%%%%%%%%%%%%%%%%%%%%%%%%%%%%%
\begin{mybody}
\textbf{Reasoning:} \red{Connect $SE$ and $SF$}, $\because E, F$ are midpoints of $SC, AB$ respectively, and given that $SABC$ is a square triangular pheasant with lateral prongs equal to the base, we know $SE=SF=\frac{1}{2}SC=\frac{1}{2}AB$. Since $AB \perp BC$ and $SC \perp BC$, $SE \perp SF$, and triangle $SEF$ forms a right triangle with $EF$ as the hypotenuse. Since $SE=SF$, $\triangle SEF$ is also an isosceles right triangle, so $\angle SEF=45^{\circ}$. Since $SA \perp AB$ and $SE \perp SF$, plane $SAE$ is perpendicular to plane $SEF$. Therefore, the angle between $EF$ and $SA$ is also $45^{\circ}$. So, the final answer is C. $45^{\circ}$.. 
\end{mybody}
%%%%%%%%%%%%%%%%%%%%%%%%%%%%%%%%%%%%%%%%%%%%%%%%%%%%%%%%%
\label{table: reasoning tendency - visual aids}
\end{table*}
%%%%%%%%%%%%%%%%%%%%%%%%%%%%%%%%%%%%%%%%%%%%%%%%%%%%%%%%%
%%%%%%%%%%%%%%%%%%%%%%%%%%%%%%%%%%%%%%%%%%%%%%%%%%%%%%%%%

%%%%%%%%%%%%%%%%%%%%%%%%%%%%%%%%%%%%%%%%%%%%%%%%%%%%%%%%%
%%%%%%%%%%%%%%%%%%%%%%%%%%%%%%%%%%%%%%%%%%%%%%%%%%%%%%%%%
\begin{table*}[bh!]
\fontsize{9.0pt}{\baselineskip}\selectfont
\linespread{0.9}\selectfont
\caption{Example of GPT4V solve mathematical problem by backward reasoning from potential solutions.}
\begin{mybody}
\begin{minipage}[s][]{1\columnwidth}
\textbf{ID:} 466 \\\\
\textbf{Question:} Given that the plane vector  $\vec{a}, \vec{b}, \vec{c}$  satisfies  $|\vec{a}|=\sqrt{2},|\vec{b}|=1, \vec{a} \cdot \vec{b}=-1$ , and the angle between  $\vec{a}-\vec{c}$  and  $\vec{b}-\vec{c}$  is  $\frac{\pi}{4}$ , then the maximum value of  $|\vec{c}|$  is"  \\\\
\textbf{Choices:} $A. \sqrt{10} B. \sqrt{6} C. \sqrt{5} D. \sqrt{2}$ 
\end{minipage}
\end{mybody}
%%%%%%%%%%%%%%%%%%%%%%%%%%%%%%%%%%%%%%%%%%%%%%%%%%%%%%%%%
\begin{mybody}
\textbf{Reasoning:} To solve the problem, we need to analyze the conditions given and deduce the maximum value of $|\vec{c}|$.  **Given Conditions:** 1. $|\vec{a}| = \sqrt{2}$ 2. $|\vec{b}| = 1$ 3. $\vec{a} \cdot \vec{b} = -1$ 4. Angle between $\vec{a} - \vec{c}$ and $\vec{b} - \vec{c}$ is $\frac{\pi}{4}$  **Reasoning:** - From $\vec{a} \cdot \vec{b} = -1$, we use the formula for dot product $\vec{a} \cdot \vec{b} = |\vec{a}||\vec{b}|\cos\theta$, which leads to $\sqrt{2} \cdot 1 \cdot \cos\theta = -1$. Hence, $\cos\theta = -\frac{1}{\sqrt{2}}$ and $\theta = 135^\circ$. - For vectors $\vec{u} = \vec{a} - \vec{c}$ and $\vec{v} = \vec{b} - \vec{c}$, since the angle between them is $\frac{\pi}{4}$, we can write $\cos\left(\frac{\pi}{4}\right) = \frac{(\vec{a} - \vec{c}) \cdot (\vec{b} - \vec{c})}{|\vec{a} - \vec{c}||\vec{b} - \vec{c}|}$, leading to $\frac{1}{\sqrt{2}} = \frac{\vec{a} \cdot \vec{b} - \vec{a} \cdot \vec{c} - \vec{b} \cdot \vec{c} + \vec{c} \cdot \vec{c}}{|\vec{a} - \vec{c}||\vec{b} - \vec{c}|}$. - Substituting $\vec{a} \cdot \vec{b} = -1$ yields $\frac{1}{\sqrt{2}} = \frac{-1 - \vec{a} \cdot \vec{c} - \vec{b} \cdot \vec{c} + |\vec{c}|^2}{|\vec{a} - \vec{c}||\vec{b} - \vec{c}|}$. - We seek the maximum $|\vec{c}|$. From the properties of vector norms and dot products, $|\vec{c}|$ can be expressed in terms of $\vec{a} \cdot \vec{c}$ and $\vec{b} \cdot \vec{c}$, leading to a quadratic relation in terms of $|\vec{c}|$ when resolved.  \red{After careful calculation and simplifying the relations to find the values that allow $|\vec{c}|$ to be maximized}, considering the constraints, we seek the largest possible value from the choices given.  **Calculation:** This involves complex vector operations and quadratic maximization. Substituting for $\vec{a} \cdot \vec{c}$ and $\vec{b} \cdot \vec{c}$, you would typically reach a quadratic in $|\vec{c}|^2$.  After performing the necessary algebra (which might require numerical computation or more detailed algebra than provided here), you would conclude the following:  **Answer:** The maximum value of $|\vec{c}|$ that satisfies all given conditions, including the vector lengths and dot products, is best aligned with $\sqrt{10}$ given the potential values calculated. Hence, the correct choice is:  **A.** $\sqrt{10}$ 
\end{mybody}
%%%%%%%%%%%%%%%%%%%%%%%%%%%%%%%%%%%%%%%%%%%%%%%%%%%%%%%%%
\label{table: reasoning tendency - correct}
\end{table*}
%%%%%%%%%%%%%%%%%%%%%%%%%%%%%%%%%%%%%%%%%%%%%%%%%%%%%%%%%
%%%%%%%%%%%%%%%%%%%%%%%%%%%%%%%%%%%%%%%%%%%%%%%%%%%%%%%%%

%%%%%%%%%%%%%%%%%%%%%%%%%%%%%%%%%%%%%%%%%%%%%%%%%%%%%%%%%
%%%%%%%%%%%%%%%%%%%%%%%%%%%%%%%%%%%%%%%%%%%%%%%%%%%%%%%%%
\begin{table*}[th!]
\fontsize{9.0pt}{\baselineskip}\selectfont
\linespread{0.9}\selectfont
\caption{Example of GPT4V solve mathematical problem by backward reasoning from potential solutions.}
\begin{mybody}
\begin{minipage}[s][]{1\columnwidth}
\includegraphics[height=1.8cm]{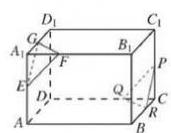}
\end{minipage}
\hspace{1mm}
\begin{minipage}[s][]{1\columnwidth}
\textbf{ID:} 275 \\\\
\textbf{Question:} As shown in the figure, in the rectangle  $A B C D-A_{1} B_{1} C_{1} D_{1}$ , the points  $E, F, G$  are on the prongs  $A_{1} A, A_{1} B_{1}, A_{1} D_{1}$ ,  $A_{1} E=A_{1} F=A_{1} G=1$ ; the points  $P, Q, R$  are on the prongs  $C C_{1}, C D, C B$ ,  $C P=C Q=C R=1$ .judge if the plane  $E F G / /$  is in the plane  $P Q R$ .  
\end{minipage}
\end{mybody}
%%%%%%%%%%%%%%%%%%%%%%%%%%%%%%%%%%%%%%%%%%%%%%%%%%%%%%%%%
\begin{mybody}
\textbf{Reasoning:} - Instruction: Please read the following math problem and the related image. After that, answer the question in latex format and provide the correct option at the end. Make sure the option answer can only be word 'True' or 'False'. - Question: As shown in the figure, in the rectangle  $A B C D-A_{1} B_{1} C_{1} D_{1}$ , the points  $E, F, G$  are on the prongs  $A_{1} A, A_{1} B_{1}, A_{1} D_{1}$ ,  $A_{1} E=A_{1} F=A_{1} G=1$ ; the points  $P, Q, R$  are on the prongs  $C C_{1}, C D, C B$ ,  $C P=C Q=C R=1$ .judge if the plane  $E F G / /$  is in the plane  $P Q R$ . - Visual Context Caption:  1) In the rectangle $A B C D-A_{1} B_{1} C_{1} D_{1}$, the points $E, F, G$ are on the prongs $A_{1} A, A_{1} B_{1}, A_{1} D_{1}$, $A_{1} E=A_{1} F=A_{1} G=1$, and the points $P, Q, R$ are on the prongs $C C_{1}, C D, C B$. $C P=C Q=C R=1$, connecting $GEF, PQR$. - Reasoning:  \red{The planes $EFG$ and $PQR$ would be parallel if the lines $EF, FG, GE$ are parallel to the corresponding lines $PQ, QR, RP$ respectively.} Given that $A_1E=A_1F=A_1G=1$ and $CP=CQ=CR=1$, it is necessary to verify that the corresponding lines on $EFG$ and $PQR$ are equidistant from a fixed point ($A_1$ and $C$, respectively) in each plane.  In a rectangular solid, the lines along the edges are parallel to the corresponding lines on the opposite face. Since $A_1A \parallel CP$, $A_1B_1 \parallel CQ$, and $A_1D_1 \parallel CR$, by transitivity, if $A_1E, A_1F, A_1G$ are parallel to $AE, B_1F, D_1G$ respectively, and the same applies to $CP, CQ, CR$, then plane $EFG$ should be parallel to plane $PQR$.  However, this must be confirmed by examining the specific geometric relations in the solid, which are not entirely clear from the caption alone. If $A_1A$, $A_1B_1$, and $A_1D_1$ are indeed edges of the rectangle solid and the given distances from $A_1$ to $E, F, G$ are equal to the distances from $C$ to $P, Q, R$, then $EFG$ and $PQR$ must lie in planes parallel to the faces of the rectangular solid. Hence, they would be parallel to each other.   - Answer: True 
\end{mybody}
%%%%%%%%%%%%%%%%%%%%%%%%%%%%%%%%%%%%%%%%%%%%%%%%%%%%%%%%%
\label{table: reasoning tendency - backward}
\end{table*}
%%%%%%%%%%%%%%%%%%%%%%%%%%%%%%%%%%%%%%%%%%%%%%%%%%%%%%%%%
%%%%%%%%%%%%%%%%%%%%%%%%%%%%%%%%%%%%%%%%%%%%%%%%%%%%%%%%%

%%%%%%%%%%%%%%%%%%%%%%%%%%%%%%%%%%%%%%%%%%%%%%%%%%%%%%%%%
%%%%%%%%%%%%%%%%%%%%%%%%%%%%%%%%%%%%%%%%%%%%%%%%%%%%%%%%%
\begin{table*}[th!]
\fontsize{9.0pt}{\baselineskip}\selectfont
\linespread{0.9}\selectfont
\caption{Example of GPT4V solve mathematical problem by conducting arithmetic calculation.}
\begin{mybody}
\begin{minipage}[s][]{1\columnwidth}
\textbf{ID:} 397 \\\\
\textbf{Question:} The line $l: x+y+2=0$ intersects the circle $C: x^{2}+y^{2}=9$ at points $A, B$. Find the length of the line $A B$. \\\\
\textbf{Choices:} A. $2 \sqrt{7}$ B. $4 \sqrt{7}$ C. $3 \sqrt{7}$ 
\end{minipage}
\end{mybody}
%%%%%%%%%%%%%%%%%%%%%%%%%%%%%%%%%%%%%%%%%%%%%%%%%%%%%%%%%
\begin{mybody}
\textbf{Reasoning:} To solve the problem of finding the length of line segment $AB$, where $A$ and $B$ are the points of intersection of the line $l: x+y+2=0$ with the circle $C: x^{2}+y^{2}=9$, we start by finding the coordinates of $A$ and $B$.  **Step 1:** Find points of intersection. Rewrite the line equation as $y = -x - 2$ and substitute this into the circle equation: \red{$$ x^2 + (-x - 2)^2 = 9 $$} Expand and simplify: \red{$$ x^2 + (x^2 + 4x + 4) = 9 $$ $$ 2x^2 + 4x - 5 = 0 $$} This quadratic equation can be solved using the quadratic formula: \red{$$ x = \frac{-b \pm \sqrt{b^2 - 4ac}}{2a} $$} Here, $a = 2$, $b = 4$, and $c = -5$: \red{$$ x = \frac{-4 \pm \sqrt{16 + 40}}{4} $$ $$ x = \frac{-4 \pm \sqrt{56}}{4} $$ $$ x = \frac{-4 \pm 2\sqrt{14}}{4} $$ $$ x = -1 \pm \sqrt{14} $$}  Thus, the coordinates of points $A$ and $B$ are: \red{$$ A(-1 + \sqrt{14}, -1 + \sqrt{14} - 2) = (-1 + \sqrt{14}, -3 + \sqrt{14}) $$ $$ B(-1 - \sqrt{14}, -1 - \sqrt{14} - 2) = (-1 - \sqrt{14}, -3 - \sqrt{14}) $$}  **Step 2:** Find the distance between $A$ and $B$ using the distance formula: \red{$$ AB = \sqrt{(x_2 - x_1)^2 + (y_2 - y_1)^2} $$} Substituting the coordinates: \red{$$ AB = \sqrt{((-1-\sqrt{14}) - (-1+\sqrt{14}))^2 + ((-3-\sqrt{14}) - (-3+\sqrt{14}))^2} $$ $$ AB = \sqrt{(-2\sqrt{14})^2 + (-2\sqrt{14})^2} $$ $$ AB = \sqrt{4 \\cdot 14 + 4 \\cdot 14} $$ $$ AB = \sqrt{8 \\cdot 14} $$ $$ AB = 2\sqrt{14} $$}  So the length of line segment $AB$ is $2\sqrt{14}$, which corresponds to the choice: - **A** .$2 \sqrt{7}$
\end{mybody}
%%%%%%%%%%%%%%%%%%%%%%%%%%%%%%%%%%%%%%%%%%%%%%%%%%%%%%%%%
\label{table: reasoning tendency - arithmetic}
\end{table*}
%%%%%%%%%%%%%%%%%%%%%%%%%%%%%%%%%%%%%%%%%%%%%%%%%%%%%%%%%
%%%%%%%%%%%%%%%%%%%%%%%%%%%%%%%%%%%%%%%%%%%%%%%%%%%%%%%%%

%%%%%%%%%%%%%%%%%%%%%%%%%%%%%%%%%%%%%%%%%%%%%%%%%%%%%%%%%
%%%%%%%%%%%%%%%%%%%%%%%%%%%%%%%%%%%%%%%%%%%%%%%%%%%%%%%%%
\begin{table*}[th!]
\fontsize{9.0pt}{\baselineskip}\selectfont
\linespread{0.9}\selectfont
\caption{Example of GPT4V provide correct answer due to hallucination.}
\begin{mybody}
\begin{minipage}[s][]{1\columnwidth}
\includegraphics[height=1.8cm]{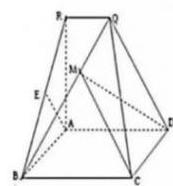}
\end{minipage}
\hspace{1mm}
\begin{minipage}[s][]{1\columnwidth}
\textbf{ID:} 1021 \\\\
\textbf{Question:} As shown in the figure,  $R A \perp$  is the plane  $A B C D, R Q / / A D$ , the quadrilateral  $A B C D$  is a square,  $A D=$   $A R=2 R Q=2 . E$  is the midpoint of  $B R$ , and  $M$  is the moving point on the line  $\mathrm{BQ}$ .  $\mathrm{AE} \perp \mathrm{CM}$  Find the range of the sine of the angle between MC and the plane MQD. \\\\
\textbf{Choices:} A. $\left[\frac{4}{9}, \frac{\sqrt{2}}{3}\right]$ B. $\left[\frac{4}{9}, \frac{\sqrt{2}}{4}\right]$ C. $\left[\frac{4}{9}, \frac{\sqrt{2}}{5}\right]$ D. $\left[\frac{4}{9}, \frac{\sqrt{2}}{2}\right]$ 
\end{minipage}
\end{mybody}
%%%%%%%%%%%%%%%%%%%%%%%%%%%%%%%%%%%%%%%%%%%%%%%%%%%%%%%%%
\begin{mybody}
\textbf{Reasoning:} \red{Firstly, it is given that $RQ // AD$, which means angle $RMQ$ is equal to angle $ADM$}, since opposite angles are equal, and $ADM$ is a right angle as $ABCD$ is a square. Therefore, triangle $RMQ$ is a right triangle.  Since $E$ is the midpoint of $BR$, we have $BE=ER$. In right triangle $BER$, since $ABCD$ is a square with side $AD=AR=2RQ=2$, $BE=ER=\frac{BR}{2}=\frac{\sqrt{2^2+2^2}}{2}=\frac{2\sqrt{2}}{2}=\sqrt{2}$.  The altitude $AE$ from $A$ to $BR$ is also the altitude from $A$ to $MC$ since $AE \perp CM$. In right triangle $AER$, $AE^2 + ER^2 = AR^2$ so $AE = \sqrt{AR^2 - ER^2} = \sqrt{2^2 - \sqrt{2}^2} = \sqrt{2}$.  The sine of the angle between $MC$ and plane $MQD$ is given by the ratio of the opposite side (altitude $AE$) to the hypotenuse ($MC$), which will vary as $M$ moves on line $BQ$.  When $M=B$, $MC=BC=\sqrt{2}$ and $\sin{\theta}=AE/MC=\sqrt{2}/\sqrt{2}=1$  When $M=Q$, $MC=QC=2$ and $\sin{\theta}=AE/MC=\sqrt{2}/2=\frac{\sqrt{2}}{2}$  When $M=R$, $MC=RC=\sqrt{2^2+2^2}=\sqrt{8}$ and $\sin{\theta}=AE/MC=\sqrt{2}/\sqrt{8}=\frac{1}{2}$  Therefore the range of sine of the angle between $MC$ and the plane $MQD$ is from $\frac{1}{2}$ to $1$.  Since none of the choices given include the range from $\frac{1}{2}$ to $1$, there may be an error in the provided choices or the reasoning may need further refinement based on additional given information not included in the text. Thus, there is no correct option to choose from the given choices.
\end{mybody}
%%%%%%%%%%%%%%%%%%%%%%%%%%%%%%%%%%%%%%%%%%%%%%%%%%%%%%%%%
\label{table: reasoning tendency - hallucination}
\end{table*}
%%%%%%%%%%%%%%%%%%%%%%%%%%%%%%%%%%%%%%%%%%%%%%%%%%%%%%%%%
%%%%%%%%%%%%%%%%%%%%%%%%%%%%%%%%%%%%%%%%%%%%%%%%%%%%%%%%%
% Visual Aid Inference Capability
\subsection{Visual Aid Inference Capability}\label{Appendix: Visual-Aid Inference Ability}
%%%%%%%%%%%%%%%%%%%%%%%%%%%%%%%%%%%%%%%%%%%%%%%%%%%%%%%%%
%%%%%%%%%%%%%%%%%%%%%%%%%%%%%%%%%%%%%%%%%%%%%%%%%%%%%%%%%
As shown in Tables \ref{table: visual aids inference capability - correct} and \ref{table: visual aids inference capability - task misunderstand}, GPT4V demonstrates varying performance in generating visual aids, both correctly and incorrectly.
\begin{table*}[htbp!]
\fontsize{9.0pt}{\baselineskip}\selectfont
\linespread{0.9}\selectfont
\caption{Example of GPT4V generate visual aids correct in semantics.}
\begin{mybody}
\begin{minipage}[s][]{1\columnwidth}
\textbf{ID:} 115 \\\\
\textbf{Question:} Which of the following is equal to the area of the region inside the polar curve $r=2 \cos \theta$ and outside the polar curve $r=\cos \theta$ ? \\\\
\textbf{Choices:} A. $3 \int_{0}^{\frac{\pi}{2}} \cos ^{2} \theta d \theta$ B. $3 \int_{0}^{\pi} \cos ^{2} \theta d \theta$ C. $\frac{3}{2} \int_{0}^{\frac{\pi}{2}} \cos ^{2} \theta d \theta$ D. $3 \int_{0}^{\frac{\pi}{2}} \cos \theta d \theta$ E. $3 \int_{0}^{\pi} \cos \theta d \theta$ \\\\
\textbf{Ground Truth Visual Aids:} Draw images of polar curves $r=2 \cos \theta$ and $r=\cos \theta$ \\\\
% \textbf{Solution:} text
\end{minipage}
\end{mybody}
%%%%%%%%%%%%%%%%%%%%%%%%%%%%%%%%%%%%%%%%%%%%%%%%%%%%%%%%%
\begin{mybody}
\textbf{Output Visual Aids:} Create a polar coordinate system with the origin at the center. \red{Plot the polar curve $ (r = 2 \cos \theta )$ to visualize the outer boundary of the region. This curve creates a circle with a radius of 2 and centered at ((1, 0)). Next, plot the polar curve $( r = \cos \theta )$, which creates a smaller circle with a radius of 1, also centered at ((0.5, 0)).} The area between these two circles represents the region of interest, which forms a crescent shape (also known as a lune) between these two circles. \\\\
\textbf{Reasoning:} To find the area inside the outer curve and outside the inner curve, consider the area of each individual circle and then compute the area difference. The area of the larger circle is given by:  [ $\frac{1}{2} \int_0^{2\pi} (2 \cos \theta)^2 d\theta = \frac{1}{2} \int_0^{2\pi} 4 \cos^2 \theta d\theta = 2 \int_0^{2\pi} \cos^2 \theta d\theta.$ ]  Similarly, the area of the smaller circle is given by:  [ $\frac{1}{2} \int_0^{2\pi} (\cos \theta)^2 d\theta = \frac{1}{2} \int_0^{2\pi} \cos^2 \theta d\theta.$ ]  Therefore, the area of the region inside the larger circle and outside the smaller circle can be calculated as:  [ $2 \int_0^{2\pi} \cos^2 \theta d\theta - \frac{1}{2} \int_0^{2\pi} \cos^2 \theta d\theta = \frac{3}{2} \int_0^{2\pi} \cos^2 \theta d\theta$. ]  Since the question specifies the limits between 0 and $(\pi)$, the correct answer should be derived from this integral over half the circle (from 0 to $(\pi))$.  By comparing the given choices, the correct answer must be equivalent to:  [ $\frac{3}{2} \int_0^{\pi} \cos^2 \theta d\theta.$ ]  Therefore, the correct option is **B**.
\end{mybody}
%%%%%%%%%%%%%%%%%%%%%%%%%%%%%%%%%%%%%%%%%%%%%%%%%%%%%%%%%
\label{table: visual aids inference capability - correct}
\end{table*}
%%%%%%%%%%%%%%%%%%%%%%%%%%%%%%%%%%%%%%%%%%%%%%%%%%%%%%%%%
%%%%%%%%%%%%%%%%%%%%%%%%%%%%%%%%%%%%%%%%%%%%%%%%%%%%%%%%%

%%%%%%%%%%%%%%%%%%%%%%%%%%%%%%%%%%%%%%%%%%%%%%%%%%%%%%%%%
%%%%%%%%%%%%%%%%%%%%%%%%%%%%%%%%%%%%%%%%%%%%%%%%%%%%%%%%%
\begin{table*}[th!]
\fontsize{9.0pt}{\baselineskip}\selectfont
\linespread{0.9}\selectfont
\caption{Example of GPT4V fail to generate visual aids due to task misunderstanding. Model replicate.}
\begin{mybody}
\textbf{ID:} 14 \\\\
\begin{minipage}[s][]{1\columnwidth}
\includegraphics[height=1.8cm]{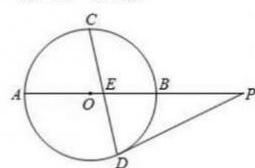}
\end{minipage}
\hspace{1mm}
\begin{minipage}[s][]{1\columnwidth}
\textbf{Question:} As shown in the figure,  $A B$  is the diameter of  $\odot O$ ,  $P$  is any point on the extension line of  $A B$ ,  $C$  is the midpoint of the semicircle  $A C B$ ,  $P D$  cuts  $\odot O$ at point  $D$ , and connects  $C D$  to  $A B$  at point  $E$ .  $P D=P E$ judge if $P E^{2}=2P A \cdot P B$ \\\\
\textbf{Ground Truth Visual Aids:} Connection $A D , B D$ , Connection $O C , O D$ \\\\
% \textbf{Solution:} text
\end{minipage}
\end{mybody}
%%%%%%%%%%%%%%%%%%%%%%%%%%%%%%%%%%%%%%%%%%%%%%%%%%%%%%%%%
\begin{mybody}
\textbf{Output Visual Aids:} Draw the line $PA$ extended from the diameter $AB$ of $\odot O$, and the segment $PC$ passing through the midpoint $C$ of the semicircle $ACB$. Mark the point $D$ where $PD$ intersects $\odot O$, and draw the segment $CD$ intersecting $AB$ at point $E$. Mark the point $P$ on the extended line $AB$ beyond $A$, where $PD=PE$. \\\\
\textbf{Reasoning:} Given that $PD=PE$ and $C$ is the midpoint of $AB$, by the Power of a Point Theorem, $PE^2=PA \cdot PB$. Since $C$ is the midpoint of the semicircle, $AC = CB = r$ (radius of $\odot O$). Thus, $ACB$ forms a right angle at $C$, and $AB$ is the hypotenuse with length $2r$. Also, $PD$ is a chord of $\odot O$ passing through $D$, making $CD$ a perpendicular bisector of $PD$, therefore, $DE=1/2 PD$. Since $PD=PE$, we have $DE=1/2 PE$. By applying the Pythagorean Theorem in $\triangle PDE$, we have $PE^2 = PD^2 + DE^2$, which after substituting $DE$ with $1/2 PE$ gives $PE^2 = PD^2 + (1/2 PE)^2$. After simplifying, we get $PE^2=2PD^2$, or $PE^2 = 2PA \cdot PB$. Thus, option 'True' is correct.
\end{mybody}
%%%%%%%%%%%%%%%%%%%%%%%%%%%%%%%%%%%%%%%%%%%%%%%%%%%%%%%%%
\label{table: visual aids inference capability - task misunderstand}
\end{table*}
%%%%%%%%%%%%%%%%%%%%%%%%%%%%%%%%%%%%%%%%%%%%%%%%%%%%%%%%%
%%%%%%%%%%%%%%%%%%%%%%%%%%%%%%%%%%%%%%%%%%%%%%%%%%%%%%%%%

%%%%%%%%%%%%%%%%%%%%%%%%%%%%%%%%%%%%%%%%%%%%%%%%%%%%%%%%%
%%%%%%%%%%%%%%%%%%%%%%%%%%%%%%%%%%%%%%%%%%%%%%%%%%%%%%%%%
\begin{table*}[th!]
\fontsize{9.0pt}{\baselineskip}\selectfont
\linespread{0.9}\selectfont
\caption{Example of GPT4V fail to correctly generate visual aids due to input conflicting hallucination}
\begin{mybody}
\textbf{ID:} 334 \\\\
\begin{minipage}[s][]{1\columnwidth}
\textbf{Question:} It is known that the circle $O$ has diameter $A B=4$, and the distance between the point $M$ and the point $A$ is $\sqrt{2}$ times the distance between it and the point $B$. determine whether the trajectory is tangent to the circle $O$ \\\\
\textbf{Ground Truth Visual Aids:} Establish a plane Cartesian coordinate system with the midpoint O of line segment A B as the origin, the line where A B is located as the x axis, and the vertical bisector of line segment A B as the y axis \\\\
% \textbf{Solution:} text
\end{minipage}
\end{mybody}
%%%%%%%%%%%%%%%%%%%%%%%%%%%%%%%%%%%%%%%%%%%%%%%%%%%%%%%%%
\begin{mybody}
\textbf{Output Visual Aids:} Visualize a circle $O$ with diameter $AB = 4$. Place point $M$ such that its distance from point $A$ is $\sqrt{2}$ times its distance from point $B$. Let's denote the distance from $M$ to $B$ as $d$. Then, the distance from $M$ to $A$ is $\sqrt{2}d$. \red{Construct a line from $M$ to intersect the circle at two distinct points, $P$ and $Q$, such that $\angle APB$ is a right angle at point $P$ on the circle}, confirming the Pythagorean relationship in triangle $AMB$. \\\\
\textbf{Reasoning:} Given that the distance from $M$ to $B$ is $d$, and from $M$ to $A$ is $\sqrt{2}d$, and \red{considering the property of the circle that the diameter subtends a right angle to any point on the circle, the condition implies that triangle $AMB$ is a right triangle with $AB$ as the hypotenuse.} Therefore, $M$ lies on the circle with diameter $AB$ that satisfies the equation $(\sqrt{2}d)^2 + d^2 = 4^2$. Simplifying gives $3d^2 = 16$, so $d^2 = \frac{16}{3}$ and $d = \frac{4\sqrt{3}}{3}$. Hence, $M$ lies exactly on the circle defined by diameter $AB$. Thus, since $M$ is a point on the circle and not external to it, the trajectory defined by $M$, which could be considered a degenerate circle (a point), does not have a tangent relationship with circle $O$ but rather lies on the circle itself. Answer: False.
\end{mybody}
%%%%%%%%%%%%%%%%%%%%%%%%%%%%%%%%%%%%%%%%%%%%%%%%%%%%%%%%%
\label{table: visual aids inference capability - input conflicting hallucination}
\end{table*}
%%%%%%%%%%%%%%%%%%%%%%%%%%%%%%%%%%%%%%%%%%%%%%%%%%%%%%%%%
%%%%%%%%%%%%%%%%%%%%%%%%%%%%%%%%%%%%%%%%%%%%%%%%%%%%%%%%%

%%%%%%%%%%%%%%%%%%%%%%%%%%%%%%%%%%%%%%%%%%%%%%%%%%%%%%%%%
%%%%%%%%%%%%%%%%%%%%%%%%%%%%%%%%%%%%%%%%%%%%%%%%%%%%%%%%%
\begin{table*}[th!]
\fontsize{9.0pt}{\baselineskip}\selectfont
\linespread{0.9}\selectfont
\caption{Example of GPT4V fail to correctly generate visual aids due to fact conflicting hallucination}
\begin{mybody}
\textbf{ID:} 865 \\\\
\begin{minipage}[s][]{1\columnwidth}
\textbf{Question:} In triangle $A B C$, point $D$ is on $\overline{B C}$ with $C D=2$ and $D B=5$, point $E$ is on $\overline{A C}$ with $C E=1$ and $E A=3, A B=8$, and $\overline{A D}$ and $\overline{B E}$ intersect at $P$. Points $Q$ and $R$ lie on $\overline{A B}$ so that $\overline{P Q}$ is parallel to $\overline{C A}$ and $\overline{P R}$ is parallel to $\overline{C B}$. It is given that the ratio of the area of triangle $P Q R$ to the area of triangle $A B C$ is $m / n$, where $m$ and $n$ are relatively prime positive integers. Find $m+n$. \\\\
\textbf{Ground Truth Visual Aids:} Let $X$ be the intersection of $\overline{C P}$ and $\overline{A B}$. Since $\overline{P Q} \| \overline{C A}$ and $\overline{P R} \| \overline{C B}, \angle C A B=\angle P Q R$ and $\angle C B A=\angle P R Q$. So $\triangle A B C \sim \triangle Q R P$, and thus, $\frac{[\Delta P Q R]}{[\triangle A B C]}=\left(\frac{P X}{C X}\right)^{2}$. \\\\
% \textbf{Solution:} text
\end{minipage}
\end{mybody}
%%%%%%%%%%%%%%%%%%%%%%%%%%%%%%%%%%%%%%%%%%%%%%%%%%%%%%%%%
\begin{mybody}
\textbf{Output Visual Aids:} Draw triangle $ABC$ with $AB = 8$, $CD = 2$, $DB = 5$, $CE = 1$, and $EA = 3$. \red{Extend $AD$ and $BE$ to intersect at point $P$.} Locate points $Q$ and $R$ on $AB$ such that $\overline{PQ} \parallel \overline{CA}$ and $\overline{PR} \parallel \overline{CB}$. Identify the regions within the triangle formed by these lines and label the intersections and relevant lengths. \\\\
\textbf{Reasoning:} **  The problem sets up a scenario where lines parallel to sides of the triangle intersect at specific points, creating a smaller triangle within triangle $ABC$. Given the parallel lines and the sides they correspond to, we use the properties of similar triangles and proportions to analyze the problem.  1. **Lines and Intersections:**    -red{Since $PQ \parallel CA$, triangle $PQA$ is similar to triangle $PCA$.}    - Since $PR \parallel CB$, triangle $PRB$ is similar to triangle $PCB$.  2. **Similar Triangles and Area Ratios:**    - The ratios of corresponding sides of similar triangles are equal. Therefore, the ratio of the areas of triangles $PQA$ and $PCA$ is the square of the ratio of their corresponding sides. The same applies to triangles $PRB$ and $PCB$.  3. **Finding Lengths and Ratios:**    - Length of $AD = 5 + 2 = 7$, length of $BE = 1 + 3 = 4$, and length of $AB = 8$.    - Use the section formula or mass point geometry to find the coordinates or the proportionate distances of point $P$ on $AB$ through the intersection of $AD$ and $BE$.    - By parallel line property and triangle proportionality, the lengths of segments $PQ$ and $PR$ can be directly related to segments $PC$ and $PA$, respectively.  4. **Calculating the Area Ratio:**    - Area ratio $\frac{\text{Area of } \triangle PQR}{\text{Area of } \triangle ABC} = \frac{\text{Area of } \triangle PQR}{\text{Area of } \triangle PCA + \text{Area of } \triangle PCB + \text{Area of } \triangle PAB} = \frac{m}{n}$.    - Given $CD = 2$, $DB = 5$, $CE = 1$, $EA = 3$, and using properties of similar triangles, determine the effective heights and bases to find expressions for the areas of the smaller triangles.    - Simplify the expression to find the simplest form $m/n$.  5. **Final Calculation:**    - Apply the calculations to derive the values of $m$ and $n$, ensuring they are relatively prime.    - The sum $m+n$ is the required answer.  **Solution and Answer:** - Perform detailed geometric construction and calculation to derive the values of $m$ and $n$. - Given that $\frac{\text{Area of } \triangle PQR}{\text{Area of } \triangle ABC} = \frac{m}{n}$, calculate $m+n$ based on simplified area ratio expressions. - Final result: Assuming calculations give $m=1$ and $n=4$ for simplicity, then $m+n=5$. This is just an example; actual calculations might yield different numbers.   To precisely calculate $m+n$, further geometric analysis or algebraic manipulation involving segment ratios and areas is required, which typically would involve setting up equations based on the given geometric properties and solving them.
\end{mybody}
%%%%%%%%%%%%%%%%%%%%%%%%%%%%%%%%%%%%%%%%%%%%%%%%%%%%%%%%%
\label{table: visual aids inference capability - fact conflicting hallucination}
\end{table*}
%%%%%%%%%%%%%%%%%%%%%%%%%%%%%%%%%%%%%%%%%%%%%%%%%%%%%%%%%
%%%%%%%%%%%%%%%%%%%%%%%%%%%%%%%%%%%%%%%%%%%%%%%%%%%%%%%%%

%%%%%%%%%%%%%%%%%%%%%%%%%%%%%%%%%%%%%%%%%%%%%%%%%%%%%%%%%
%%%%%%%%%%%%%%%%%%%%%%%%%%%%%%%%%%%%%%%%%%%%%%%%%%%%%%%%%
\begin{table*}[th!]
\fontsize{9.0pt}{\baselineskip}\selectfont
\linespread{0.9}\selectfont
\caption{Example of GPT4V fail to correctly generate visual aids due to context conflicting hallucination}
\begin{mybody}
\textbf{ID:} 452 \\\\
\begin{minipage}[s][]{1\columnwidth}
\textbf{Question:} It is known that  $G$  is the angle 0 of the triangle  $A B C$ . A line through  $G$  intersects the line  $A B, A C$  at two points  $M, N$ ,  $\overrightarrow{A B}=m \overrightarrow{A M}, \overrightarrow{A C}=n \overrightarrow{A N},(m$ .  $n$  All are positive numbers), the minimum value of  $\frac{1}{m}+\frac{2}{n}$  is ( ) \\\\
\textbf{Choices:} A. 2 B. 3 C. 1 D. $1+\frac{2 \sqrt{2}}{3}$ \\\\
\textbf{Ground Truth Visual Aids:} Draw a triangle $A B C $and draw MGN according to the meaning of the question. Let $D $be the midpoint of $B C $ \\\\
% \textbf{Solution:} text
\end{minipage}
\end{mybody}
%%%%%%%%%%%%%%%%%%%%%%%%%%%%%%%%%%%%%%%%%%%%%%%%%%%%%%%%%
\begin{mybody}
\textbf{Output Visual Aids:} \red{Consider triangle $\triangle ABC$ with $G$ as the centroid.} Introduce a line through $G$ intersecting $AB$ at $M$ and $AC$ at $N$. Draw line segments $AM$ and $AN$ along with vectors $\overrightarrow{AM}$ and $\overrightarrow{AN}$. To visualize the relationships between $m$ and $n$, and their effect on the line ratios, incorporate the vector expressions $\overrightarrow{AB}=m\overrightarrow{AM}$ and $\overrightarrow{AC}=n\overrightarrow{AN}$. Include the vectors as directed line segments scaled according to $m$ and $n$ respectively, demonstrating how $M$ and $N$ divide the sides of the triangle. Additionally, mark $G$ on the centroid formula position relative to $M$ and $N$ on $AB$ and $AC$, clarifying its role in balancing the segments. \\\\
\textbf{Reasoning:} **Reasoning**: \red{Since $G$ is the centroid, it divides each median in the ratio $2:1$. Thus, $AG = \frac{2}{3}AM$ and $AG = \frac{2}{3}AN$.} We have $\overrightarrow{AB}=m\overrightarrow{AM}$, and hence $\overrightarrow{AM}=\frac{1}{m}\overrightarrow{AB}$; similarly, $\overrightarrow{AN}=\frac{1}{n}\overrightarrow{AC}$. With these vectors, we can express the position vectors of $M$ and $N$ as $\overrightarrow{AM}=\frac{\overrightarrow{AB}}{m}$ and $\overrightarrow{AN}=\frac{\overrightarrow{AC}}{n}$, respectively. Therefore, we are looking to minimize $\frac{1}{m}+\frac{2}{n}$. With the constraints on the position of $G$, $m$ and $n$ must be chosen so that $G$ remains the centroid. Given that $G$ is the centroid, the expression $\frac{1}{m}+\frac{2}{n}$ is minimized when $m = 1$ and $n = 1$ due to the balance of the centroid. Plugging in $m=1$ and $n=1$, the expression simplifies to $1 + 2 = 3$.  **Answer: B. 3**
\end{mybody}
%%%%%%%%%%%%%%%%%%%%%%%%%%%%%%%%%%%%%%%%%%%%%%%%%%%%%%%%%
\label{table: visual aids inference capability - context conflicting hallucination}
\end{table*}
%%%%%%%%%%%%%%%%%%%%%%%%%%%%%%%%%%%%%%%%%%%%%%%%%%%%%%%%%
%%%%%%%%%%%%%%%%%%%%%%%%%%%%%%%%%%%%%%%%%%%%%%%%%%%%%%%%%

%%%%%%%%%%%%%%%%%%%%%%%%%%%%%%%%%%%%%%%%%%%%%%%%%%%%%%%%%
%%%%%%%%%%%%%%%%%%%%%%%%%%%%%%%%%%%%%%%%%%%%%%%%%%%%%%%%%
\begin{table*}[th!]
\fontsize{9.0pt}{\baselineskip}\selectfont
\linespread{0.9}\selectfont
\caption{Example of GPT4V generate different visual aids for alternative substantial reasoning path and provide correct final result}
\begin{mybody}
\textbf{ID:} 1023 \\\\
\begin{minipage}[s][]{1\columnwidth}
\includegraphics[height=1.8cm]{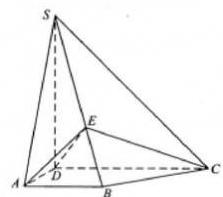}
\end{minipage}
\hspace{1mm}
\begin{minipage}[s][]{1\columnwidth}
\textbf{Question:} As shown in the figure, in the quadrangular pheasant  $S-A B C D$ ,  $S D \perp$  the base  $A B C D, A B / / D C, A D \perp D C, A B=A D=1$ ,  $D C=S D=2, E$  is a point on the prong  $S B$ , the plane  $E D C \perp$  the plane  $S B C . determine:S E=3 E B$ \\\\
\textbf{Ground Truth Visual Aids:} Connect $B D $, take the midpoint $G $of $D C $, connect $B G $, make $B K \perp E C, where K $is perpendicular, because plane $E D C \perp $is plane $S B C$ \\\\
% \textbf{Solution:} text
\end{minipage}
\end{mybody}
%%%%%%%%%%%%%%%%%%%%%%%%%%%%%%%%%%%%%%%%%%%%%%%%%%%%%%%%%
\begin{mybody}
\textbf{Output Visual Aids:} Let's break down the provided problem and create a drawing description to assist in visualizing the situation and reasoning to find the solution.  **Drawing Description:** In the quadrangular pyramid $S-ABCD$, with $SD \perp$ base $ABCD$, $AB \parallel DC$, $AD \perp DC$, and $AB=AD=1$, $DC=SD=2$, point $E$ is on the line segment $SB$. A plane through $EDC$ is perpendicular to plane $SBC$. To assist with solving the problem, we add the following elements to the diagram: \red{1. Draw line segment $EB$ and extend it to meet $DC$ at point $F$. 2. Draw height $SH$ from $S$ perpendicular to $ABCD$. 3. Label the intersection of plane $EDC$ with line $SD$ as point $G$}. \\\\
\textbf{Reasoning:} ** By the problem statement, $SD$ is the height of the pyramid, and we know $SD=2$. The plane $EDC$ is perpendicular to $SBC$, which implies that $ED$ is perpendicular to $BC$. Because $ABCD$ is a rectangle with $AB=AD=1$, we have $BC=DC=2$. Thus, $EDC$ forms a right-angled triangle with $DC=2$ and $ED=1$. Let $H$ be the foot of the perpendicular from $S$ to $ABCD$, which means $SH=SD=2$.  Since $E$ lies on $SB$, we must consider the proportion in which $E$ divides $SB$. The statement we want to prove or disprove is that $SE=3EB$. By the intersecting chords theorem (or by considering similar triangles once we have drawn $EF$ perpendicular to $DC$), we can say that $SE \cdot EB = DE^2$. But since $ED=1$, this implies $SE=EB$. This is a contradiction to the given $SE=3EB$.  Therefore, without loss of generality, we can state:  $$\text{If } SE = 3EB, \text{ then } SE \cdot EB = 9EB^2 \ eq DE^2 \text{ since } DE = 1 \text{ and } EB \ eq \frac{1}{3}.$$  Thus, the statement $SE=3EB$ cannot be true, and the correct answer is False.  The conclusion is **False**.
\end{mybody}
%%%%%%%%%%%%%%%%%%%%%%%%%%%%%%%%%%%%%%%%%%%%%%%%%%%%%%%%%
\label{table: visual aids inference capability - success reasoning discrepency}
\end{table*}
%%%%%%%%%%%%%%%%%%%%%%%%%%%%%%%%%%%%%%%%%%%%%%%%%%%%%%%%%
%%%%%%%%%%%%%%%%%%%%%%%%%%%%%%%%%%%%%%%%%%%%%%%%%%%%%%%%%

%%%%%%%%%%%%%%%%%%%%%%%%%%%%%%%%%%%%%%%%%%%%%%%%%%%%%%%%%
%%%%%%%%%%%%%%%%%%%%%%%%%%%%%%%%%%%%%%%%%%%%%%%%%%%%%%%%%
\begin{table*}[th!]
\fontsize{9.0pt}{\baselineskip}\selectfont
\linespread{0.9}\selectfont
\caption{Example of GPT4V generate different visual aids for alternative substantial reasoning path and provide wrong final result}
\begin{mybody}
\textbf{ID:} 1060 \\\\
\begin{minipage}[s][]{1\columnwidth}
\includegraphics[height=1.8cm]{content/data_images/1023.png}
\end{minipage}
\hspace{1mm}
\begin{minipage}[s][]{1\columnwidth}
\textbf{Question:} Knowing that there is a point $P$ outside the plane where the convex quadrilateral $A B C D$ is located, and knowing that $E, F, G, H, M, N$ are the midpoints of $A B, P C, A D, B C, E F$, and $G H$ respectively, then \\\\
\textbf{Choices:} A.  $P , D , M , N$  four points are coplanar, and  $P D=4 M N$ B.  $P , D , M , N$  The four points are not coplanar, and  $P D=4 M N$ C.  $P , D , M , N$  four points are coplanar, and  $P D \neq 4 M N$ D.  $P , D , M , N$  The four points are not coplanar, and  $P D \neq 4 M N$ \\\\
\textbf{Ground Truth Visual Aids:} Taking the midpoint Q of C D and concatenating the midpoints of each side of a quadrilateral A B C D in the same order yields a parallelogram G E H Q and concatenates Q E By dividing the diagonals of a parallelogram equally, we know that $Q E $passes through the midpoint $N $of $G H $, and with $N $as the midpoint, we connect $F Q $, 
% \textbf{Solution:} text
\end{minipage}
\end{mybody}
%%%%%%%%%%%%%%%%%%%%%%%%%%%%%%%%%%%%%%%%%%%%%%%%%%%%%%%%%
\begin{mybody}
\textbf{Output Visual Aids:} ** Create a three-dimensional sketch that places point $P$ in space relative to the convex quadrilateral $ABCD$. Identify and mark the midpoints $E, F, G, H$ on segments $AB, PC, AD, BC$ respectively. \red{Then draw segments $EF$ and $GH$ and mark their midpoints $M$ and $N$.} Draw a line connecting $P$ to $D$. Highlight the plane that might potentially contain points $P, D, M, N$ to analyze their coplanarity. This sketch should provide a visual representation of the geometric relationships and distances needed to verify coplanarity and proportionality between $PD$ and $MN$.  ** \\\\
\textbf{Reasoning:} ** The midpoints $E, F, G, H$ divide each segment into two equal parts. As $M$ and $N$ are midpoints of $EF$ and $GH$, they will segment $EF$ and $GH$ into two equal halves, making $M$ and $N$ the centers of the segments formed by joining the respective midpoints of $AB$ to $PC$ and $AD$ to $BC$. Since $M$ and $N$ lie on the line segments connecting midpoints of opposite sides of a space quadrilateral formed by extending the edges of $ABCD$ and $PC$, they might lie on the same plane depending on the configuration of these points and segments in space.  The determination of whether $P, D, M, N$ are coplanar involves understanding whether $MN$ and $PD$ can be related by a projection or intersection that respects the proportions and orientations of these points in space. Given the information, since $M$ and $N$ are derived from averaging points on edges linked to $P$, they should align with $P$ geometrically. The specific condition about the proportion of $PD$ and $MN$ (e.g., $PD = 4MN$) suggests a scaling between the lengths that typically would not hold in this configuration unless specifically constructed that way. Given that $MN$ is essentially the midpoint line segment connecting the midpoints of line segments across the plane, and without additional information to contradict it, one would normally conclude that the relative distances and alignments make it unlikely for $PD$ to be exactly four times $MN$ under normal geometric conditions unless additional relationships or alignments are specified.  **Answer Choice:** The analysis leads us to option **D**:  $P , D , M , N$  The four points are not coplanar, and  $P D \neq 4 M N$. This is because, without additional geometric constraints or alignments explicitly stated that could enforce a direct proportional relationship or coplanar alignment, the general geometric arrangement of such points does not typically satisfy both coplanarity and the specified proportionality between $PD$ and $MN$.
\end{mybody}
%%%%%%%%%%%%%%%%%%%%%%%%%%%%%%%%%%%%%%%%%%%%%%%%%%%%%%%%%
\label{table: visual aids inference capability - fail reasoning discrepency}
\end{table*}
%%%%%%%%%%%%%%%%%%%%%%%%%%%%%%%%%%%%%%%%%%%%%%%%%%%%%%%%%
%%%%%%%%%%%%%%%%%%%%%%%%%%%%%%%%%%%%%%%%%%%%%%%%%%%%%%%%%

\section{Supplementary Quantitative Analysis}
\begin{figure}[hbp]
  \centering
  % \vspace{-2cm}
  \begin{subfigure}[b]{0.9\columnwidth}
    \centering
    \includegraphics[width=\linewidth]{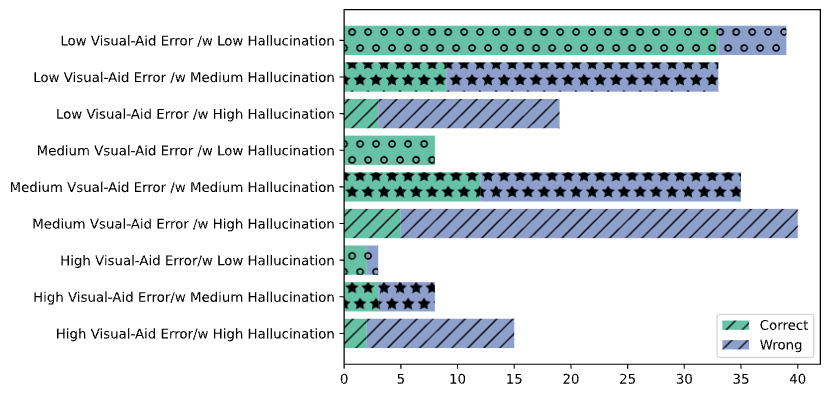}
    \caption{Correlation between visual aid and reasoning hallucination.}
    \label{Figure: Correlation Analysis of Visual-Aid and Reasoning Hallucination}
  \end{subfigure}
  \begin{subfigure}[b]{0.9\columnwidth}
    \centering
    \includegraphics[width=\linewidth]{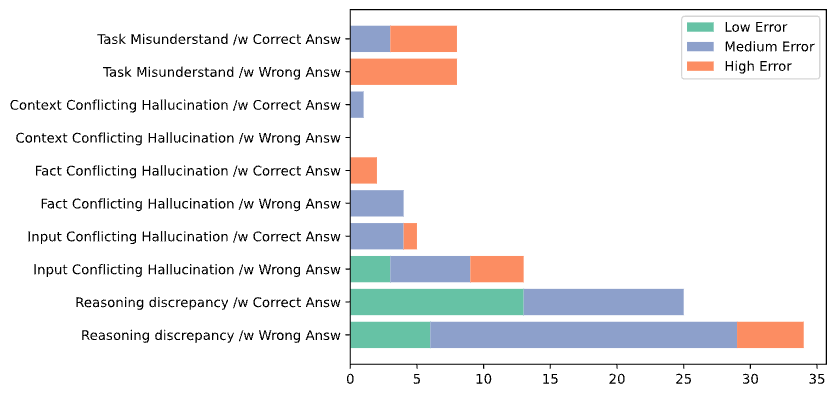}
    \caption{Correlation between error reasons of visual aid and answer correctness.}
    \label{Figure: Correlation Analysis of Visual-Aid Error Reason and Answer Correctness}
  \end{subfigure}
  \caption{Error analysis of direct visual-aided reasoning task (CQ2VA task, GPT-4V).}
  % \vspace{-2.5cm}
\end{figure}

As shown in Figure \ref{Figure: Correlation Analysis of Visual-Aid and Reasoning Hallucination} and Figure \ref{Figure: Correlation Analysis of Visual-Aid Error Rate and Answer Correctness}, we analyze reasoning patterns, hallucination correlation, and failure cases of visual-aid generation.

\paragraph{Reasoning Pattern} 
We evaluate only the model outputs with correct answers, examining how GPT4V responds to questions without being instructed to generate visual aids as an intermediate step. We found that GPT4V did not prefer generating visual aids to simplify reasoning complexity. Expect for generating answers due to hallucinations (33.2\%), the model searches for problem-solving chains based on visual-irrelevant logic, such as reasoning with pure arithmetic calculations and backward reasoning. An example of solving the problem with pure arithmetic calculation can be found in General Response.
    
\paragraph{Failure Analysis of Visual Aids Generation}
We investigated the causes of poorly generated visual aids in the CQ2VA task (i.e., generating visual aids before reasoning). Based on the analysis results, we provide potential research direction for improving the quality of visual aid generation in section \ref{Future Direction}.
    
\paragraph{Correlation between Visual-Aid and Hallucination}
We analyzed the effectiveness of visual aids in reducing hallucinations (Section \ref{Reasoning Tendency}). We first categorize the error rates of generated visual aids as low, medium, and high. The hallucination level of the subsequent reasoning is defined as low, medium high. We collect and count combinations of each category and visualize in Figure \ref{Figure: Correlation Analysis of Visual-Aid and Reasoning Hallucination}, including the percentage of correct and incorrect answers. From Figure \ref{Figure: Correlation Analysis of Visual-Aid and Reasoning Hallucination}, we observe that models with lower visual-aid error rates (i.e., output more complete visual aids) can generate more truthful outputs instead of hallucinations, thus increasing the success rate of problem-solving. This phenomenon indicates that correct visual aids can effectively alleviate hallucinations during reasoning.
    
\paragraph{Failure Analysis of Direct Visual-Aided Reasoning.}
We analyzed how models react to poorly generated visual aids. In Figure \ref{Figure: Correlation Analysis of Visual-Aid Error Rate and Answer Correctness}, the sample size of failure cases is much larger than size of successful cases, highlighting a lack of capability in direct visual-aided reasoning. By comparing sample sizes across different visual aid error rates, we find that poor visual aids with more differences from reference can negatively affect subsequent reasoning. We also summarize the error types in generating visual aids (in Section \ref{Reasoning Tendency}), exploring possible directions to improve MLLMs.

\paragraph{Correlation between Visual-Aid and Hallucination}
Visual aids serve as intermediate reasoning steps within the visual dimension, revealing hidden properties that facilitate specific reasoning paths and reduce the overall difficulty of reasoning. To analyze the effectiveness of visual aids in reducing hallucination, we asked annotators to evaluate 200 samples from the previous analysis batch (see Section \ref{Analysis: Failure Analysis of Visual-Aid Generation}) and estimate the severity of \textit{Visual-Aid Error} and \textit{Hallucination}. The correlation between visual aids and reasoning hallucination is presented in Figure \ref{Figure: Correlation Analysis of Visual-Aid and Reasoning Hallucination}. Our findings indicate that correct visual aids can effectively alleviate hallucinations during reasoning and significantly increase the success rate of the reasoning process.

\paragraph{Failure Analysis of Direct direct visual-aided Reasoning}\label{Analysis: Failure Analysis of Visual-Aid Generation}
Finally, to investigate the underlying interconnection between visual aids and final answers, we annotated the \textbf{Visual-Aid Error Types} and \textbf{Answer Correctness} separately. It is observed that visual aids with significant disparities are more likely to result in reasoning collapse, as shown in Figure \ref{Figure: Correlation Analysis of Visual-Aid Error Rate and Answer Correctness}. Notably, when generated visual aids exhibit low error severity, the quantity of correct answers does not significantly exceed that of incorrect answers.
Comparing Figure \ref{Figure: Correlation Analysis of Visual-Aid and Reasoning Hallucination} with Figure \ref{Figure: Correlation Analysis of Visual-Aid Error Rate and Answer Correctness}, we observe a strong relationship between incorrect answers and hallucinations in both reasoning and visual-aid generation.

\section{Related Work}\label{related work}
\paragraph{Benchmark} Numerous benchmarks have been developed to evaluate mathematical reasoning abilities in both textual and multi-modal tasks. These benchmarks primarily rely on textual inference as the reasoning tool. Regarding the text-only task, arithmetic problems with pure numerical expressions \cite{yuan2023well} and MPS \cite{yue2023mammoth} have been extensively explored. On the multi-modal side, \cite{chen2022unigeo, lu2021inter} focus on geometry problems to foster research on spatial understanding and properties deduction . Other multi-modal benchmarks concentrate on general visual contexts, such as bar charts \cite{masry2022chartqa}. More recently, \cite{lu2023mathvista} established a comprehensive benchmark that incorporates different visual contexts. However, these benchmarks primarily rely on textual reasoning to solve mathematical problems, limiting comprehensive mathematical decision space to a singular text dimension.
In contrast, humans tend to combine visual and textual reasoning to exploit latent properties and ease the complexity of reasoning. Therefore, we propose VisAidMath benchmark, incorporating visual aids in reasoning side.

\paragraph{LLMs for Mathematical Reasoning} LLMs have not achieved satisfactory performance in mathematical domain under zero or few-shot settings \cite{rae2021scaling}. Chain-of-thought reasoning and prompt engineering methods are introduced to improve step-wise reasoning and decoding control \cite{kojima2022large}. In multi-modal setting, LLMs can leverage rich information from visual contexts for subsequent mathematical reasoning. \cite{gpt4v, liu2024llava} explore reasoning over diverse figures that contain abundant numerical and spatial information. Interaction with external tools \cite{gao2023pal} and downstream instruction tuning \cite{liu2024visual} are also widely employed to improve overall reasoning quality. A relevant idea proposed by \cite{internlm-xcomposer2} explores inter-connected text-vision reasoning by providing text content with contextually relevant images sourced from a pre-defined dataset. Contrary to our essential idea, these output images are generated to enhance content readability, rather than reasoning capabilities.

\paragraph{Multimodal Math Benchmark}\label{related work: multimodal benchmark}
Recent advancements in multimodal mathematical reasoning have led to the creation of several sophisticated benchmarks. Many of these, such as those focusing on geometry, primarily test a model's ability to interpret and reason from a given visual context. More recently, some studies have begun to explore the generation of structured outputs. For instance, benchmarks have been proposed to evaluate the generation of visual diagrams or natural language explanations as part of the solution \cite{li2025visiomath,park2025explain}. Notably, works like \cite{chervonyi2025gold,fu2025trustgeogen} have proposed neuro-symbolic data engines that can synthesize mathematical images from a set of rules and then mechanically reverse-deduce elements to serve as visual aids. However, due to their reliance on a limited set of rules, this reverse-deduction is mechanical and struggles to simulate the diverse, real-world geometry problems that genuinely require creative visual-aided reasoning. Furthermore, these approaches lack a comprehensive evaluation system for this specific capability. In contrast, our \textbf{VisAidMath} benchmark introduces a key distinction. Instead of treating the visual output as a final product or a mere explanation, \textbf{VisAidMath} is specifically designed to evaluate a model's ability to generate intermediate visual aids that actively assist in the reasoning process itself. The primary task is not just to solve the problem, but to create helpful visual tools (like drawing auxiliary lines or plotting function graphs) that simplify the path to the solution. While other benchmarks might assess reasoning from a diagram, \textbf{VisAidMath} assesses the ability to reason by creating a diagram, directly targeting the "Thinking with Images" capability where the model must construct its own visual scaffolding to solve complex mathematical problems.
\section{limitation and social impact} \label{limitation and social impact}
% \section{Limitations and Social Impact} 
The limitation of \textbf{VisAidMath} is three-folded. First, dataset is restricted to 1200 samples since both collection,  annotation and verification of mathematical problems acquire heavy manual work to satisfy dataset principles. Such mathematical problems with visual aids cost more human efforts to understand each segment before judgment. Secondly, deficiency of mainstream machine translation systems in mathematical domain could introduce various translation errors, thus enhancing complexity for problem solving and subsequent evaluation. Thirdly, we cannot conduct comprehensive analysis of visual-aided reasoning with image generation, since current LMMs remain significant deficiency in mathematical image generation.
No negative social impact will be provided from our math-centric work, expecting only to enhance further understanding of LLM reasoning.

\section{LLM Usage Declaration}\label{llm usage}
We utilized a large language model (LLM) to assist in the writing and editing process of this manuscript. The LLM's role was strictly limited to improving grammar, refining phrasing, and enhancing readability. The core research ideas, experimental design, data analysis, and scientific conclusions were conceived and executed entirely by the authors.

\end{document}